\documentclass[journal]{IEEEtran}

\usepackage{graphicx,  amsbsy, amsmath, cuted}
\usepackage{subfigure} 
\usepackage{multirow}
\usepackage{booktabs}
\usepackage{threeparttable}
\usepackage{enumitem}
\usepackage{diagbox}
\usepackage{amsfonts,amssymb}
\usepackage{algorithm}
\usepackage{algorithmic}
\usepackage{upgreek}
\usepackage{xcolor}

\def\aphy{a_{phy}}
\def\Rrs{R_{rs}}

\begin{document}

\title{\huge Variational Autoencoder Framework for Hyperspectral Retrievals (Hyper-VAE) of Phytoplankton Absorption and Chlorophyll a in Coastal Waters for NASA's EMIT and PACE Missions} 

% \author{
%     \IEEEauthorblockN{Jiadong Lou\IEEEauthorrefmark{1}, Bingqing Liu\IEEEauthorrefmark{2}, Yuanheng Xiong\IEEEauthorrefmark{3}, Xiaodong Zhang\IEEEauthorrefmark{3}, and Xu Yuan\IEEEauthorrefmark{1}}\\
%     \IEEEauthorblockA{\IEEEauthorrefmark{1}University of Delaware,  \IEEEauthorrefmark{2}University of Louisiana at Lafayette, \IEEEauthorrefmark{3}University of Southern Mississippi}
% }

\author{
    \thanks{This work was supported by NASA EMIT program through grant no. 80NSSC24K0865, as well as the NOAA NCCOS Hypoxia program under grant no. NA23NOS4780285. (Corresponding author: Bingqing Liu).}
    
    Jiadong Lou\thanks{Jiadong Lou is with Department of Computer \& Information Sciences, University of Delaware, Newark, DE 19716 USA (e-mail: loujd@udel.edu).}, 
    Bingqing Liu\thanks{Bingqing Liu is with School of Geosciences, University of Louisiana at Lafayette, Lafayette, LA 70504 USA (e-mail bingqing.liu@louisiana.edu).}, 
    Yuanheng Xiong\thanks{Yuanheng Xiong is with School of Ocean Science and Engineering, The University of Southern Mississippi, Stennis Space Center, MS 39525 USA (e-mail: yuanheng.xiong@usm.edu).}, 
    Xiaodong Zhang\thanks{Xiaodong Zhang is with School of Ocean Science and Engineering, The University of Southern Mississippi, Stennis Space Center, MS 39525 USA (e-mail: xiaodong.zhang@usm.edu)}, and
    Xu Yuan\thanks{Xu Yuan is with Department of Computer \& Information Sciences, University of Delaware, Newark, DE 19716 USA (e-mail: xyuan@udel.edu).}
}

% make the title area
\maketitle

\begin{abstract}
Phytoplankton absorb and scatter light in unique ways, subtly altering the color of water-changes that are often minor for human eyes to detect but can be captured by sensitive ocean color instruments onboard satellites from space. 
Hyperspectral sensors, paired with advanced algorithms, are expected to significantly enhance the characterization of phytoplankton community composition (PCC), especially in coastal waters where ocean color remote sensing applications have historically encountered significant challenges. 
This study presents novel machine learning-based solutions for NASA's hyperspectral missions including EMIT (7nm), and PACE (2.5 nm), tackling high-fidelity retrievals of phytoplankton absorption coefficient ($\aphy$) and chlorophyll a (Chl-a) from their hyperspectral remote sensing reflectance ($\Rrs$). 
Given that a single $\Rrs$ spectrum may correspond to varied combinations of inherent optical properties (IOPs) and associated concentrations (e.g., Chl-a), the Variational Autoencoder (VAE) is used as a backbone in this study to handle such multi-distribution prediction problems. 
We first time tailor the VAE model with innovative designs to achieve hyperspectral retrievals of  $\aphy$ and of Chl-a from hyperspectral $\Rrs$ in optically complex estuarine-coastal waters. 
Validation with extensive experimental observation demonstrates superior performance of the VAE models with high precision and low bias.
The in-depth analysis of VAE's advanced model structures and learning designs highlights the improvement and advantages of VAE-based solutions over the mixture density network (MDN) approach, particularly on high-dimensional data, such as PACE. 
Our study provides strong evidence that current EMIT and PACE hyperspectral data as well as the upcoming Surface Biology Geology (SBG) mission will open new pathways toward a better understanding of phytoplankton community dynamics in aquatic ecosystems when integrated with AI technologies.
\end{abstract}


\section{Introduction}

Phytoplankton is an extremely diverse set of microorganisms, varying in cell morphologies, biogeochemical functions, and physiological responses to environmental disturbances~\cite{henson2021future}. 
As the primary producer of ocean's food web, phytoplankton produces approximately 50 percent of Earth's oxygen, regulate the global carbon cycle and climate, and support various ecosystem services, such as fisheries, water quality, and biodiversity~\cite{cael2020information}. 
Therefore, knowledge of phytoplankton biomass and their community composition is critical to understanding the food web structure, higher trophic level production (e.g., fisheries), and biological shifts among other complex Earth Science questions, especially in the context of degraded water quality (e.g., eutrophication) and climate change (e.g., warming temperatures), demanding attention at local, regional, and global scales~\cite{chase2020evaluation}. 
As such, there is an increasing interdisciplinary interest in studying phytoplankton community dynamics in estuarine-coastal waters, where massive riverine inputs of nutrient-rich freshwaters often lead to eutrophication, harmful algal blooms (HABs), and the annual recurrence of bottom-water hypoxia events, which cause widespread and severe impacts on the aquatic ecosystem~\cite{bianchi2010science,rabalais2001hypoxia,wang2022ground, hill2020habnet,sun2013hyperspectral}.

In the field of ocean color remote sensing, the concentration of chlorophyll a (Chl-a) and phytoplankton absorption properties ($\aphy$) are two of the most commonly used metrics for assessing phytoplankton abundance and diversity in aquatic environments~\cite{pahlevan2021hyperspectral,tiwari2013evaluation}. Those phytoplankton-related satellite algorithms are rooted in the physical principle that remote sensing reflectance ($\Rrs$, $sr^{-1)}$), the ratio of water-leaving radiance to the total downwelling irradiance just above water, is determined by the inherent optical properties (IOPs), most importantly the total backscattering coefficient ($b_{total}$, $m^{-1}$) and the total absorption coefficient ($a_{total}$, $m^{-1}$)~\cite{mobley1994light}. 
These components are interrelated through the radiative transfer theory, presented here in its simplified form~\cite{mobley1994light}: 
\begin{equation}
    R_{rs}({\lambda})=c\times(\frac{bb_{total}}{a_{total}+bb_{total}})\ ,
\end{equation}
where $R_{rs}({\lambda})$ represents remote sensing reflectance at wavelength $\lambda$; c is the proportionality constant that accounts for factors such as the geometry of the observation and specific conditions of the water body. Further, $a_{total}$ can be separated into absorption by four major constituents (Roesler and Perry, 1995) including water ($a_w$), colored dissolved organic matter ($a_{CDOM}$), non-algal particle ($a_{NAP}$) and phytoplankton ($a_{phy}$), i.e.,
\begin{equation}
    a_{total}=a_w+a_{CDOM}+a_{NAP}+a_{phy}\ .
\end{equation}

Recognizing this physical principle, Chl-a, a key indicator of phytoplankton biomass, has been widely estimated from multi-spectral satellite data using semi-analytical approaches that firstly infer $a_{phy}$ peaks within the red portions of the spectrum from $R_{rs}$ and then estimate Chl-a~\cite{gons2002chlorophyll,lee2002deriving,liu2021biogeographical,schroeder2007retrieval}.  
Alternatively, empirical methods that utilize band-ratio metrics, have a long history of being employed to estimate Chl-a directly from $R_{rs}$~\cite{liu2023dissolved,khan2021spectral,sass2007understanding}. Band ratios, particularly those in the red and near-infrared (NIR) regions~\cite{gitelson1992peak}, as well as fluorescence line-height (FLH) approaches that utilize three bands in either red or red-NIR bands to compute characteristic spectral reflectance peaks, are commonly used for Chl-a estimation in estuarine and coastal waters~\cite{gurlin2011remote,li2011estimation}. Additionally, recent phytoplankton-related algorithms have increasingly favored machine learning techniques, such as Mixture Density Networks (MDNs)~\cite{pahlevan2020seamless, smith2021chlorophyll}, to enhance the accuracy and efficiency based on large amounts of historical data, such as GLORIA~\cite{lehmann2023gloria}. Pahlevan et al. (2021)~\cite{pahlevan2021hyperspectral} not only applied MDNs—originally developed for multispectral sensors (e.g., Sentinel-3 OLCI) to enhance the estimation of Chl-a and, more critically, $a_{phy}$, but also adapted this approach for the Hyperspectral Imager for the Coastal Ocean (HICO), the first spaceborne hyperspectral imaging spectrometer designed specifically for sampling coastal oceans. This adaptation paves new possibilities for ocean color algorithms, especially in the context of the increasing availability of hyperspectral missions, such as PACE (Fig. 1).

Hyperspectral sensors, with their narrow spectral bands from ultraviolet to NIR, capture far more spectral details than traditional multispectral sensors and, when paired with advanced retrieval algorithms, can extend beyond phytoplankton abundance (e.g., Chl-a) to quantify certain aspects of phytoplankton community composition (PCC). Hyperspectral $\aphy$ is a crucial optical proxy for characterizing PCC as the variability in phytoplankton cellular pigment compositions significantly affects the magnitude and spectral shape of $\aphy$, which is further reflected in $\Rrs$~\cite{bricaud2004natural,ciotti2002assessment,king2021spectral,liu2019floodwater,xi2015hyperspectral,xi2017phytoplankton}. 
Semi-analytical models~\cite{lee2002deriving,roesler2003spectral}, and machine learning approaches~\cite{pahlevan2021hyperspectral,balasubramanian2020robust,tiwari2013evaluation} have been popular methods for solving the inverse problem of estimating IOPs (e.g., $\aphy$) from $\Rrs$, most of which generally perform well in the open ocean and are well-suited to multispectral satellites. Studies focusing on the retrieval of $\aphy$ in bio-geo-chemically and optically complex estuarine and coastal waters are comparatively sparse. 
This is because coastal waters, typically rich in colored dissolved organic matter (CDOM) and/or non-algal particles (NAP), can overwhelm phytoplankton optical signals~\cite{liu2019multi,d2018galveston,d2019biogeochemical}, which further leads to uncertainty regarding the degree to which PCC information (e.g., taxa, function, or size) can be accurately gleaned from satellite data in coastal waters and the levels of confidence. 
Last but not least, different combinations of IOPs and associated concentrations that could yield similar or identical $\Rrs$ spectra~\cite{pahlevan2021hyperspectral} are further exacerbated by the scarcity of spectral bands available in current multispectral satellite sensors, complicating the inversion of IOPs (e.g., $\aphy$) from $\Rrs$ spectra.  {MDN-based machine learning approaches~\cite{pahlevan2021hyperspectral} have been proposed, bringing progress toward addressing this “one to many” inversion challenge. While MDN's performance in predicting IOPs—compared to heritage inversion algorithms such as the Quasi Analytical Algorithm (QAA)~\cite{lee2002deriving} and Generalized IOP Inversion (GIOP)~\cite{werdell2013generalized}, demonstrate their potential in coastal waters, the ``one-to-many'' issue still requires further exploration from both algorithmic and data perspectives. This is because the GLORIA dataset lacks true ``one-to-many'' features, and thus, capabilities of learning models designed to handle one-to-many inversion problems have yet to be fully explored. Nevertheless, the global Chl-a and $\aphy$ algorithms in coastal waters, especially hyperspectral retrievals of $\aphy$, are still in the early stages of development. }

\begin{figure}
	\centering
	\includegraphics[width=0.9\linewidth]{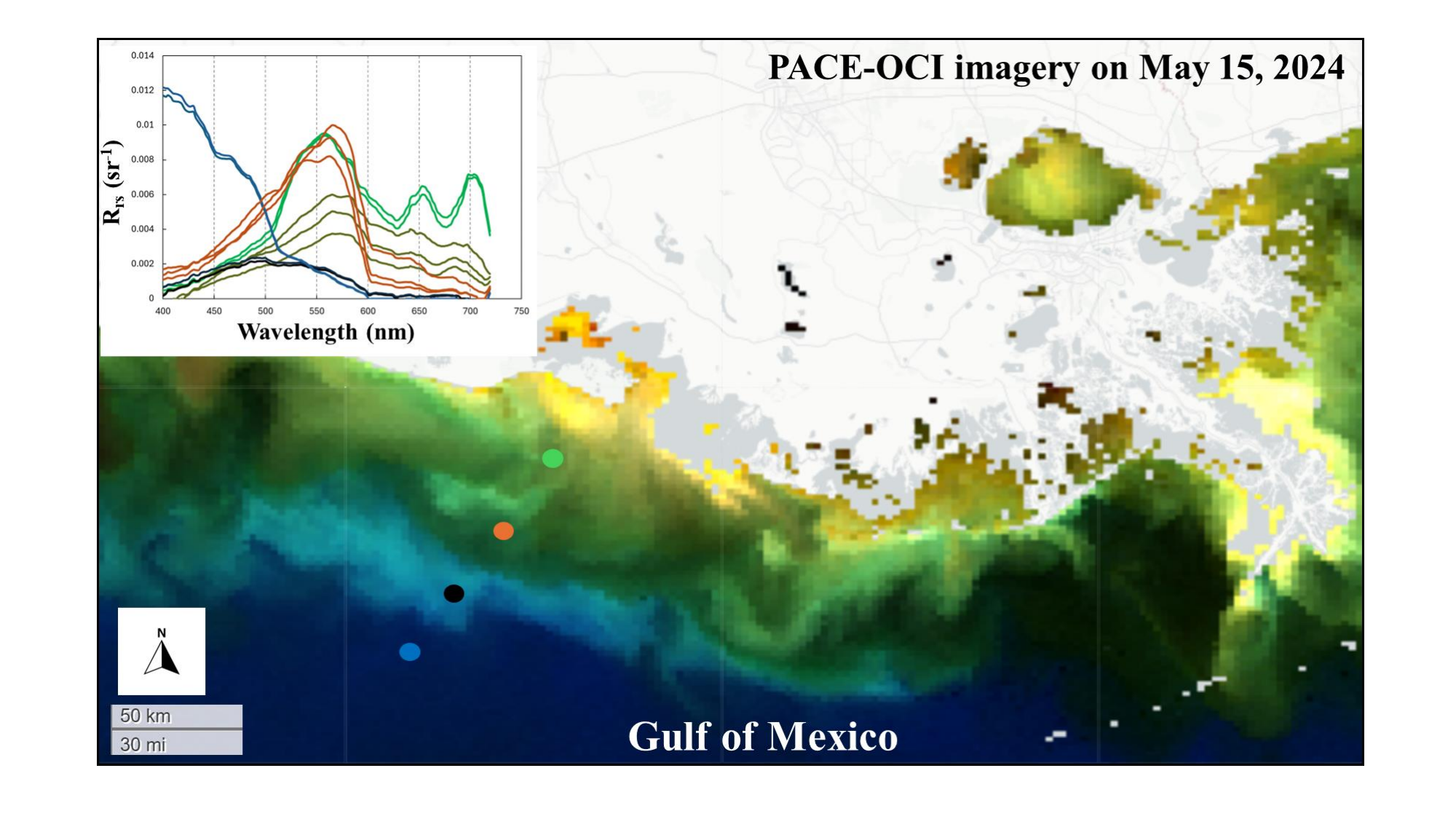}
	\caption{NASA’s PACE-OCI Level 2 AOP data obtained on May 15, 2024, covers diverse water types in the northern Gulf of Mexico. The data is displayed using a band combination of remote sensing reflectance ($\Rrs$) at 440, 560, and 650 nm and processed using the HyperCoast open-source tool (https://hypercoast.org/).}
	\label{fig:VAE}
\end{figure}

With the launch of new hyperspectral sensors like NASA's Ocean Color Instrument (OCI) on the Plankton, Aerosol, Cloud, Ocean Ecosystem (PACE) mission (Fig. 1), which features narrow spectral bands at 2.5 nm spanning ultraviolet (350 nm) to NIR (885 nm) and provides 2-day global coverage~\cite{cetinic2024phytoplankton,dierssen2021living}, there is an urgent need for global coastal algorithms to retrieve inherent IOPs and biogeochemical metrics, such as Chl-a and $\aphy$, in optically complex waters. Further, the Earth Surface Mineral Dust Source Investigation (EMIT) instrument, a precursor to the Surface Biology Geology (SBG) mission, offers both high spectral (380-2500 nm, 7.4 nm resolution) and much higher spatial (60 m) resolutions, focusing on coastal oceans~\cite{green2021nasa,sousa2023topological}. 
Both PACE and EMIT represent critical advancements for monitoring water quality in coastal zones. However, to fully utilize their potential, this is an urgent need for globally applicable algorithms for the retrieval of IOPs, tailored to these new hyperspectral missions.

{
To this end, this study aims to propose novel machine learning-based solutions, notably Variational Autoencoders (VAE)~\cite{higgins2017beta,kingma2013auto,makhzani2015adversarial} for NASA’s hyperspectral missions, e.g., PACE and EMIT, to achieve high fidelity retrievals of $\aphy$ and Chl-a from the hyperspectral $\Rrs$.}
Given the complexity of a single $\Rrs$ spectrum, which may correspond to varied combinations of IOPs and concentrations, particularly in optically complex waters, the VAE model is ideally suited as a backbone to manage such multi-distribution prediction problems because it captures the essential and high-level features of the input data, thereby effectively learning and characterizing the most important patterns. 
VAEs excel in balancing uncertainty and reliability for multi-value prediction tasks by effectively mitigating overfitting caused by noise in the training data, and learning latent patterns through a lower-dimensional latent space~\cite{chen2016variational,kingma2016improved,sonderby2016ladder,zhao2017infovae}, thereby providing robust high-dimensional of predictions. 
This manuscript will introduce the innovative design of VAE-based models tailored for PACE and EMIT, and their applications in predicting  $\aphy$ and Chl-a. 
The performance is evaluated using the dataset presented in~\cite{pahlevan2021hyperspectral,o2023hyperspectral} along with bio-optical datasets collected in Galveston Bay, TX, USA~\cite{liu2019floodwater}. 
The advantages of the VAE-based solution, including its model structure and learning designs, for hyper-retrieval predictions are also discussed.
%, targeting addressing the ``one-to-many'' problem.

\section{Dataset and Method}

\subsection{Development Dataset}

For the retrievals of $\aphy$, we utilized paired $\Rrs$-$\aphy$ data from optically complex waters in the U.S. from the SeaWiFS Bio-optical Archive and Storage System (SeaBASS) dataset in conjunction with data from lakes in Asia, all of which were compiled and shared by Pahlevan et al. (2021)~\cite{pahlevan2021hyperspectral} with additional data added from~\cite{o2023hyperspectral}. 
It was documented that the majority of $\Rrs$ spectra were acquired using the HyperOCR® radiometers, while a small portion was obtained through a handheld radiometer (e.g., ASD FieldSpec®)~\cite{mueller2003ocean}.  
All $\aphy$ spectra were measured using the standard quantitative filter pad technique (QFT)~\cite{roesler2018spectrophotometric}. 
In comparison to paired $\Rrs$-$\aphy$ dataset, the $\Rrs$-Chl-a covers a markedly wider range of Chl-a concentrations, better representing global coastal water quality conditions~\cite{pahlevan2021hyperspectral}. 

\subsection{Data Preprocessing}

\begin{figure}[h]
	\centering
	\subfigure[Spectral distribution of the $R_{rs}$]{
		\includegraphics[width=0.8\linewidth]{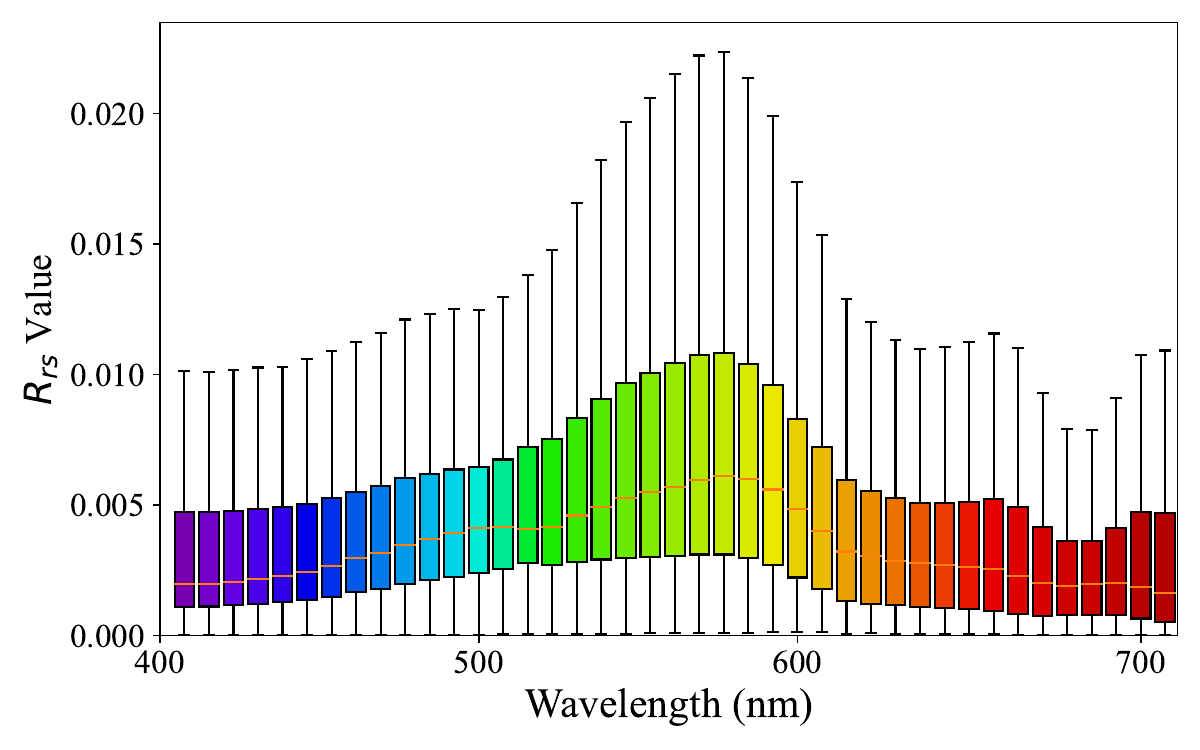}
		\label{Fig:EMIT_3_plots1}
	}	
	\subfigure[Spectral distribution of the $a_{phy}$]{
		\includegraphics[width=0.8\linewidth]{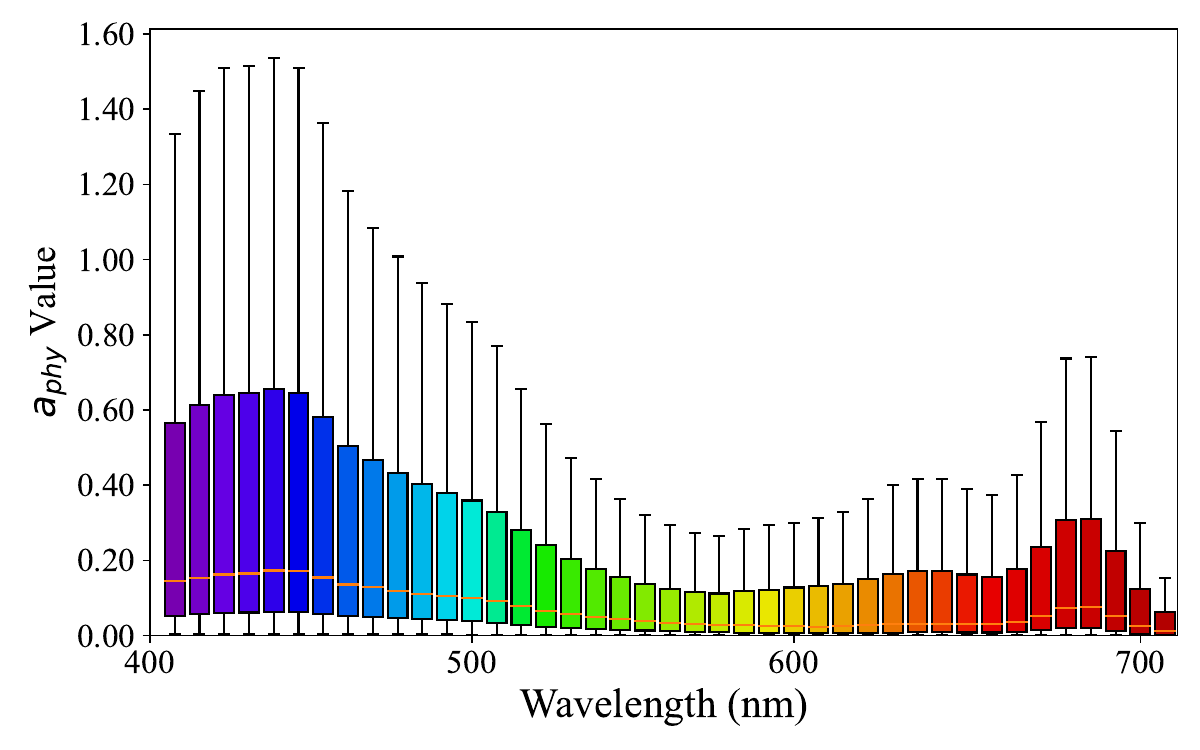}
		\label{Fig:EMIT_3_plots2}
	}
	\caption{Spectral distribution of the $\Rrs$-$\aphy$ dataset, with bars denoting the minimum and maximum values of (a) $\Rrs$ ($sr^{-1}$) and (b) $\aphy$ ($m^{-1}$) using EMIT spectral setting.  }
	\label{fig:EMIT_3_plots}
\end{figure}

Data quality control was conducted to exclude outliers (e.g. the negative values) and invalid values (e.g., NaN) from both  $\Rrs$-$\aphy$ and $\Rrs$-Chl-a datasets. 
The $\aphy$ spectra that are very noisy or exhibit zigzag patterns, were removed, and the same criteria were applied to the $\Rrs$ spectra. 
This quality control resulted in 2114 and 6111 pairs of valid $\Rrs$-$\aphy$ and $\Rrs$-Chl-a datasets, respectively.
We resampled hyperspectral $\Rrs$-$\aphy$ data to the wavelength settings corresponding to NASA's latest hyperspectral missions, including PACE (2.5 nm) and EMIT (7.4 nm) within the 400-700 nm range. 
For $\Rrs$-$\aphy$ prediction, the VAE-$\aphy$ models take 141, and 41 bands of $\Rrs$ and $\aphy$ as both input and output data for PACE, and EMIT, respectively. For $\Rrs$-Chl-a prediction, the input data for the VAE-Chl-a model consists of 141, and 41 bands of $\Rrs$ vectors for PACE, and EMIT, all with one-element output for the Chl-a value.
Fig.~\ref{fig:EMIT_3_plots} shows the spectral distribution of the $\Rrs$-$\aphy$ dataset corresponding to EMIT wavelengths. For the predictions of  $\aphy$ and Chl-a concentrations, 70\% of preprocessed data is used for training, while the remaining 30\% is reserved for testing.
Moreover, for $\Rrs$-$\aphy$ data, the min-max normalization with empirical parameters from training data is applied to each $\Rrs$ data sample, i.e., the $\Rrs$ spectra are scaled to fit into the designated value range, to account for the wide range of values across different wavelengths. 
As Chl-a is log-normally distributed in the natural environment, a logarithmic scale is performed on Chl-a to account for its natural variability.

\subsection{In-situ Hyperspectral Measurements}

Additional hyperspectral $\Rrs$-$\aphy$ field data are used to test the VAE’s robustness and generalizability. 
Phytoplankton absorption coefficient ($\aphy$) was obtained from surface water samples collected during two surveys conducted in September and October of 2017, in Galveston Bay, TX, USA. The water samples were filtered through 0.7 $\mu$m GF/F filters for the measurement of absorption by particles and non-algal particles \textit{via} a Perkin-Elmer Lambda 850 spectrophotometer fitted with a 15-cm diameter integrating sphere~\cite{liu2019floodwater} using the internally mounted sample in the IS-mode~\cite{roesler2018spectrophotometric}.
Phytoplankton absorption coefficient ($\aphy$) was then calculated as the difference between the adsorption of particles and non-algal particles~\cite{liu2021biogeographical}. The $\Rrs$ values were obtained using a GER 1500 512iHR spectroradiometer across the 350–1050 nm spectral range. At each station, sky radiance, plate radiance, and water radiance were recorded (each repeated three times) and processed to derive the above-water remote sensing reflectance~\cite{liu2019floodwater}.

{
\subsection{Revisiting MDN-based Methods}
\label{revisingMDN}
}
{
Mixture Density Networks (MDN)-based methods have been developed for IOP retrievals (e.g., Chl-a and $\aphy$) in ocean color remote sensing and tailored for applications to diverse hyperspectral missions, such as HICO and PRISMA~\cite{pahlevan2020seamless,balasubramanian2020robust,o2023hyperspectral,smith2021chlorophyll,pahlevan2021hyperspectral}. MDN is a class of neural networks designed to model complex conditional probability distributions. In contrast to traditional deterministic models, which predict a single output for a given input, MDNs learn a mixture of Gaussian distributions to capture the multi-modal nature of the target variable~\cite{pahlevan2020seamless}. 
By estimating both the mean and variance of multiple Gaussian components, MDNs can in principle include uncertainty in regression tasks. Due to this capability, MDNs have been utilized to tackle the one-to-many problem in IOPs retrieval~\cite{o2023hyperspectral,pahlevan2020seamless,pahlevan2021hyperspectral}, where the same  $\Rrs$ can correspond to multiple combinations of IOPs and associated concentrations.
Despite this potential, the existing MDN implementations~\cite{o2023hyperspectral,pahlevan2020seamless,pahlevan2021hyperspectral} 
either select the mean of the highest-weighted Gaussian component or compute a weighted average of the means of all components as the final predictions, which renders the inference process entirely deterministic. 
Once trained, the model consistently produces the same prediction for a given $\Rrs$, regardless of how many times it is executed.
While this approach ensures stability, it inherently collapses the multimodal nature of the problem to a single-point estimate, limiting MDNs' ability to address the one-to-many challenge.
Further, upon analyzing the GLORIA dataset used in the existing MDN work, it became evident that the dataset does not contain one-to-many features—no different Chl-a values or $\aphy$ spectra correspond to the same or similar $\Rrs$ spectra. This explains why the current implementations of MDNs as reported in ~\cite{o2023hyperspectral,pahlevan2020seamless,pahlevan2021hyperspectral} perform well when tested on the GLORIA and additional supplementary datasets. However, in real-world scenarios, the one-to-many problem can still arise, highlighting the need for developing methods that can more effectively address this challenge.
}

{To fully utilize the intrinsic capability of MDNs for true one-to-many mapping, rather than directly selecting the Gaussian mean with the highest weight or weighted average of the means of all components as the output as reported in previous studies, a feasible solution is to sample $\aphy$ or Chl-a from the learned mixture of Gaussian distributions. This modification ensures that the uncertainty captured in the learned distribution is effectively incorporated in MDN predictions for the same $\Rrs$ input.  However, this modification also increases variability in predictions due to the inherent stochastic nature of the sampling process. 
In Section~\ref{sec:exp}, we implement this modified MDN approach, referred to as M-MDN, and evaluate its performance in addressing one-to-many problems. }

\section{Hyper-VAE}
\label{sec:VAE}

\subsection{Variational Autoencoder}
In this study, Variational Autoencoder (VAE) was introduced and for the first time applied in the ocean color remote sensing field for inversion retrieval of IOPs and concentrations.  
As a category of generative models, VAEs were originally designed to generate new data that resemble the original input. The encoding phase aims to capture the essential high-level features of the input data and compress them into a low-dimensional representation of the input data, known as a latent vector, which characterizes the most important patterns in the data. 
In the subsequent decoding phase, this latent vector is decoded back into the original data space, enabling the reconstruction of the input data from the distributions learned in the latent space. 
This encoder-decoder process allows the VAE to generate new data that closely resemble the original input, preserving their key features while introducing variations. 
The VAE consists of two core components: an encoder and a decoder, each implemented as a neural network (Fig.~\ref{fig:VAE}).

\begin{figure*}[ht]
	\centering
	\includegraphics[width=0.75\linewidth]{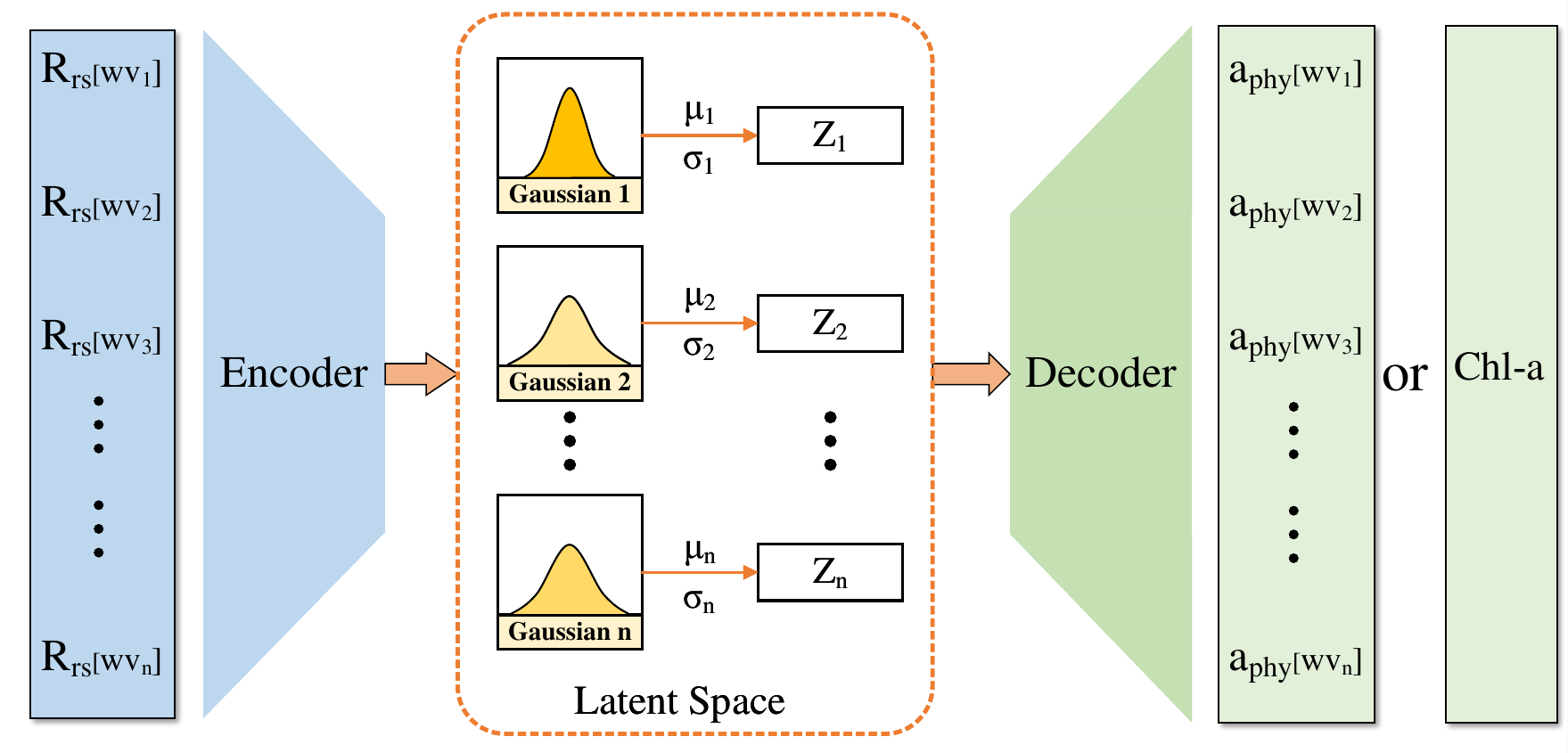}
	\caption{The structure of VAE for Predicting $a_{phy}$ and Chl-a.}
	\label{fig:VAE}
\end{figure*}

Mathematically, the encoder maps the input data,  denoted as 
$\mathbf{x}$, to a latent vector $\mathbf{z}$, where each element in $\mathbf{z}$ follows a Gaussian distribution, yielding 
\begin{equation}
    q(\mathbf{z}|\mathbf{x}) = \mathcal{N}(\mathbf{z}; \mu(\mathbf{x}), \sigma^2(\mathbf{x}))\ ,
\end{equation}
where $\mu(\mathbf{x})$  and $\sigma^2(\mathbf{x})$ represent the mean and the variance of Gaussian distributions, learned from the input data $\mathbf{x}$.

{
The key feature of the VAE that allows it to address the one-to-many prediction problem is the generation of the latent vector, where stochasticity is introduced through its latent space. Mathematically, to enable backpropagation calculation~\cite{rumelhart1986learning} during the model training process, the reparameterization trick~\cite{kingma2013auto} is typically used to sample each element $z$ in the latent vector $\mathbf{z}$, expressed as:
\begin{equation}
\label{repara}
    z = \mu({x}) + \sigma({x}) \odot \epsilon\ ,
\end{equation}
where \( \epsilon \sim \mathcal{N}(0, I) \) is a standard normal noise, and \(\odot\) denotes the element-wise multiplication.
Due to the incorporation of standard normal noise,  VAE models can effectively learn the uncertainty mapping between the $\Rrs$ and the target IOP. 
This enables the model to represent the inherent ambiguity of the inverse problem, where the same $\Rrs$ can correspond to multiple valid IOP combinations.
Furthermore, during the inference stage, the reparameterization trick is also adopted to ensure that even for the same input $\Rrs$, different latent vectors can be generated. % in different training iterations. 
This crucial property allows the VAE to produce diverse IOP predictions from the same input, effectively addressing the one-to-many mapping problem. }

{
To obtain a single output, the decoder takes the latent vector $\mathbf{z}$ as input and then outputs the reconstructed data, denoted as $\mathbf{x}^\prime$. Mathematically, the generation of $\mathbf{x}^\prime$ involves sampling from a conditional distribution $p(\mathbf{x}|\mathbf{z})$. In practice, this conditional distribution is assumed to be a Gaussian distribution with a fixed variance $\sigma^2$, while the mean is provided by the decoder output, yielding:
\begin{equation}
    p(\mathbf{x}|\mathbf{z}) = \mathcal{N}(\mathbf{x}; \mathbf{x}^\prime, \sigma^2)\ ,
\end{equation}
where $\sigma^2$ is the fixed variance.
}

{
While the parameterization step introduces noise, the decoder learns a structured mapping from $p(\mathbf{x}|\mathbf{z})$ to the desired IOP that ensures the predictions remain consistent and stable. Essentially, the decoder translates variations in the latent space into realistic, physically meaningful predictions, rather than erratic fluctuations.}

{
Furthermore, the training objective of VAE is to minimize the loss calculated from two aspects, i.e.,
\begin{equation}
    \mathcal{L}=-\mathbb{E}_{ q(\mathbf{z}|\mathbf{x})}[\log p(\mathbf{x}|\mathbf{z})]+\lambda\cdot KL\big(q(\mathbf{z}|\mathbf{x})||p(z)\big)\ ,
\end{equation}
where $\lambda$ is the weight to balance the two terms.
The loss function aligns with the standard VAE loss introduced in~\cite{kingma2013auto} in their seminal work on VAEs and has been extensively used in many VAE-based studies~\cite{higgins2017beta}. 
Here, the first term characterizes the reconstruction loss, which is practically calculated as the difference between the reconstructed data $\mathbf{x}^\prime$ and original data $\mathbf{x}$.
The second term is the KL divergence to regularize the latent space to be close to the prior distribution $p(\mathbf{z})$, which is usually chosen to be a standard normal distribution.
Assigning a weight to the KL divergence term and combining two terms together allow
for a trade-off between accurate reconstruction and meaningful latent space structure.
In our study, we carefully tuned this $\lambda$ to capture the diversity of input-output relationships while avoiding over-regularization.
%, which could lead to poor reconstructions.
}

\subsection{VAE-based Model for Predicting $a_{phy}$ and Chl-a}
While the VAE was originally designed for data generation and reconstruction, it can also be applied for prediction tasks.
{In this study, we tailor basic VAE structures into two models,
VAE-$a_{phy}$ and VAE-Chl-a, to predict $a_{phy}$ and Chl-a, separately. Since this is the first application of VAE-based models for IOP retrievals, we intentionally designed separate models for each parameter to fully explore the core capability of VAEs in one-to-many mapping. 
Notably, the model structure and design logic implemented in this study can be readily modified for the simultaneous prediction of multiple parameters, which is discussed in Section~\ref{sub:forward}.
}

In both VAE-$\mathit{a}_{phy}$ and VAE-Chl-a models,  the input is $\mathit{R}_{rs}$, and the outputs are $\mathit{a}_{phy}$ and Chl-a, respectively.
The encoders map the $\mathit{R}_{rs}$ into the latent vectors $z_{\mathit{a}_{phy}}$ and $z_{\mathit{Chla}}$, respectively, i.e.,
\begin{align}
    q(z_{\mathit{a}_{phy}}|\mathit{R}_{rs}) &= \mathcal{N}(z_{\mathit{a}_{phy}}; \mu(\mathbf{\mathit{R}_{rs}}), \sigma^2(\mathbf{\mathit{R}_{rs}}))\ , \\
    q(z_{\mathit{Chla}}|\mathit{R}_{rs}) &= \mathcal{N}(z_{\mathit{Chla}}; \mu(\mathbf{\mathit{R}_{rs}}), \sigma^2(\mathbf{\mathit{R}_{rs}}))\ . 
\end{align}
The decoders of VAE-$\mathit{a}_{phy}$ and VAE-Chl-a then map their latent vectors back to the $\mathit{a}_{phy}$ and Chl-a data space for reconstructing these data, denoted as $\mathit{a}_{phy}^\prime$ and $\mathit{Chla}^\prime$, yielding 
\begin{align}
    p(\mathit{a}_{phy}|z_{\mathit{a}_{phy}}) &= \mathcal{N}(\mathit{a}_{phy}; \mathit{a}_{phy}^\prime, \sigma^2)\ , \\
    p(\mathit{Chla}|z_{\mathit{Chla}}) &= \mathcal{N}(\mathit{Chla}; \mathit{Chla}^\prime, \sigma^2)\ . 
\end{align}
Hence, the two VAE models will output the reconstructed $\mathit{a}_{phy}^{\prime}$ and $\mathit{Chla}^{\prime}$, respectively, which are able to further predict the $\mathit{a}_{phy}$ and Chl-a values from new $\mathit{R}_{rs}$ data.  During the model training process, the reconstruction loss is tailored to be the L-1 loss between the predicted value $\mathit{a}_{phy}^\prime$  ($\mathit{Chla}^\prime$) and its ground truth value. The overall training loss for each VAE model is expressed as:
\begin{footnotesize}
\begin{align}
\label{loss}
    \mathcal{L}_{\mathit{a}_{phy}} &= ||\mathit{a}_{phy}^{\prime}-\mathit{a}_{phy}||_1 + \lambda\cdot KL\big(q(z_{\mathit{a}_{phy}}|\mathit{R}_{rs}) || p(z_{\mathit{a}_{phy}})\big)\ ,\\
    \mathcal{L}_{\mathit{Chla}} &= ||\mathit{Chla}^{\prime}-\mathit{Chla} ||_1 + \lambda\cdot KL\big(q(z_{\mathit{Chla}}|\mathit{R}_{rs})|| p(z_{\mathit{Chla}})\big)\ ,
\end{align}
\end{footnotesize}
where $p(z_{\mathit{a}_{phy}})$ and $p(z_{\mathit{Chla}})$ are considered as the standard normal distribution.

By adapting the VAE models in this way, we harness its capability to learn meaningful latent representations, enabling effective prediction of $\mathit{a}_{phy}$ and Chl-a.

\subsection{VAE-$\mathit{a}_{phy}$ and VAE-Chl-a Models Configurations}
To accommodate different data dimension requirements, the configuration of VAE-$\mathit{a}_{phy}$ and VAE-Chl-a models vary, with their parameters listed in Tables~\ref{Tab:model:aph} and~\ref{Tab:model:chla}.

For VAE-$\mathit{a}_{phy}$, the encoder consists of two fully connected layers, each followed by Batch Normalization and a LeakyReLU activation function. 
The dimension of the input layer is indicated with a variable {input}\_{dim}, which can be flexibly adjusted according to the dimension of $R_{rs}$. 
The first layer maps the input data dimension to 512 dimensions, and the second layer maps from 512 to 1024 dimensions. 
Two parallel fully connected layers then generate a 256-dimensional latent vector, producing the means and variances for 256 Gaussian distributions by mapping the 1024-dimensional output of the encoder to the 256 dimensions.
The decoder consists of three fully connected layers. 
The first two layers, each followed by Batch Normalization and a LeakyReLU activation function, map from 256 dimensions to 512 dimensions and from 512 dimensions to 1024 dimensions, respectively. 
The last layer maps from 1024 dimensions to the output dimensions, indicated as output\_dim, which are the dimensions of predicted $\mathit{a}_{phy}$ vectors. 
Here, the Softplus activation function is applied to ensure that the value is positive.

\begin{table*}[]
\centering
\caption{Model structure for VAE-$\mathit{a}_{phy}$}
\vspace{-0.5em}
\resizebox{0.8\linewidth}{!}{
\begin{tabular}{llll}
\toprule
\multicolumn{1}{c}{\textbf{Layer Type}} & \multicolumn{1}{c}{\textbf{Input Dimension}} & \multicolumn{1}{c}{\textbf{Output Dimension}} & \multicolumn{1}{c}{\textbf{Additional Operations}} \\
\midrule
\textbf{Encoder}                        &                                              &                                               &                                                    \\
Fully Connected                       & input\_dim                                   & 512                                           & Batch Normalization,   LeakyReLU(0.2)              \\
Fully   Connected                       & 512                                          & 1024                                            & Batch Normalization,   LeakyReLU(0.2)              \\
Fully   Connected                       & 1024                                           & 256                                            & Output Mean                                        \\
Fully   Connected                       & 1024                                           & 256                                            & Output Log Variance                                \\
\midrule
\textbf{Decoder}                        &                                              &                                               &                                                    \\
Fully   Connected                       & 256                                           & 512                                            & Batch Normalization,   LeakyReLU(0.2)              \\
Fully   Connected                       & 512                                           & 1024                                           & Batch Normalization,   LeakyReLU(0.2)              \\
Fully   Connected                       & 1024                                          & output\_dim                                   & Softplus Activation   \\
\bottomrule
\end{tabular}}
\label{Tab:model:aph}
\end{table*}

\begin{table*}[]
\centering
\caption{Model structure for VAE-Chl-a}
\vspace{-0.5em}
\resizebox{0.8\linewidth}{!}{
\begin{tabular}{llll}
\toprule
\multicolumn{1}{c}{\textbf{Layer Type}} & \multicolumn{1}{c}{\textbf{Input Dimension}} & \multicolumn{1}{c}{\textbf{Output Dimension}} & \multicolumn{1}{c}{\textbf{Additional Operations}} \\
\midrule
\textbf{Encoder}                        &                                              &                                               &                                                    \\
Fully Connected                       & input\_dim                                   & 256                                           & Batch Normalization,   LeakyReLU(0.2)              \\
Fully   Connected                       & 256                                          & 128                                            & Batch Normalization,   LeakyReLU(0.2)              \\
Fully   Connected                       & 128                                           & 64                                            & Output Mean                                        \\
Fully   Connected                       & 128                                           & 64                                            & Output Log Variance                                \\
\midrule
\textbf{Decoder}                        &                                              &                                               &                                                    \\
Fully   Connected                       & 64                                           & 64                                            & Batch Normalization,   LeakyReLU(0.2)              \\
Fully   Connected                       & 64                                           & 64                                           & Batch Normalization,   LeakyReLU(0.2)              \\
Fully   Connected                       & 64                                          & 1                                   &    \\
\bottomrule
\end{tabular}}
\label{Tab:model:chla}
\end{table*}

Similarly, for VAE-Chl-a, the encoder consists of two fully connected layers, each followed by Batch Normalization and LeakyReLU activation function.  
The input layer's dimension is also indicated with a variable {input}\_{dim},
The first layer maps the input data dimensions to 256 dimensions, and the second layer maps from 256 dimensions to 128 dimensions. 
For the 64-dimensional latent vector generation, two parallel fully connected layers produce the means and variances for 64 Gaussian distributions by mapping the 128-dimensional output of the encoder to 64 dimensions.
The decoder consists of three fully connected layers. 
The first two layers, each followed by Batch Normalization and a LeakyReLU activation function, both map from 64 dimensions to 64 dimensions.  
The last layer maps from 64 dimensions to the output dimension 1, which is the predicted Chl-a value.

\subsection{Performance Metrics}
Following previous studies~\cite{o2023hyperspectral,pahlevan2020seamless}, we adopt eight metrics to comprehensively evaluate the prediction performance of our proposed VAE models. 
For the following metric formulas, we collectively denote $E$ as the predicted value of our model and $M$ as the actual value. 
Additionally, since $\mathit{a}_{phy}$ and Chl-a values span a large range, we follow the approach in~\cite{o2023hyperspectral,pahlevan2020seamless} by employing a logarithmic transformation for calculating several metrics, where `log' represents $\log_{10}$. 
These evaluation metrics can be categorized into three groups.

{First, we use Mean Absolute Log-Error (MALE), Root Mean Square Error (RMSE), and Root Mean Square Log-Error (RMSLE) to assess prediction errors, i.e., the magnitude difference between the predicted values and the actual values. These metrics are calculated as below,}
%by Eqs.~(\ref{Eqn:MALE}),~(\ref{Eqn:RMSE}), and (\ref{Eqn:RMSLE})}, 
%\begin{small}
\begin{flalign}
    &\text{MALE} = 10^Y, \ Y = \frac{\sum_{i=1}^{N} |\log(E_i) - \log(M_i)|}{n}, \label{Eqn:MALE}\\
    &\text{RMSE} = \left[ \frac{\sum_{i=1}^{N} (E_i - M_i)^2}{n} \right]^{\frac{1}{2}},\label{Eqn:RMSE} \\
    &\text{RMSLE} = \left[ \frac{\sum_{i=1}^{N} (\log(E_i) - \log(M_i))^2}{n} \right]^{\frac{1}{2}}.
    \label{Eqn:RMSLE}
\end{flalign}
The three metrics collectively provide a comprehensive evaluation of a model's performance by capturing different aspects of the error distribution. Specifically, MALE calculates the average magnitude of errors, providing the most straightforward measure of overall prediction accuracy. RMSE emphasizes larger errors by squaring them before averaging,  highlighting whether a model suffers from larger deviations in some $\mathit{a}_{phy}$ and Chl-a predictions. Furthermore, RMSLE captures the logarithmic differences between predicted and actual values, making it particularly useful for the wide range of $\mathit{a}_{phy}$ and Chl-a values. By applying a logarithmic transformation, RMSLE reduces the influence of large errors relative to small ones, providing a more balanced assessment of prediction accuracy. The closer these three metrics are to zero, the better the model's performance, indicating higher precision in predictions.

Second, we adopt another two metrics, i.e.,  { Log-Bias} and {Slope}, to evaluate whether there is a systematic deviation between the predicted values and true values. The { Log-Bias} is calculated as follows:
\begin{flalign}
&\text{Log-Bias} = 10^Z, \ Z = \frac{\sum_{i=1}^{N} (\log(E_i) - \log(M_i))}{n}. \label{Eqn:bias}
\end{flalign}
It reflects the overall direction of prediction bias in VAE models. Additionally, {\em Slope} is determined as the slope of the line derived from the linear regression of predicted values versus actual values, expressed as
\begin{flalign}
&\text{Slope} = \frac{\sum_{i=1}^{N} (M_i - \bar{M})(E_i - \bar{E})}{\sum_{i=1}^{N} (M_i - \bar{M})^2}, \label{Eqn:slope
}
\end{flalign}
where $\bar{M}$ and $\bar{E}$ respectively represent the means of the actual and predicted values. The closer the two metrics are to 1, the better the model's performance, indicating that the model does not systematically overestimate or underestimate the target values.

Third, we also follow the work~\cite{o2023hyperspectral} to report three median value-oriented metrics, {\em MAPE}, ${\epsilon}$, and ${\beta}$, as calculated below.
\begin{flalign}
    &\text{MAPE} = 100 \times \text{md} \left( \frac{|E - M|}{M} \right), \label{Eqn:MAPE}\\
    &\epsilon = 100 \times \left( 10^{\text{md}\left(\left| \log \left( \frac{E}{M} \right) \right|\right)} - 1 \right), \label{Eqn:eps}  \\
    &\beta = 100 \times \text{sign}(Z) ( 10^{|Z|} - 1 ),  \ Z = \text{md} \left( \log \frac{{E}}{M} \right)\label{Eqn:beta},
\end{flalign}
%\end{small}
where $md(\cdot)$ represents the median value.
These three metrics assess the prediction error and bias of the median value, aiming to minimize the impact of prediction outliers. However, it is worth noting that outliers in the prediction of $\mathit{a}_{phy}$ and Chl-a are important, as they sometimes reflect extreme water conditions, such as algal blooms. Therefore, relying solely on the median value cannot provide a comprehensive assessment of model performance.

\begin{figure*}[h]
	\centering
	\subfigure[VAE on PACE 440 nm]{
		\includegraphics[width=0.22\linewidth]{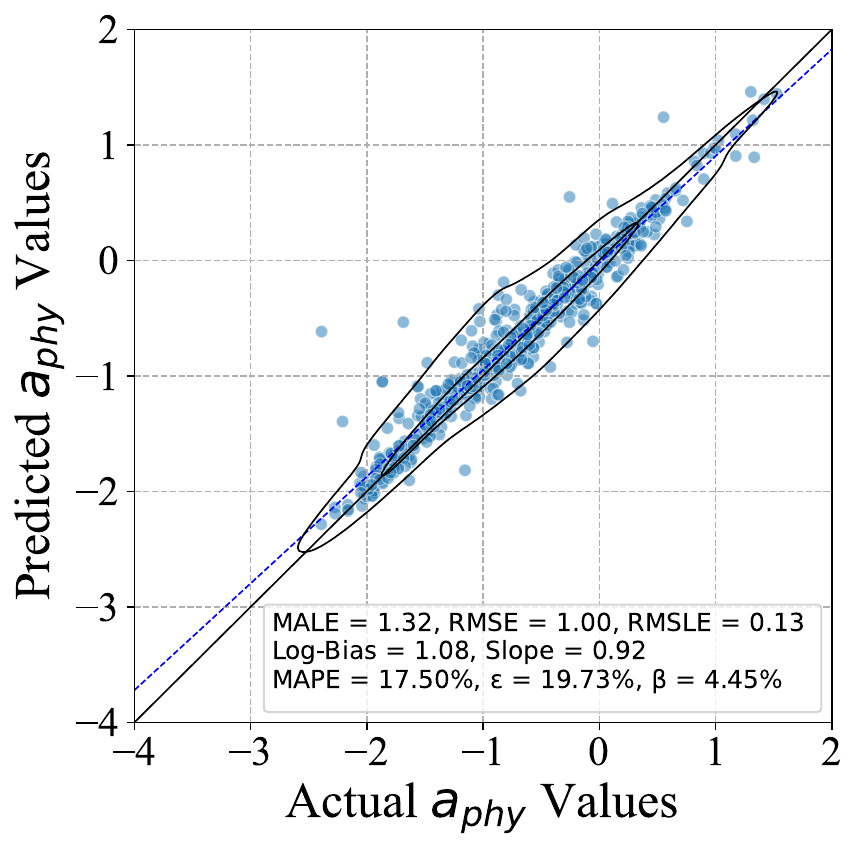}
		\label{Fig:PACE_3_plots1}
	}	
        \subfigure[MDN on PACE 440 nm]{
		\includegraphics[width=0.22\linewidth]{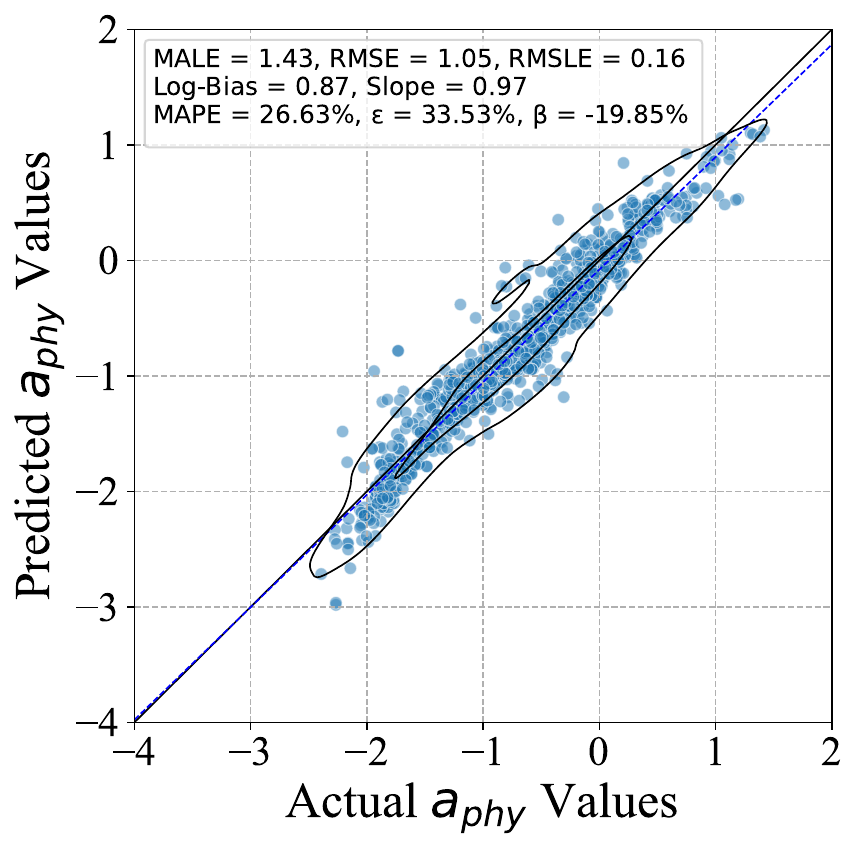}
		\label{Fig:PACE_3_plots4}
	}
 	\subfigure[VAE on EMIT 440 nm]{
		\includegraphics[width=0.22\linewidth]{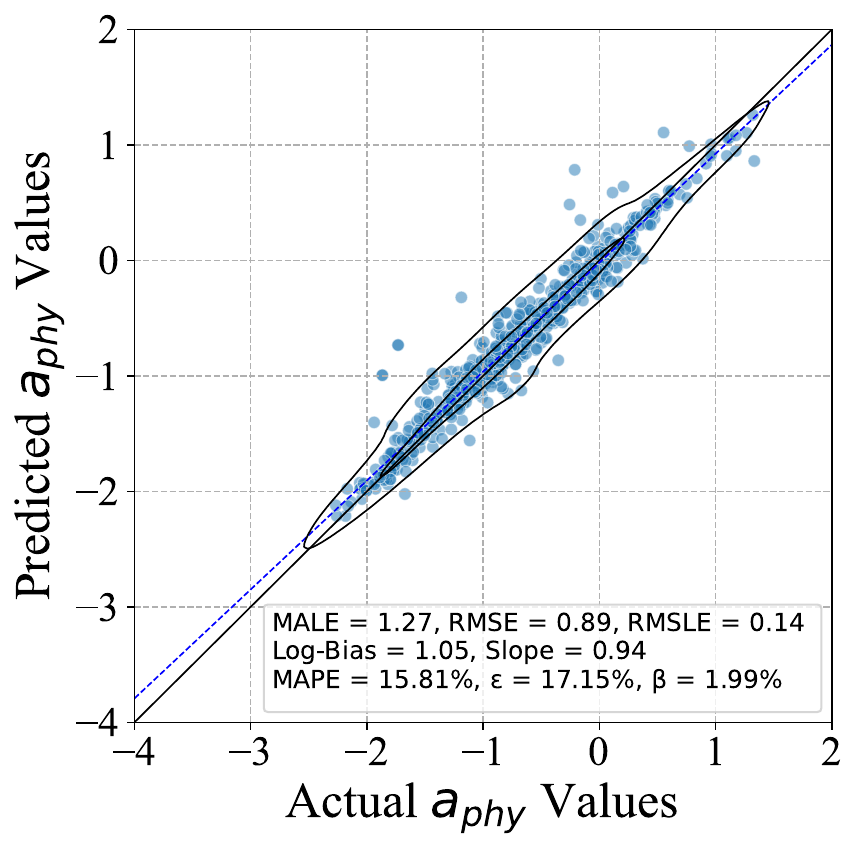}
		\label{Fig:EMIT_3_plots1}
	}	
 	\subfigure[MDN on EMIT 440 nm]{
		\includegraphics[width=0.22\linewidth]{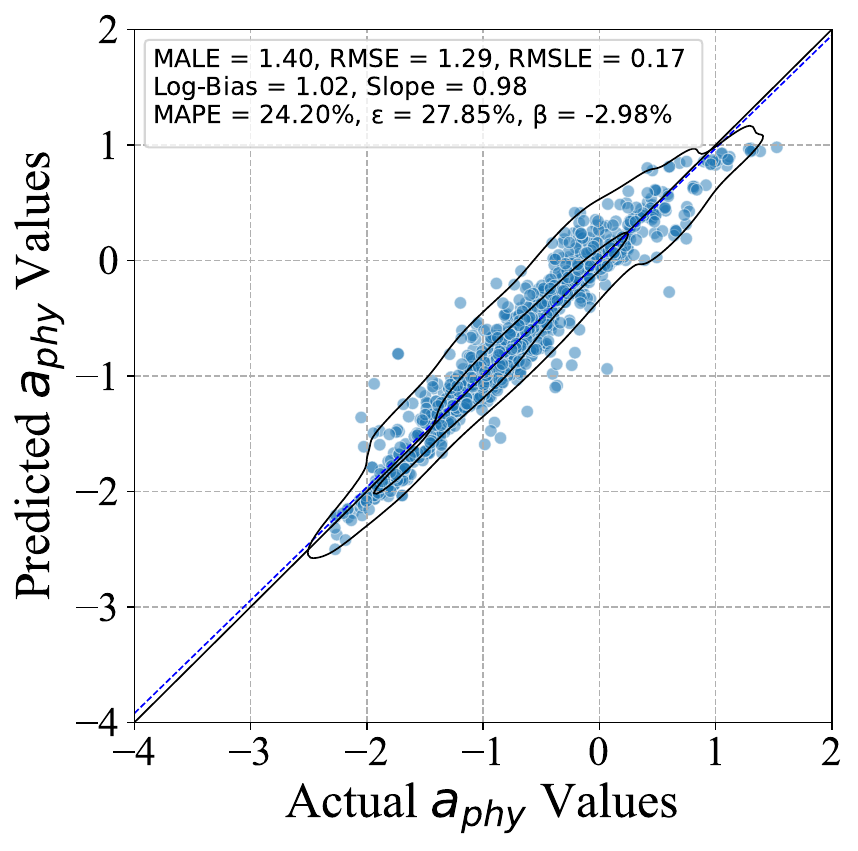}
		\label{Fig:EMIT_3_plots4}
	}
	\subfigure[VAE on PACE 620 nm]{
		\includegraphics[width=0.22\linewidth]{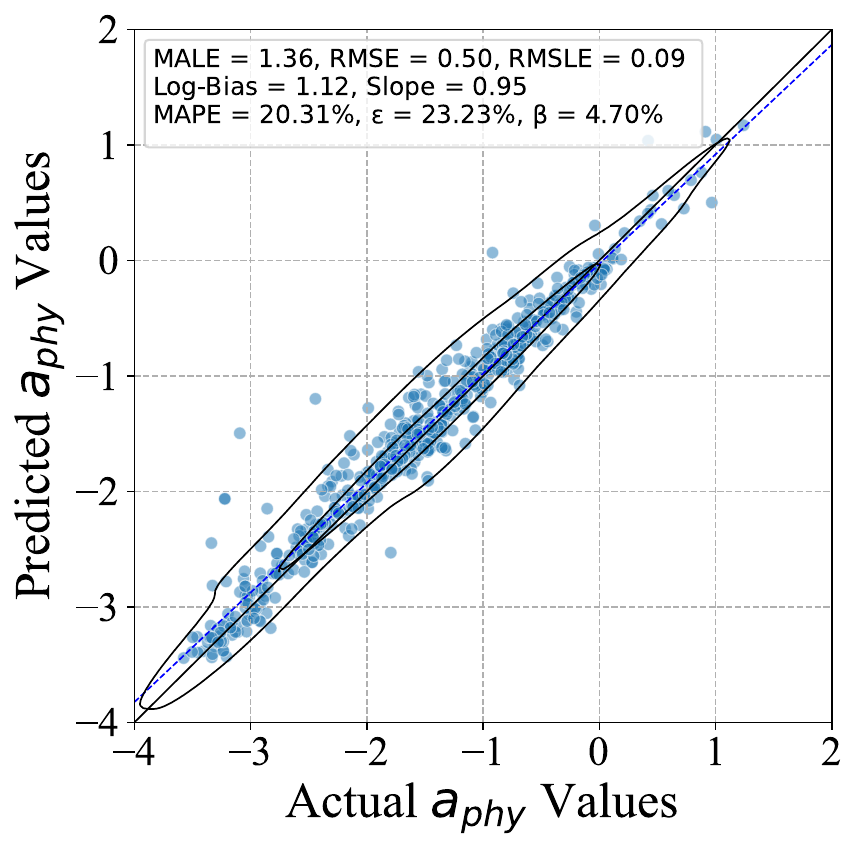}
		\label{Fig:PACE_3_plots2}
	}
 	\subfigure[MDN on PACE 620 nm]{
		\includegraphics[width=0.22\linewidth]{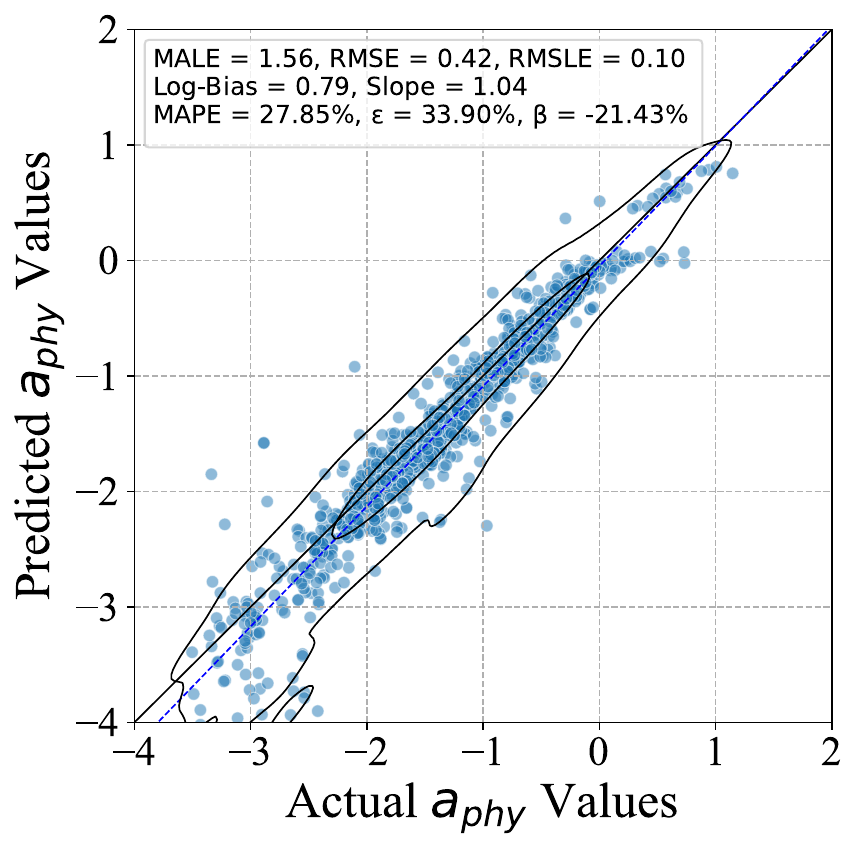}
		\label{Fig:PACE_3_plots5}
	}	
 	\subfigure[VAE on EMIT 618 nm]{
		\includegraphics[width=0.22\linewidth]{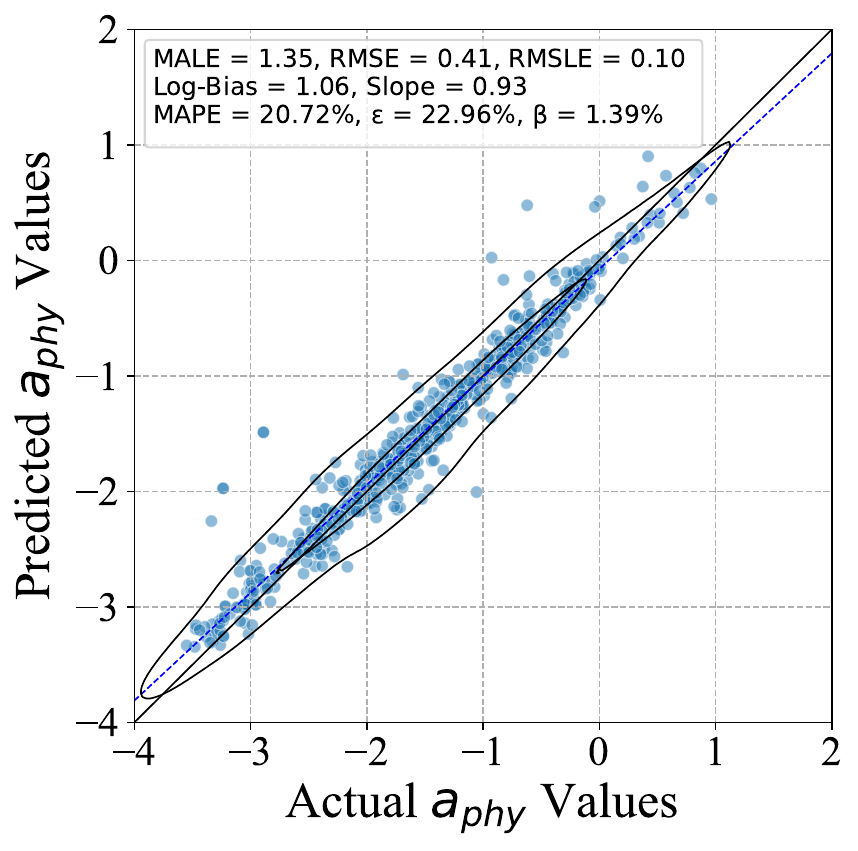}
		\label{Fig:EMIT_3_plots2}
	}
	\subfigure[MDN on EMIT 618 nm]{
		\includegraphics[width=0.22\linewidth]{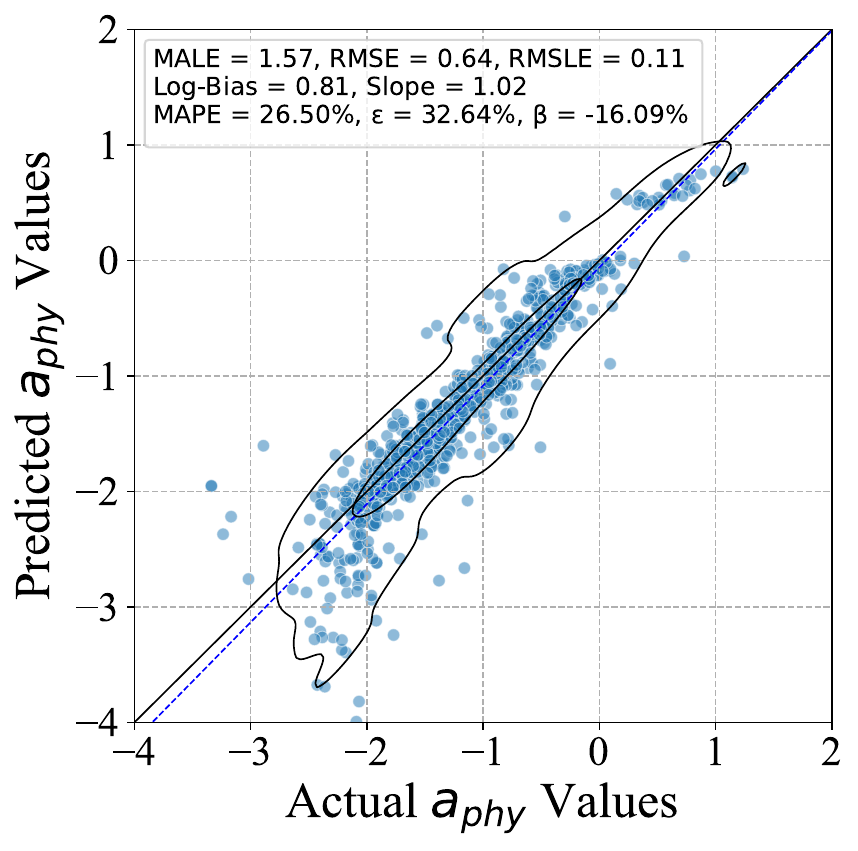}
		\label{Fig:EMIT_3_plots5}
	}	
	\subfigure[VAE on PACE 670 nm]{
		\includegraphics[width=0.22\linewidth]{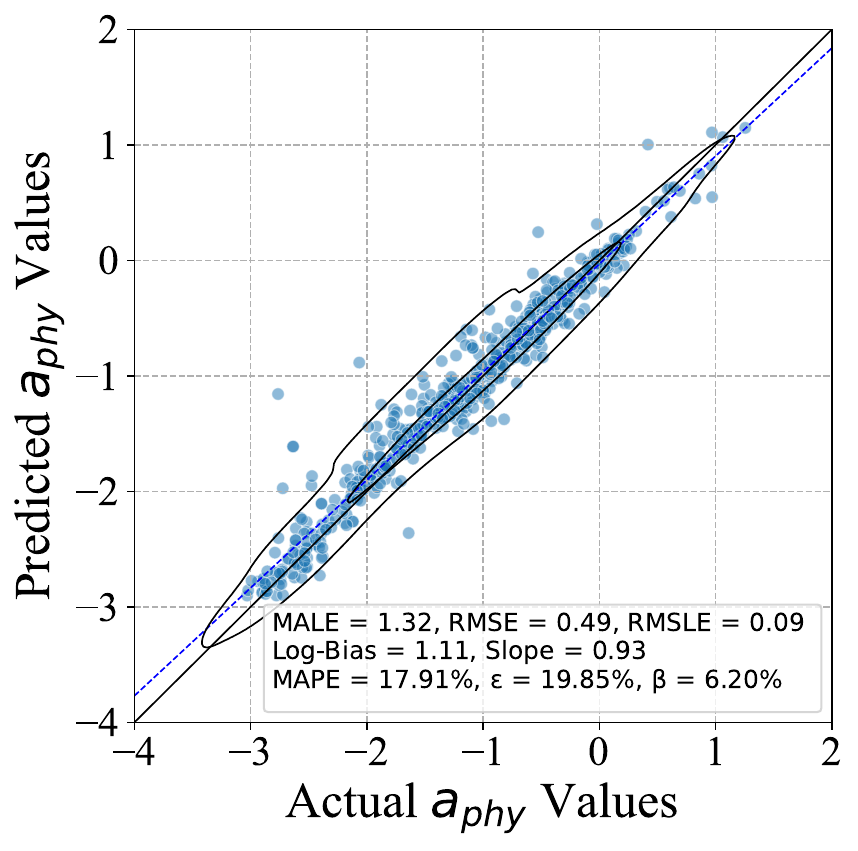}
		\label{Fig:PACE_3_plots3}
	}	
	\subfigure[MDN on PACE 670 nm]{
		\includegraphics[width=0.22\linewidth]{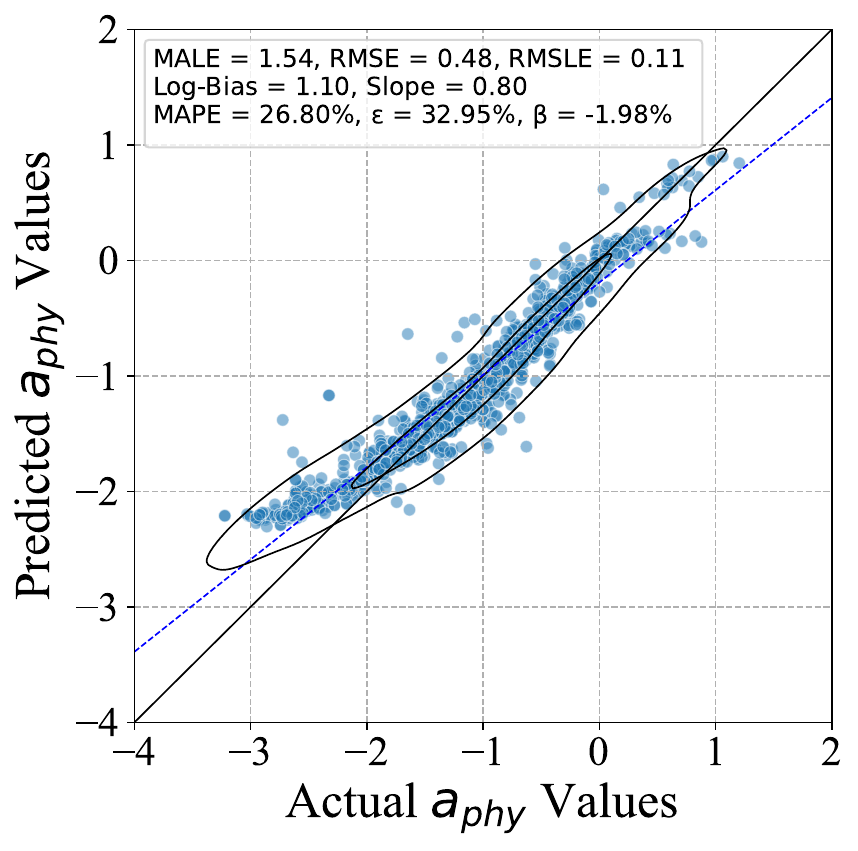}
		\label{Fig:PACE_3_plots6}
	}
 	\subfigure[VAE on EMIT 671 nm]{
		\includegraphics[width=0.22\linewidth]{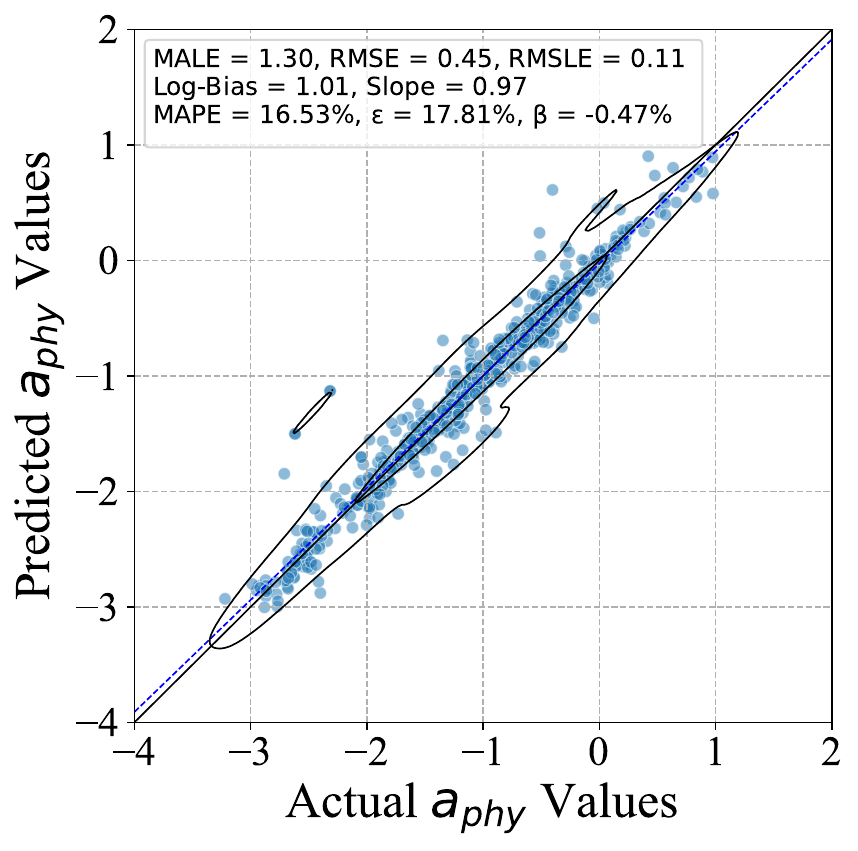}
		\label{Fig:EMIT_3_plots3}
	}	
	\subfigure[MDN on EMIT 671 nm]{
		\includegraphics[width=0.22\linewidth]{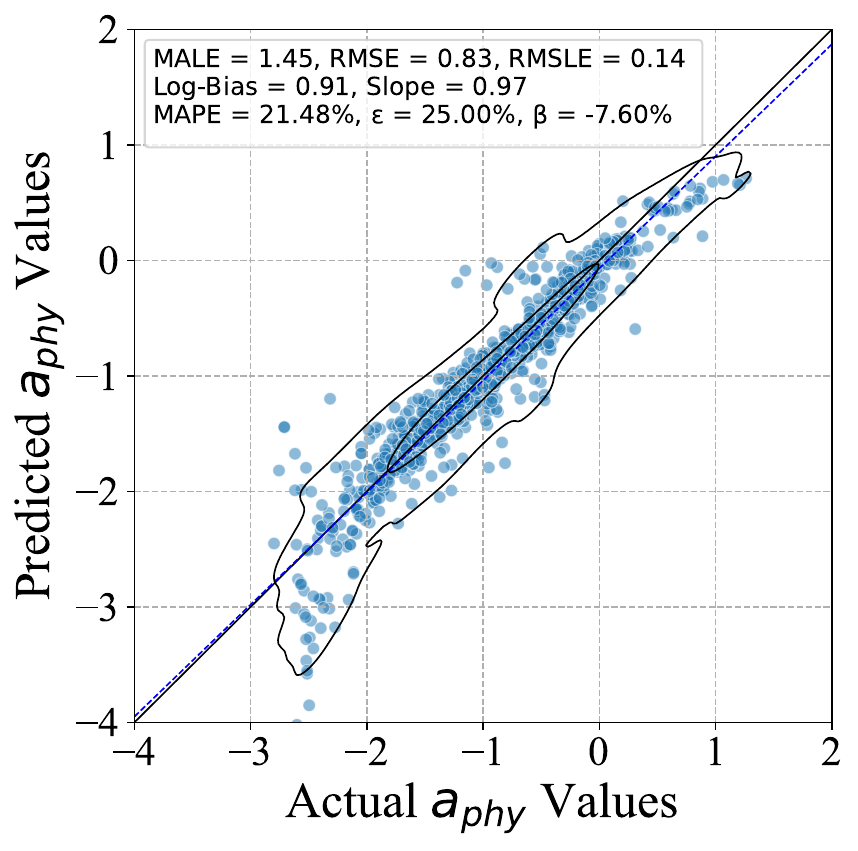}
		\label{Fig:EMIT_3_plots6}
	}
	\caption{Scatter plots and evaluation metrics for VAE and MDN predictions of $\aphy$ at three representative wavelengths for PACE and EMIT. (a)–(c) VAE for PACE and EMIT at 440 nm, (d)–(f) MDN for PACE and EMIT at 440 nm, (g)–(i) VAE for PACE and EMIT at ~620 nm, and (j)–(l) MDN for PACE and EMIT at ~670 nm.}
	\label{fig:PACE_EMIT_plots}
\end{figure*}

% \begin{figure*}[h]
% 	\centering
% 	\subfigure[VAE on 444nm]{
% 		\includegraphics[width=0.3\linewidth]{figs/HICO_444nm.pdf}
% 		\label{Fig:HICO_3_plots_1}
% 	}	
% 	\subfigure[VAE on 621nm]{
% 		\includegraphics[width=0.3\linewidth]{figs/HICO_621nm.pdf}
% 		\label{Fig:HICO_3_plots_2}
% 	}
% 	\subfigure[VAE on 673nm]{
% 		\includegraphics[width=0.3\linewidth]{figs/HICO_673nm.pdf}
% 		\label{Fig:HICO_3_plots_3}
% 	}	
% 	\subfigure[MDN on 444nm]{
% 		\includegraphics[width=0.3\linewidth]{figs/MDN_HICO_444nm.pdf}
% 		\label{Fig:HICO_3_plots_4}
% 	}
% 	\subfigure[MDN on 621nm]{
% 		\includegraphics[width=0.3\linewidth]{figs/MDN_HICO_621nm.pdf}
% 		\label{Fig:HICO_3_plots_5}
% 	}	
% 	\subfigure[MDN on 673nm]{
% 		\includegraphics[width=0.3\linewidth]{figs/MDN_HICO_673nm.pdf}
% 		\label{Fig:HICO_3_plots_6}
% 	}
% 	\caption{Scatter plots and corresponding evaluation metrics for our VAE-$\aphy$ and MDN in $\aphy$ prediction tasks at three representative wavelengths for HICO, i.e., 444nm, 621nm, and 673nm.}
% 	\label{fig:HICO_3_plots}
% \end{figure*}

\section{Results}
\label{sec:exp}

{
\subsection{MDN Basline Settings }
To ensure a comprehensive comparison with prior work, we train the MDN model following the same approach as described in~\cite{o2023hyperspectral}. 
Specifically, we reproduce their model based on the reported settings, including layer parameters, the number of Gaussian distributions, and the selection strategy of the highest-weighted Gaussian mean instead of sampling. 
The model consists of five layers, each with 100 neurons and five Gaussian distribution components. 
However, differences in dataset preprocessing and hardware environment compared to prior work necessitate adjustments to the learning rate and training epochs to achieve optimal performance.
The learning rate is set to $1e^{-3}$ with Adam optimizer and the training epoch is 2000, where the model with the best performance among these 2000 epochs is stored for use in prediction tasks.
This setup ensures a fair and direct comparison between both MDN implementations under consistent conditions.
Moreover, MDN involves matrix calculations where the full covariance matrix becomes increasingly unstable and difficult to optimize as the input dimensionality grows. 
Specifically, the number of parameters in a full covariance matrix grows quadratically with the input dimensionality.  
Consequently, under PACE spectral settings, the covariance matrix might become ill-conditioned, leading to numerical instability. This can cause gradient explosion or vanishing gradients during optimization, leading to poor convergence. To address these issues, we replace the full covariance matrix with a diagonal covariance matrix approximation, which significantly reduces computational complexity, improves numerical stability, and facilitates more reliable optimization, ultimately enhancing the performance of MDN in high-dimensional settings. 
The prediction results from the modified version of MDN, i.e., M-MDN, as we have presented in Section~\ref{revisingMDN}, are also considered to assess the MDN-based approach's capability for one-to-many mapping. 
}

\subsection{Evaluations of VAE-$\aphy$ Predictions}
We have conducted experiments using VAE-$\aphy$ for predicting $\aphy$ at PACE and EMIT spectral settings in the range of 400-700 nm, comparing the results with those from MDN models~\cite{o2023hyperspectral}.
%Fig.~\ref{fig:PACE_EMIT_plots} presents the 
Regarding predictions of $\aphy$ for VAE and MDN models, we present three important wavelengths, including 440 nm at Chl-a blue absorption, 670 nm at Chl-a red absorption, and 620 nm at phycocyanin absorption.

Figure~\ref{fig:PACE_EMIT_plots} illustrates the performance of VAE-$\aphy$ and MDN models across eight metrics, in predicting $\aphy$ at PACE and EMIT wavelengths closest to 440 nm, 620 nm, and 670 nm. 
The results demonstrate that VAE-$\aphy$ model performs better than MDN across most metrics for both PACE and EMIT, with lower errors and improved agreement with actual $\aphy$ values. 
Specifically, for PACE wavelength, the VAE model achieves lower MALE values across all these three bands, with 1.32 at 440 nm, 1.36 at 620 nm, and 1.32 at 670 nm, while the MDN model yields comparatively higher values for PACE, at 1.43, 1.56, and 1.54, respectively. 
Besides, regarding the other two prediction error metrics, RMSE and RMSLE, VAE-$\aphy$ also exhibits smaller values, reinforcing its stable predictive accuracy. 
Beyond prediction errors, VAE-$\aphy$ also achieves better Log-Bias and slope values, remaining closer to 1, which suggests a more accurate and reliable fit between predicted and actual values. 
In contrast, MDN displays greater variability and deviation at extreme values, such as overestimation of $\aphy$ (670) for PACE, as shown in Figure~\ref{Fig:PACE_3_plots6}).
%These consistently superior performances across various metrics underscore the advanced prediction capabilities of VAE-$\aphy$ for the latest hyperspectral space missions, i.e., PACE and EMIT, demonstrating its robust generalization ability. 
Furthermore, for the three median-value oriented metrics, VAE-$\aphy$ still exhibits superior performance, except for $\beta$ for PACE at 670 nm. 
%The overall lower MAPE values further highlight the VAE’s advantage, {\bf with 17.50\% at 440 nm, 17.31\% at 620 nm, and 17.31\% at 670 nm}
The stability and accuracy of VAE in hyperspectral $\aphy$ retrievals suggest that it is a suitable machine learning framework for one-to-many inversion problems in ocean color remote sensing, highlighting its potential for future hyperspectral applications, such as GLIMR and SBG.

\begin{figure*}%[ht]
	\centering
	\subfigure[RMSE on VAE ]{
		\includegraphics[width=0.3\linewidth]{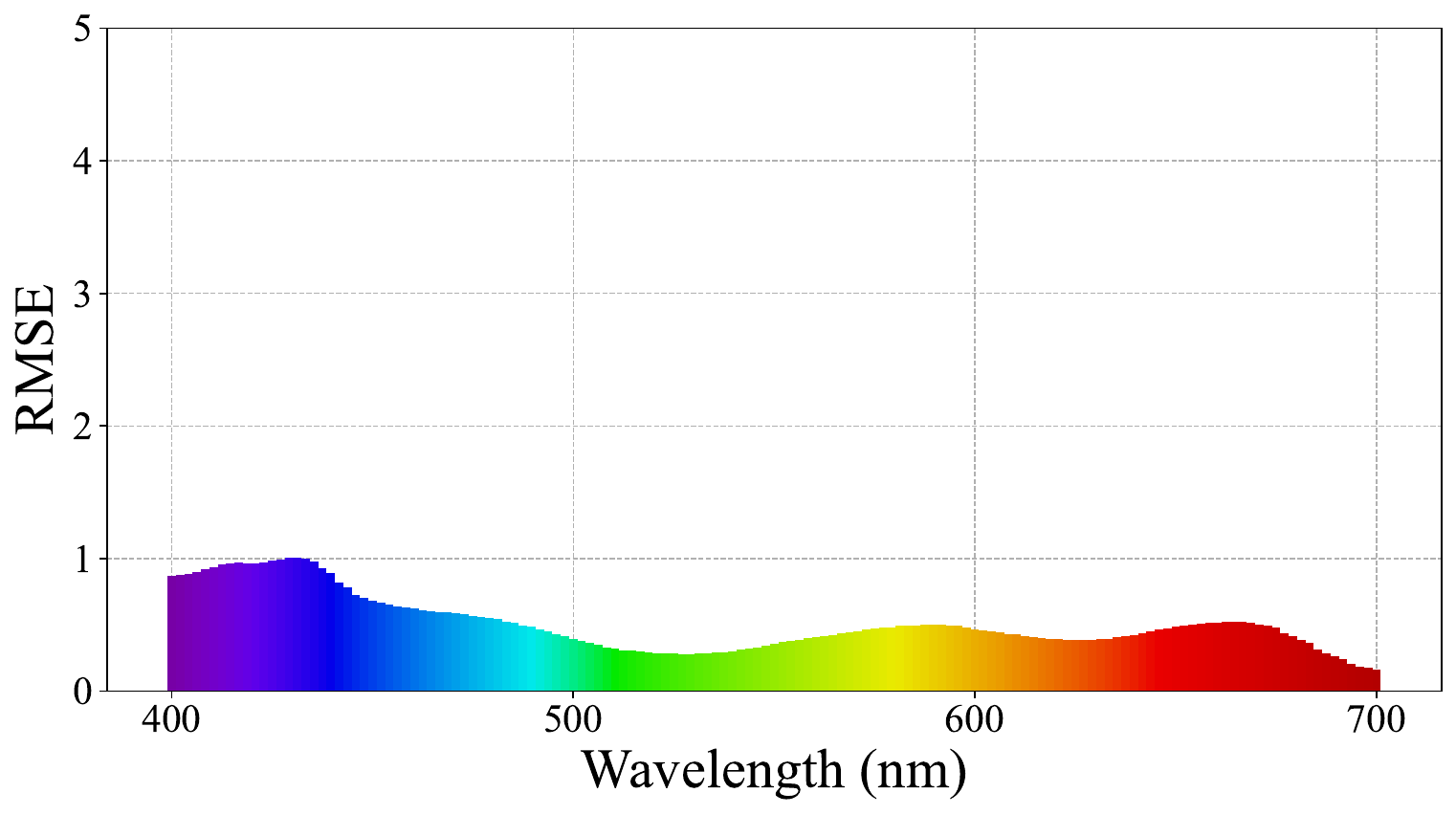}
		\label{fig:metrics:PACE:1}
	}	
	\subfigure[Log-Bias on VAE]{
		\includegraphics[width=0.3\linewidth]{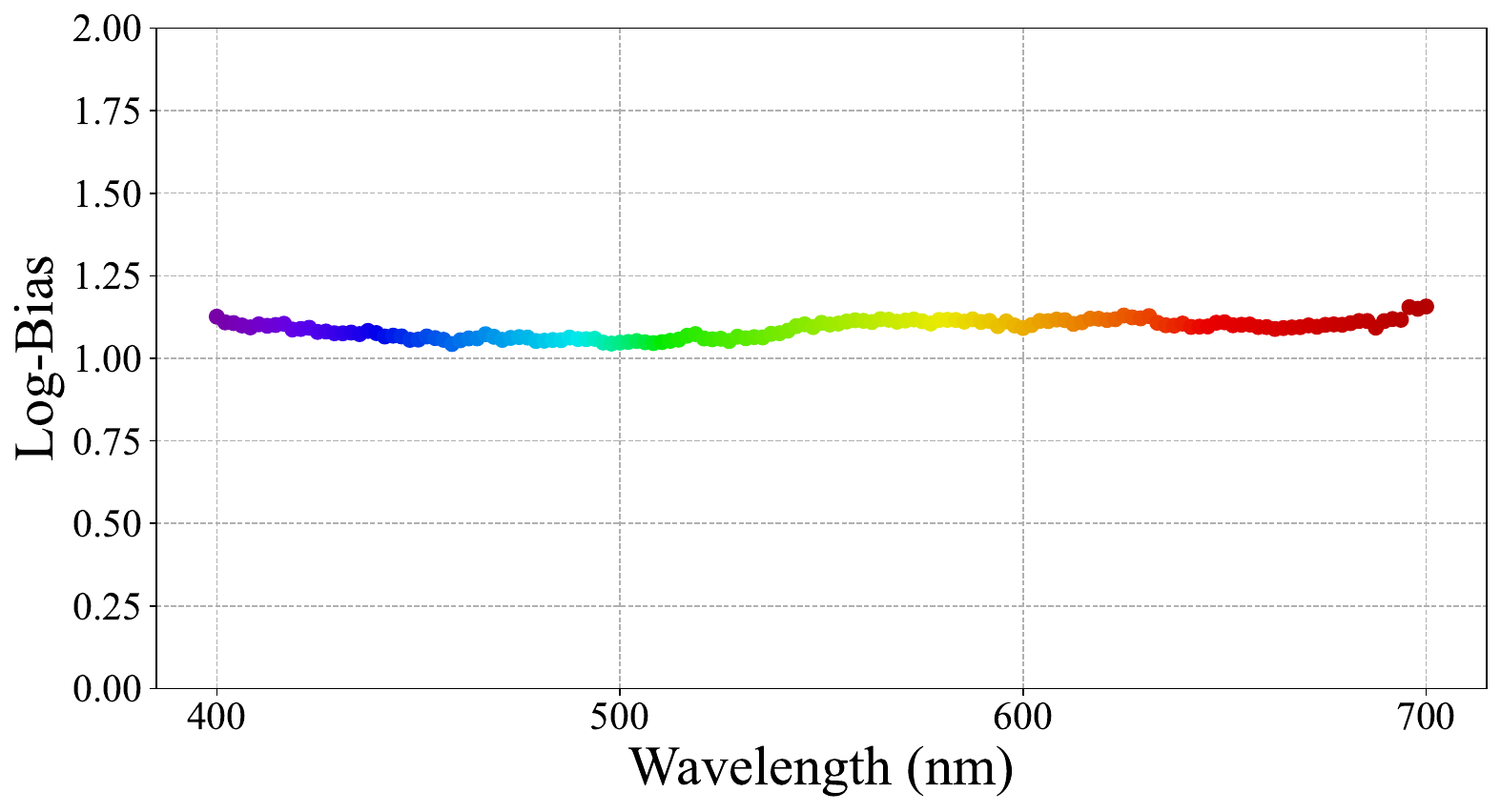}
		\label{fig:metrics:PACE:2}
	}
	\subfigure[$\beta$ on VAE]{
		\includegraphics[width=0.3\linewidth]{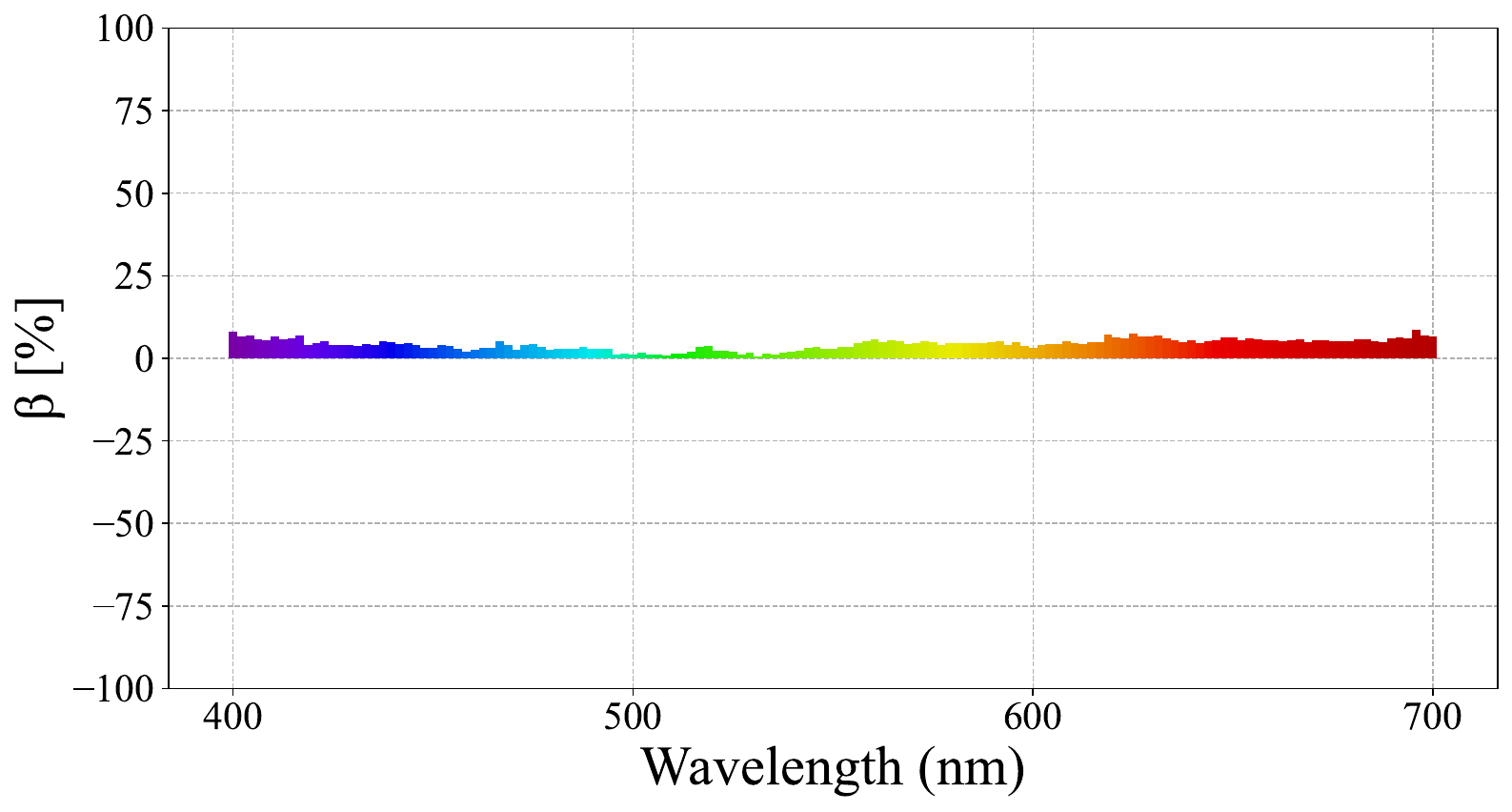}
		\label{fig:metrics:PACE:3}
	}	
	\subfigure[RMSE on MDN ]{
		\includegraphics[width=0.3\linewidth]{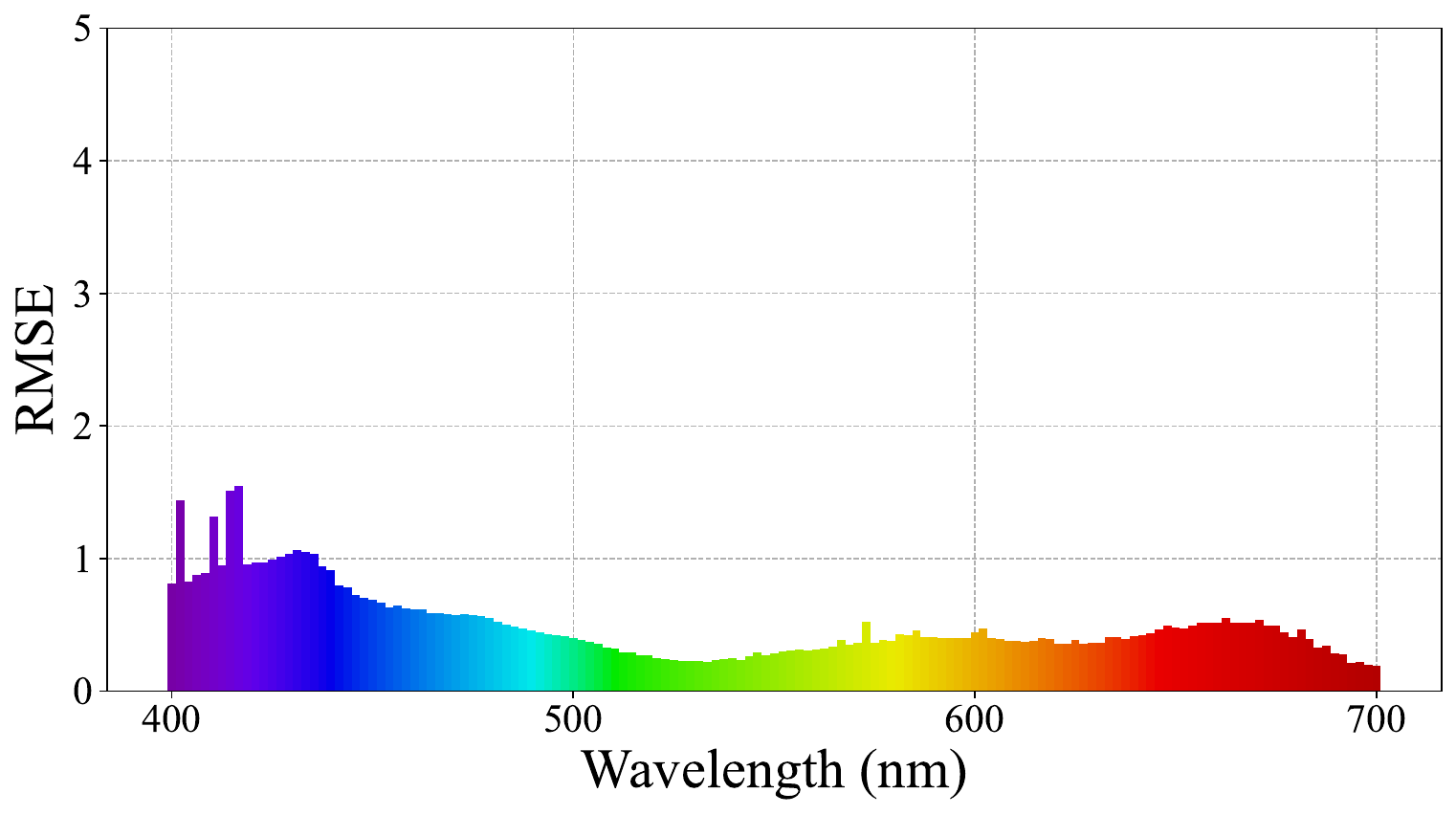}
		\label{fig:metrics:PACE:4}
	}
	\subfigure[Log-Bias on MDN ]{
		\includegraphics[width=0.3\linewidth]{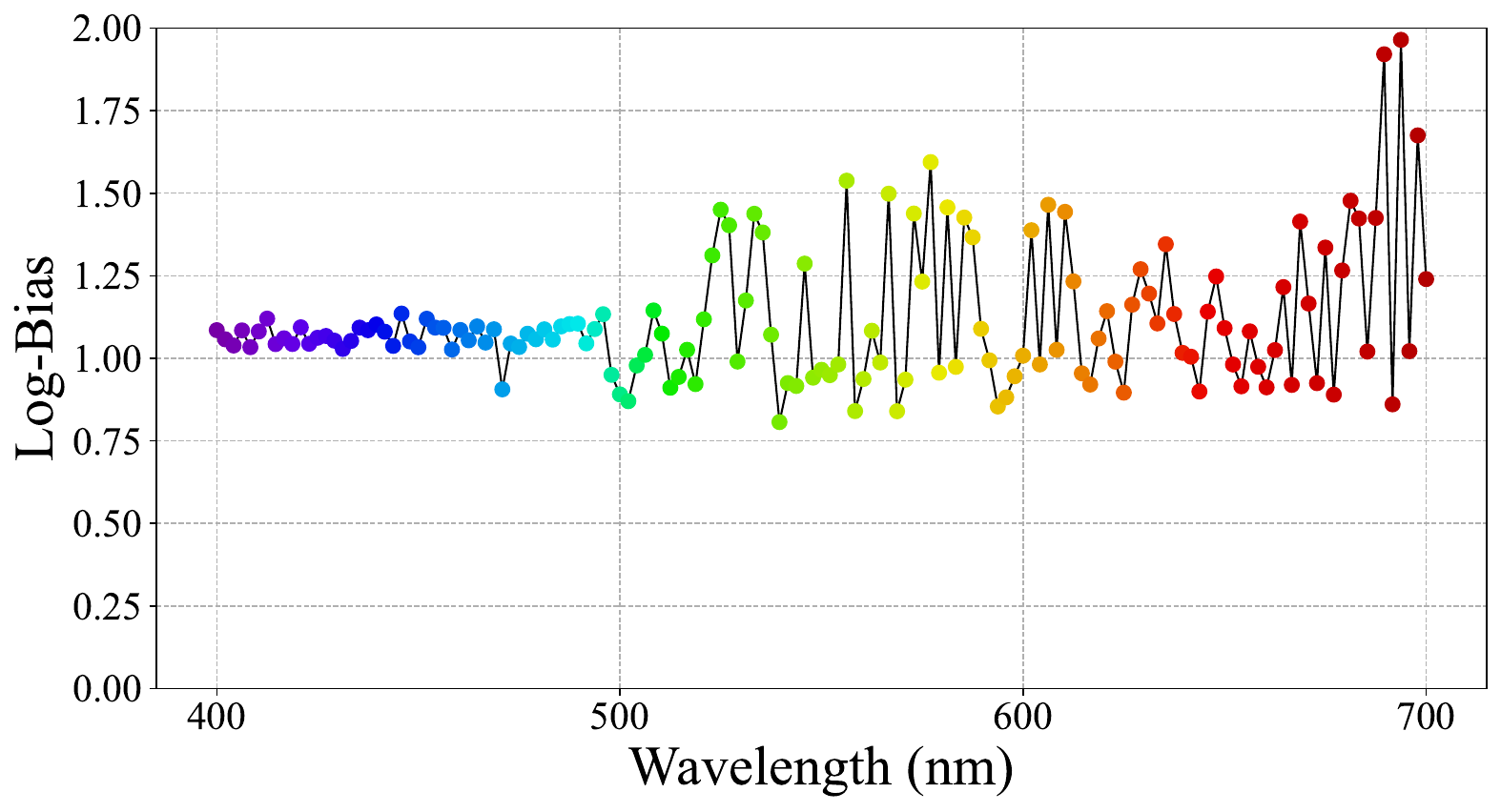}
		\label{fig:metrics:PACE:5}
	}	
	\subfigure[$\beta$ on MDN]{
		\includegraphics[width=0.3\linewidth]{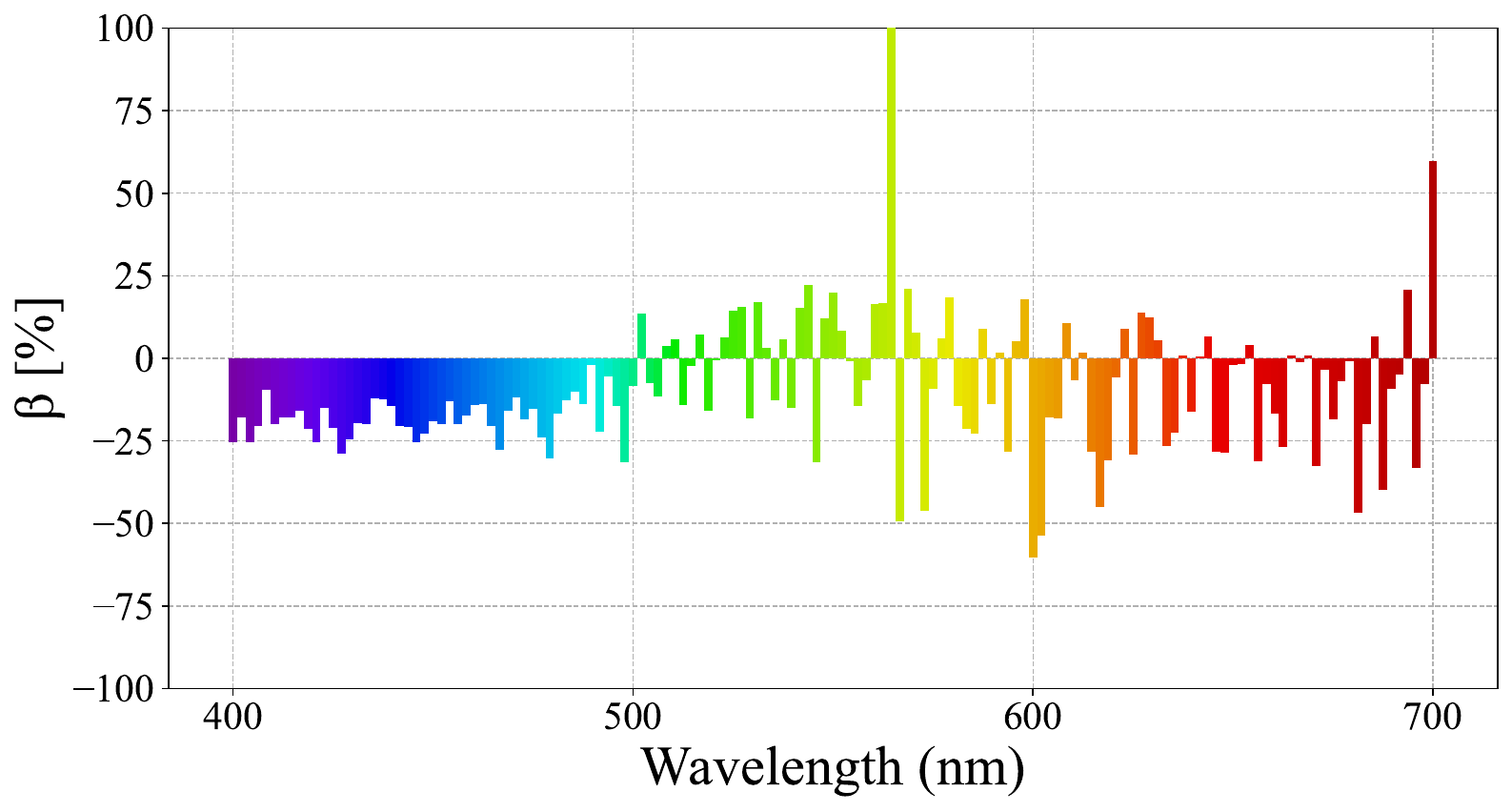}
		\label{fig:metrics:PACE:6}
	}
	\caption{Performance of VAE  in terms of RMSE, Log-Bias, and $\beta$ for $\aphy$ prediction on PACE wavelengths across 400nm-700nm.}
	\label{fig:metrics:PACE}
\end{figure*}

\begin{figure*}%[ht]
	\centering
	\subfigure[RMSE on VAE ]{
		\includegraphics[width=0.3\linewidth]{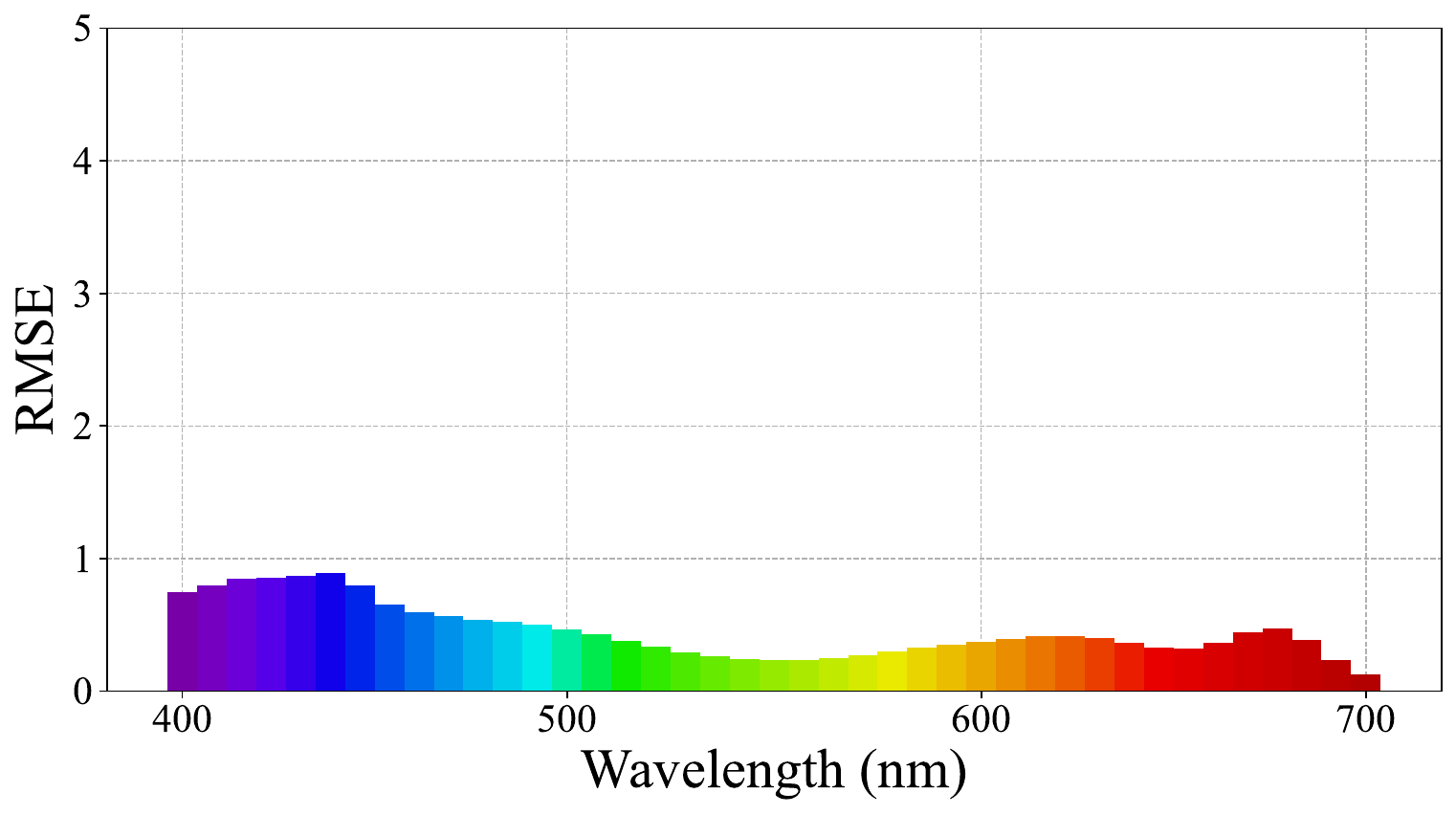}
		\label{fig:metrics:EMIT:1}
	}	
	\subfigure[Log-Bias on VAE]{
		\includegraphics[width=0.3\linewidth]{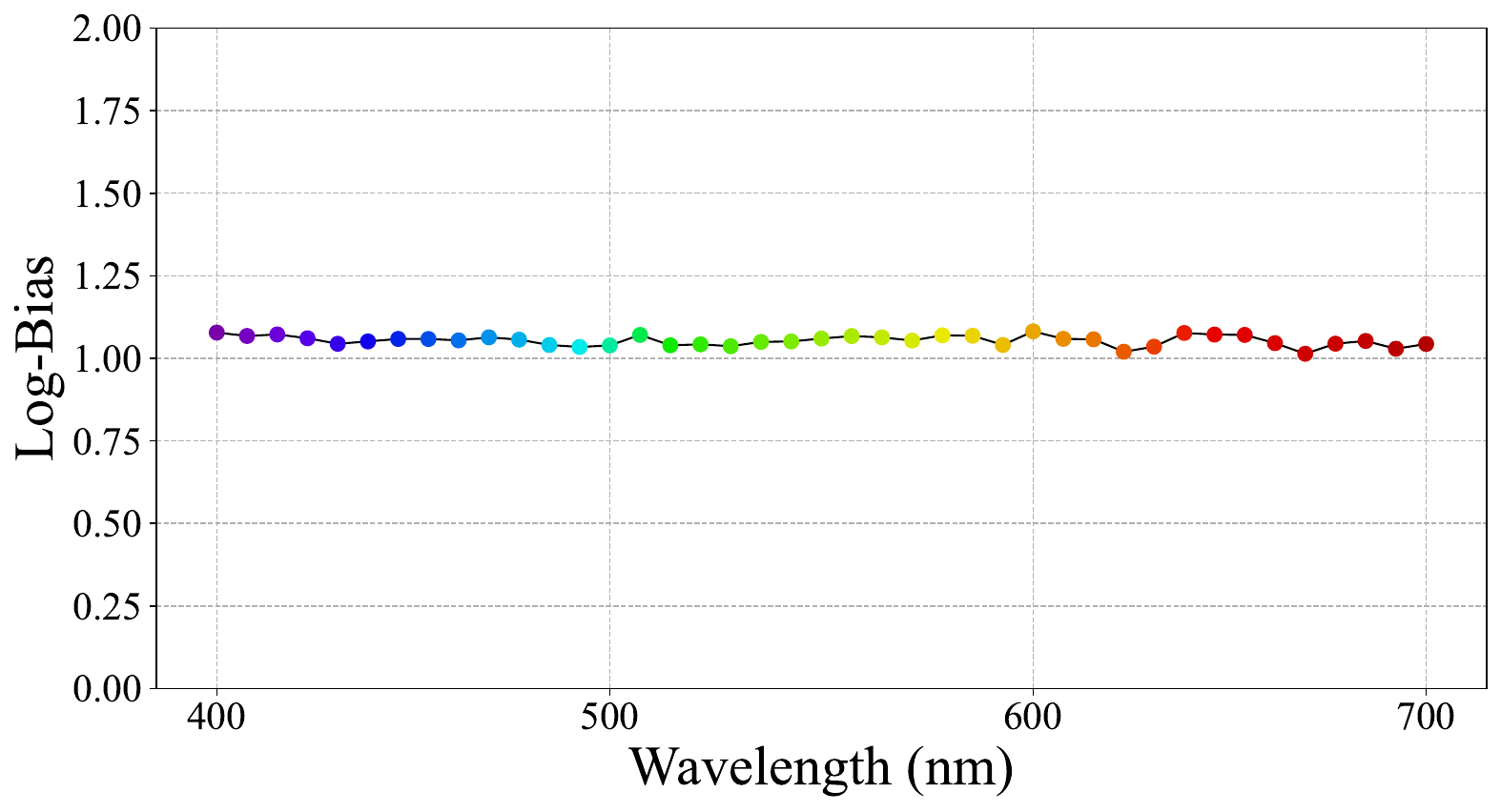}
		\label{fig:metrics:EMIT:2}
	}
	\subfigure[$\beta$ on VAE]{
		\includegraphics[width=0.3\linewidth]{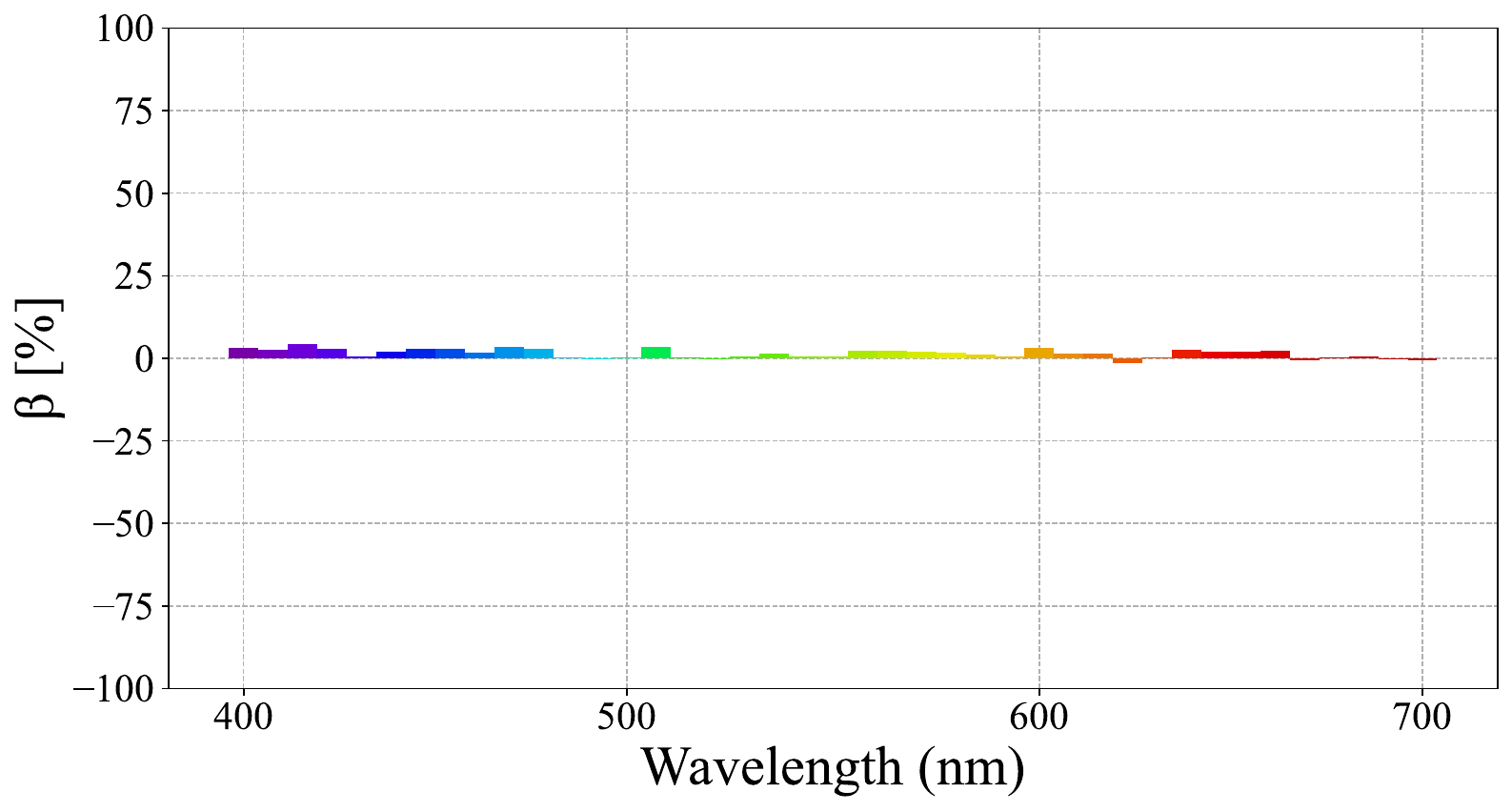}
		\label{fig:metrics:EMIT:3}
	}	
	\subfigure[RMSE on MDN ]{
		\includegraphics[width=0.3\linewidth]{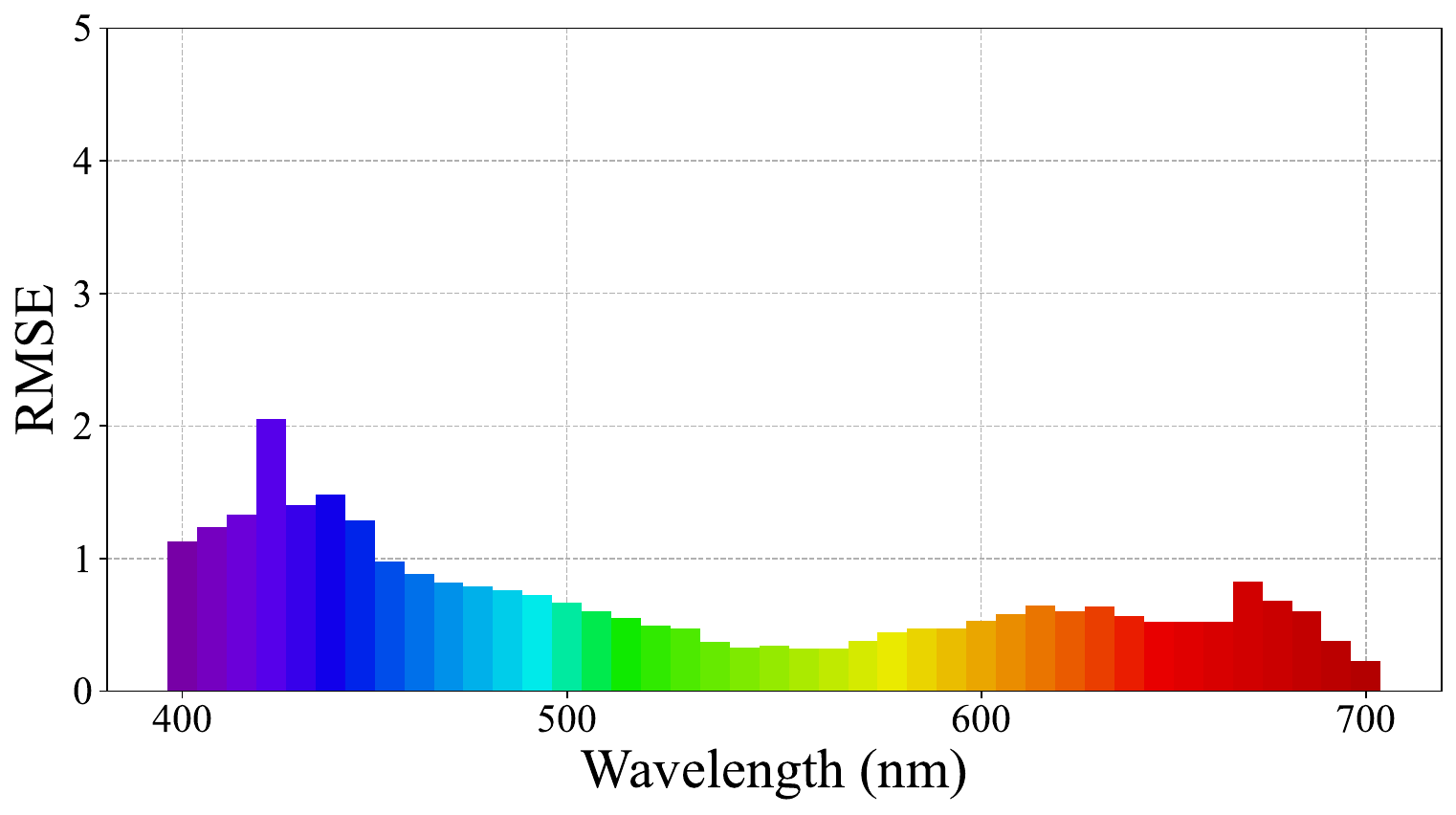}
		\label{fig:metrics:EMIT:4}
	}
	\subfigure[Log-Bias on MDN ]{
		\includegraphics[width=0.3\linewidth]{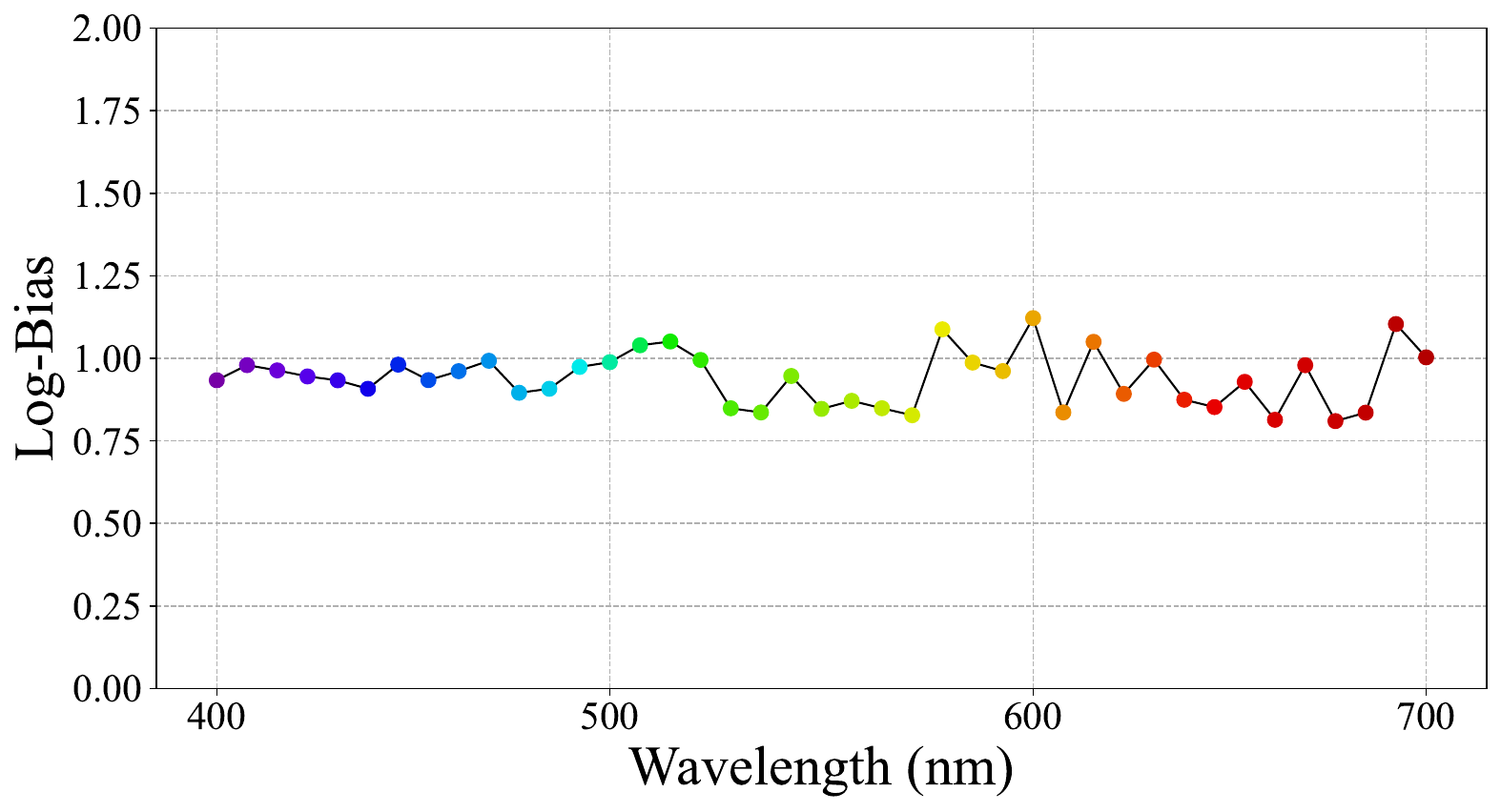}
		\label{fig:metrics:EMIT:5}
	}	
	\subfigure[$\beta$ on MDN]{
		\includegraphics[width=0.3\linewidth]{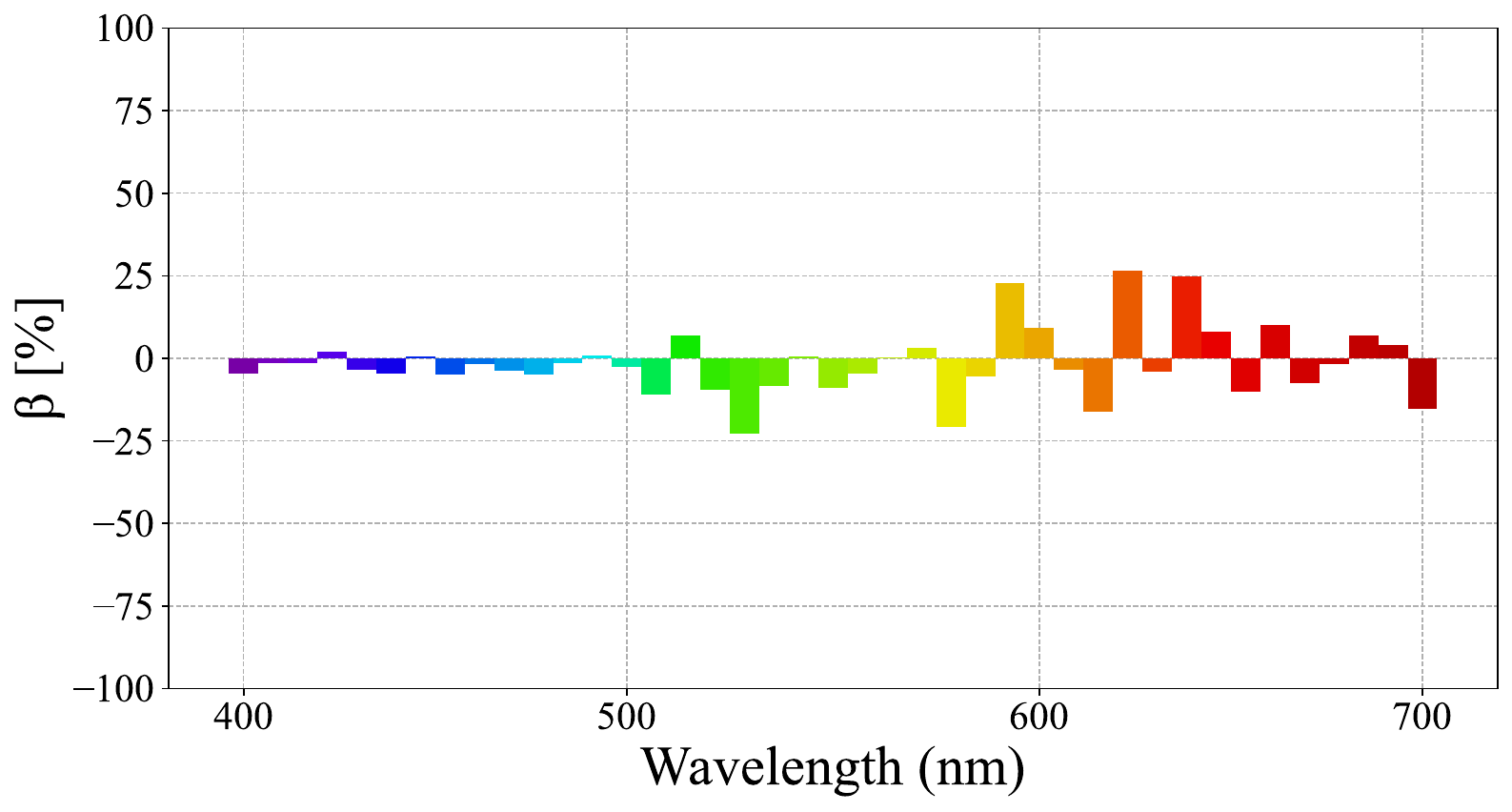}
		\label{fig:metrics:EMIT:6}
	}
  \vspace{-0.5em}
	\caption{Performance of VAE  in terms of RMSE, Log-Bias, and $\beta$ for $\aphy$ prediction with EMIT spectral setting across 400nm-700nm.}
 \vspace{-0.5em}
	\label{fig:metrics:EMIT}
\end{figure*}

To comprehensively assess the prediction performance across the 400-700 nm wavelength range, we compared three evaluation metrics, including RMSE, Log-Bias, and $\beta$,  between VAE and MDN models across all PACE and EMIT wavelengths, with results showing in Figure~\ref{fig:metrics:PACE} and~Figure~\ref{fig:metrics:EMIT}. 
Examining RMSE trends for PACE, which are shown in Figs.~\ref{fig:metrics:PACE:1} and~\ref{fig:metrics:PACE:4}, both VAE and MDN models show comparable performance across most wavelengths. 
While MDN exhibits slightly higher RMSE values in the shorter wavelength range (400–500 nm), within the 600–700 nm range, RMSE values for both models remain stable. 
In contrast, for EMIT, VAE maintains an RMSE under 1.0, whereas MDN peaks at values slightly above 2.0 at shorter wavelengths, indicating higher prediction errors for MDN in these regions. 
At longer wavelengths (600–700 nm), VAE maintains a smooth trend, with RMSE values stabilizing below 0.5, whereas MDN exhibits mild fluctuations, remaining below 0.8.
In terms of Log-Bias, VAE-$\aphy$ also highlights the stability of the model for both PACE  and EMIT.
For instance, the VAE model maintains Log-Bias values close to the ideal value of 1 across the entire wavelength range (400–700 nm), demonstrating its ability to produce unbiased estimates. 
In contrast, the MDN model exhibits certain fluctuations, particularly at longer wavelengths (600–700 nm), where deviations from 1 become increasingly pronounced. 
Lastly, the VAE model exhibits more stable and lower $\beta$ values compared to MDN across both PACE and EMIT.
For VAE, $\beta$ fluctuations remain within $\pm$ 10\% across the spectrum for both sensors, demonstrating its robust predictive capability. 
In contrast, MDN experiences substantial variations, exceeding $\pm$ 50\% at several wavelengths, with a notable increase in deviation in longer wavelengths (600–700 nm). 
When comparing PACE and EMIT, MDN exhibits larger $\beta$ deviations for PACE, particularly at wavelengths beyond 600 nm.
This suggests that MDN struggles more with PACE’s spectral characteristics, likely due to its higher spectral resolution, leading to greater instability in its predictions. 
Additionally, while EMIT shows moderate fluctuations in MDN’s $\beta$ values at shorter wavelengths (~400–500 nm), PACE exhibits more pronounced variations across the spectrum, reinforcing the idea that MDN is less stable for PACE than EMIT. 
Despite these differences, VAE maintains consistently low fluctuations across both datasets, which is crucial in hyperspectral applications where each wavelength provides critical input to construct the spectra.
Collectively, these results demonstrate that VAE is a suitable framework for PACE and EMIT, as it offers a more stable, precise, and unbiased estimation of $\aphy$ values across their spectral settings. 
Accuracy across wavelengths is crucial because detailed and precise $\aphy$ spectral predictions are key to accurately determining phytoplankton community composition (PCC).
The ability of VAE to maintain consistent Log-Bias and low $\beta$ variations also underscores its potential for future hyperspectral remote sensing missions, such as GLIMR, and SBG, where VAE's spectral settings can be easily tuned and adapted to meet mission-specifc configurations.

\begin{figure*}%[ht]
	\centering
	\subfigure[VAE on PACE]{
		\includegraphics[width=0.3\linewidth]{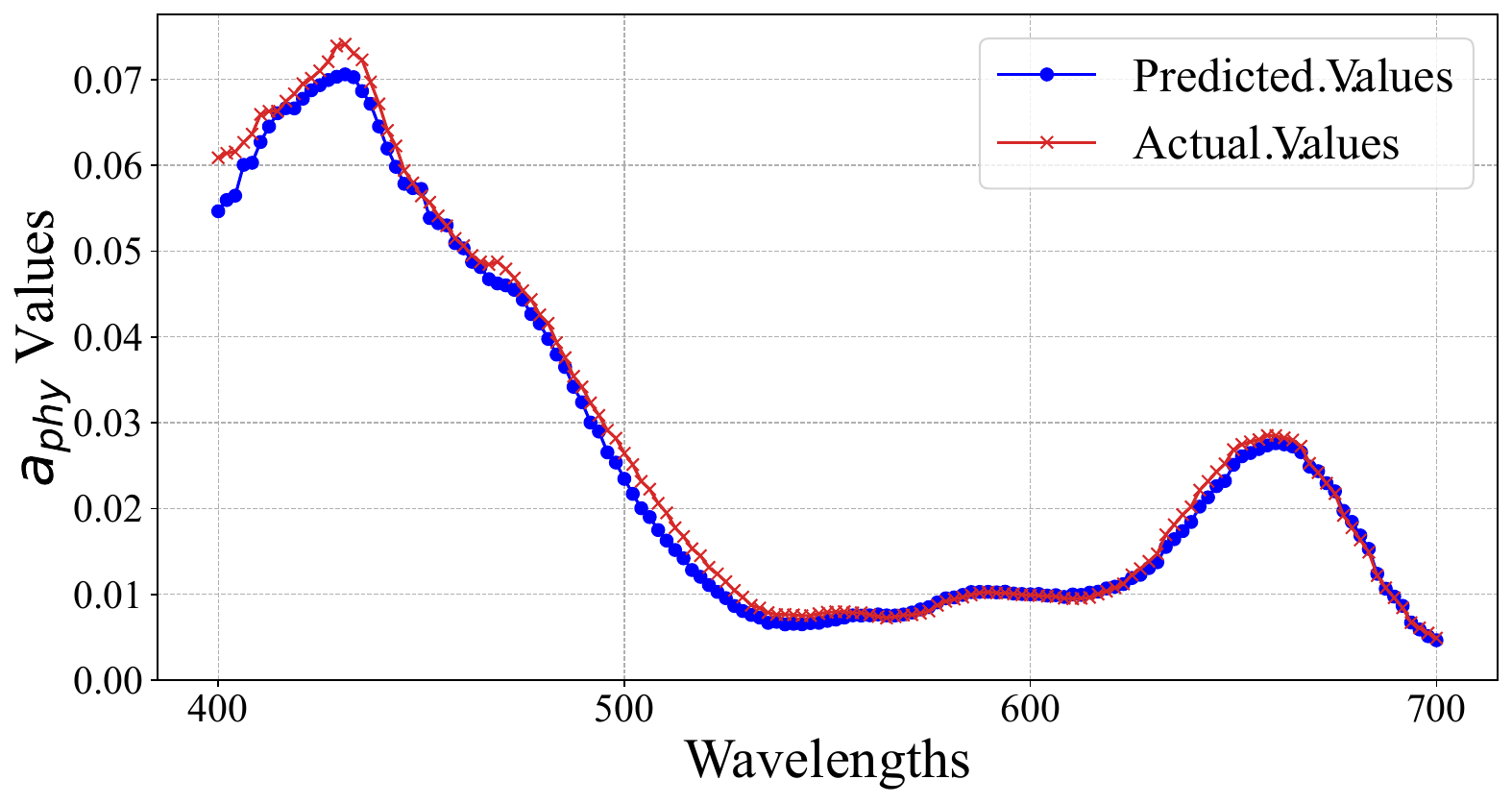}
		\label{PACE_line}
	}	
	\subfigure[MDN on PACE]{
		\includegraphics[width=0.3\linewidth]{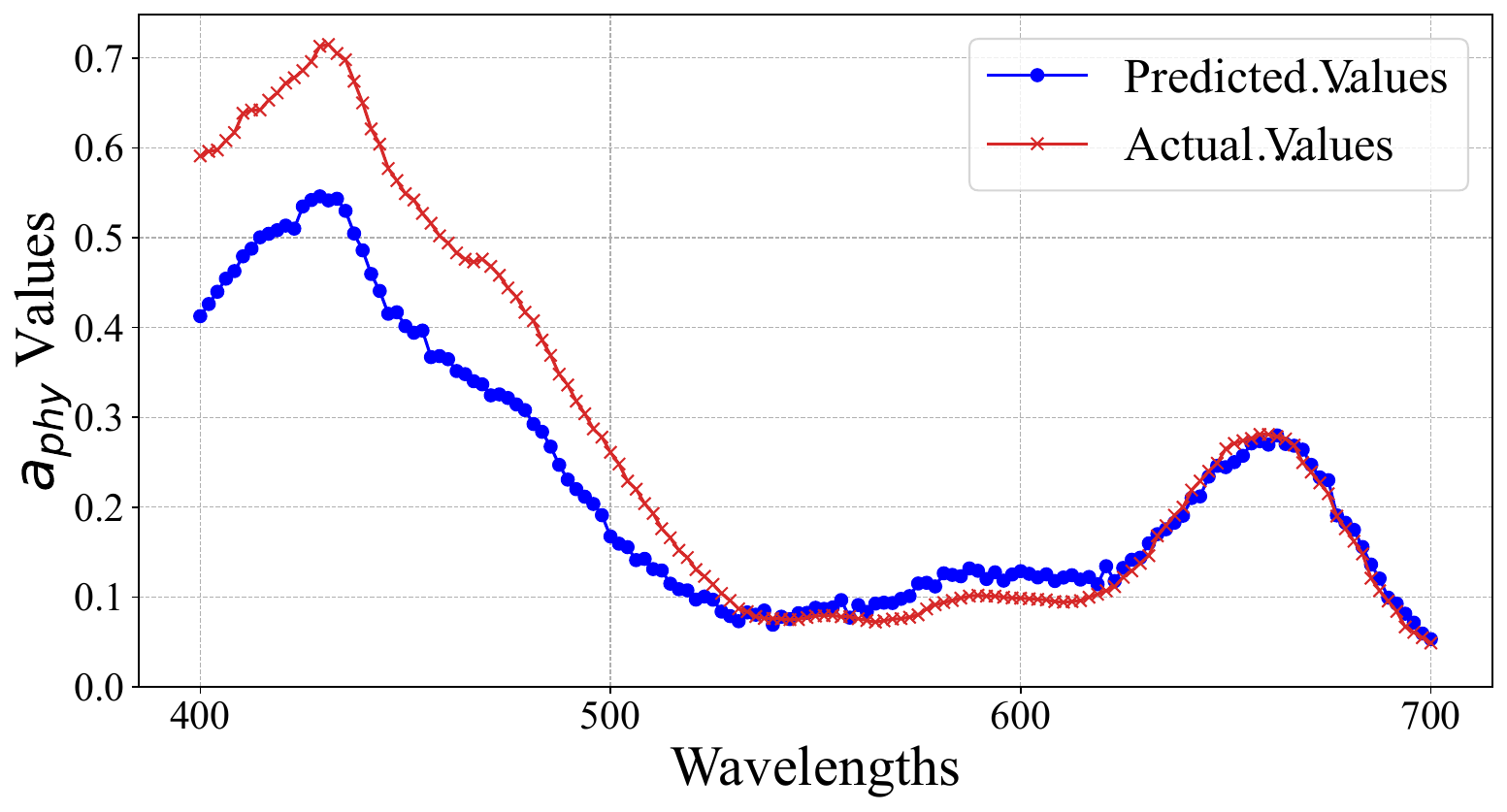}
		\label{MDN_PACE_line}
	}
	\subfigure[M-MDN on PACE]{
		\includegraphics[width=0.3\linewidth]{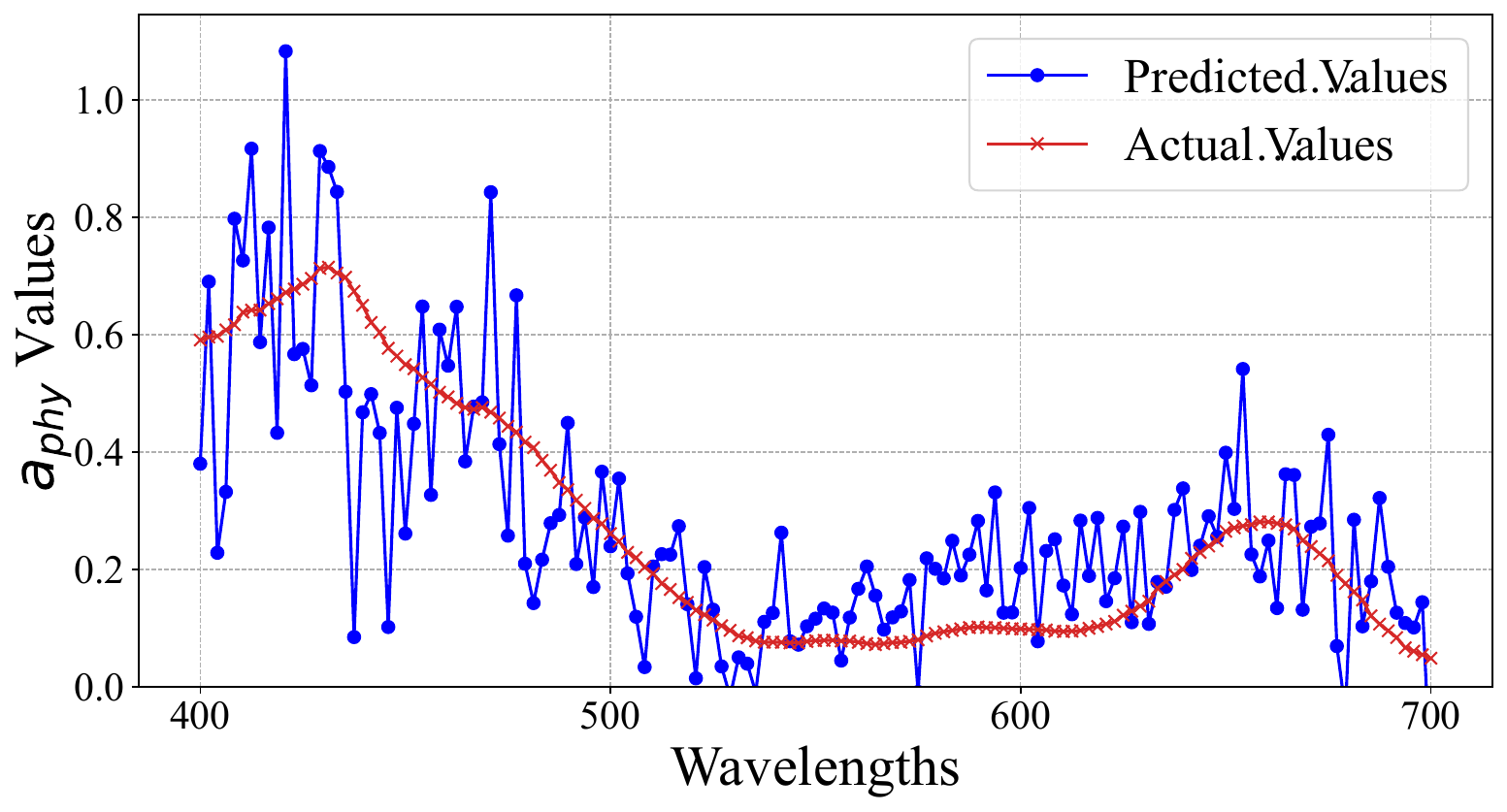}
		\label{MMDN_PACE_line}
	}	
	\subfigure[VAE on EMIT]{
		\includegraphics[width=0.3\linewidth]{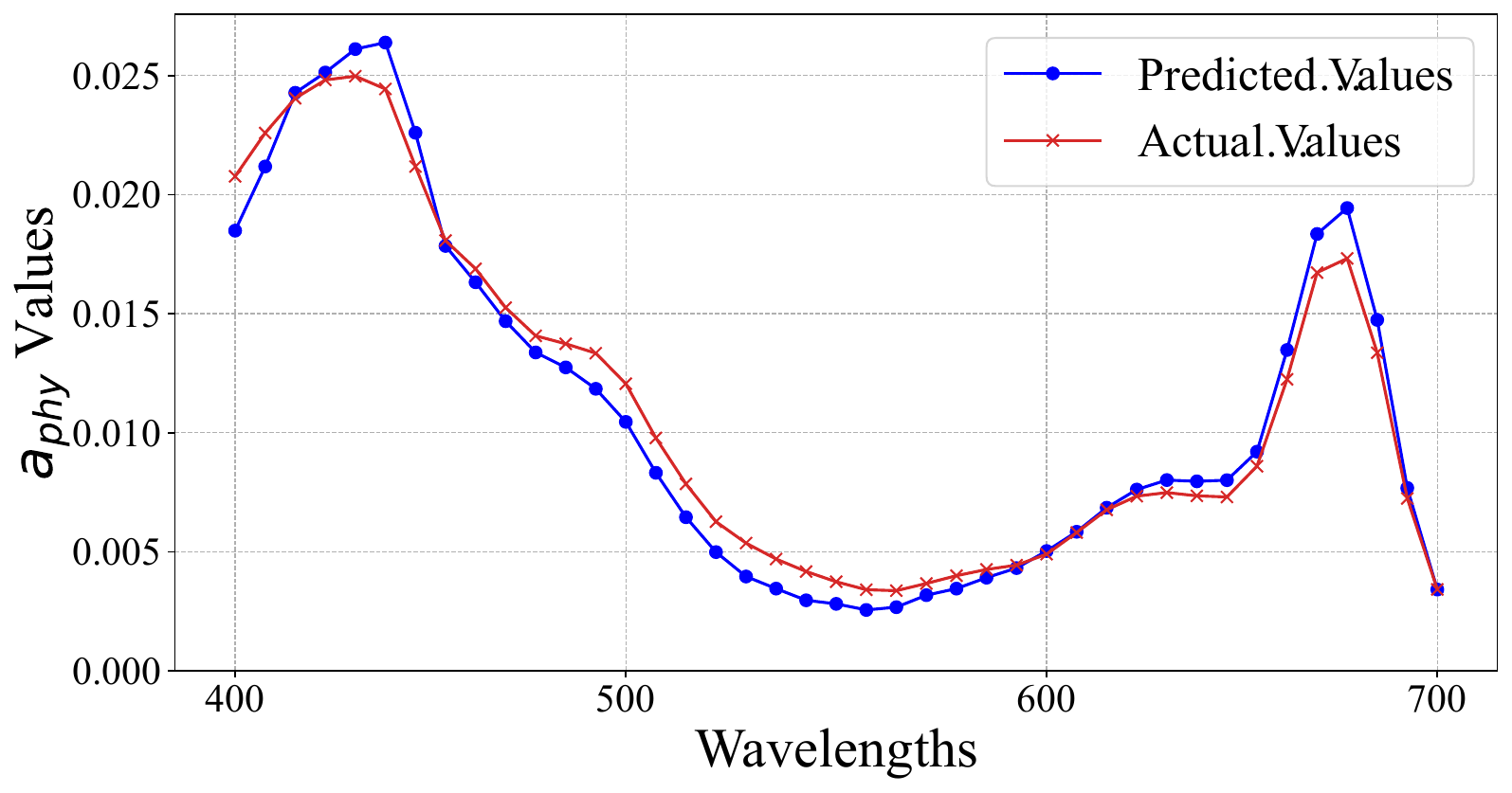}
		\label{EMIT_line}
	}
	\subfigure[MDN on EMIT]{
		\includegraphics[width=0.3\linewidth]{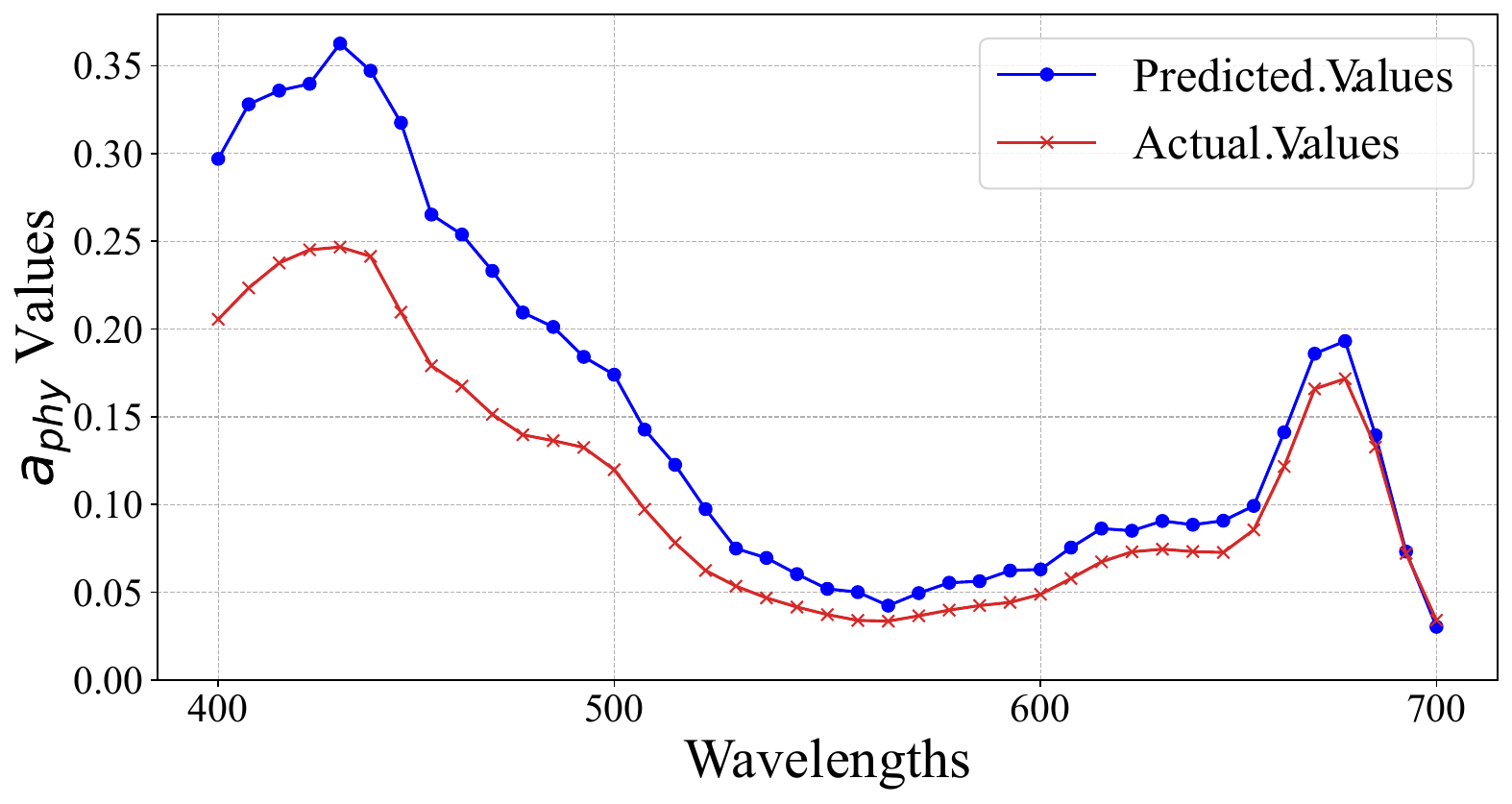}
		\label{MDN_EMIT_line}
	}	
	\subfigure[M-MDN on EMIT]{
		\includegraphics[width=0.3\linewidth]{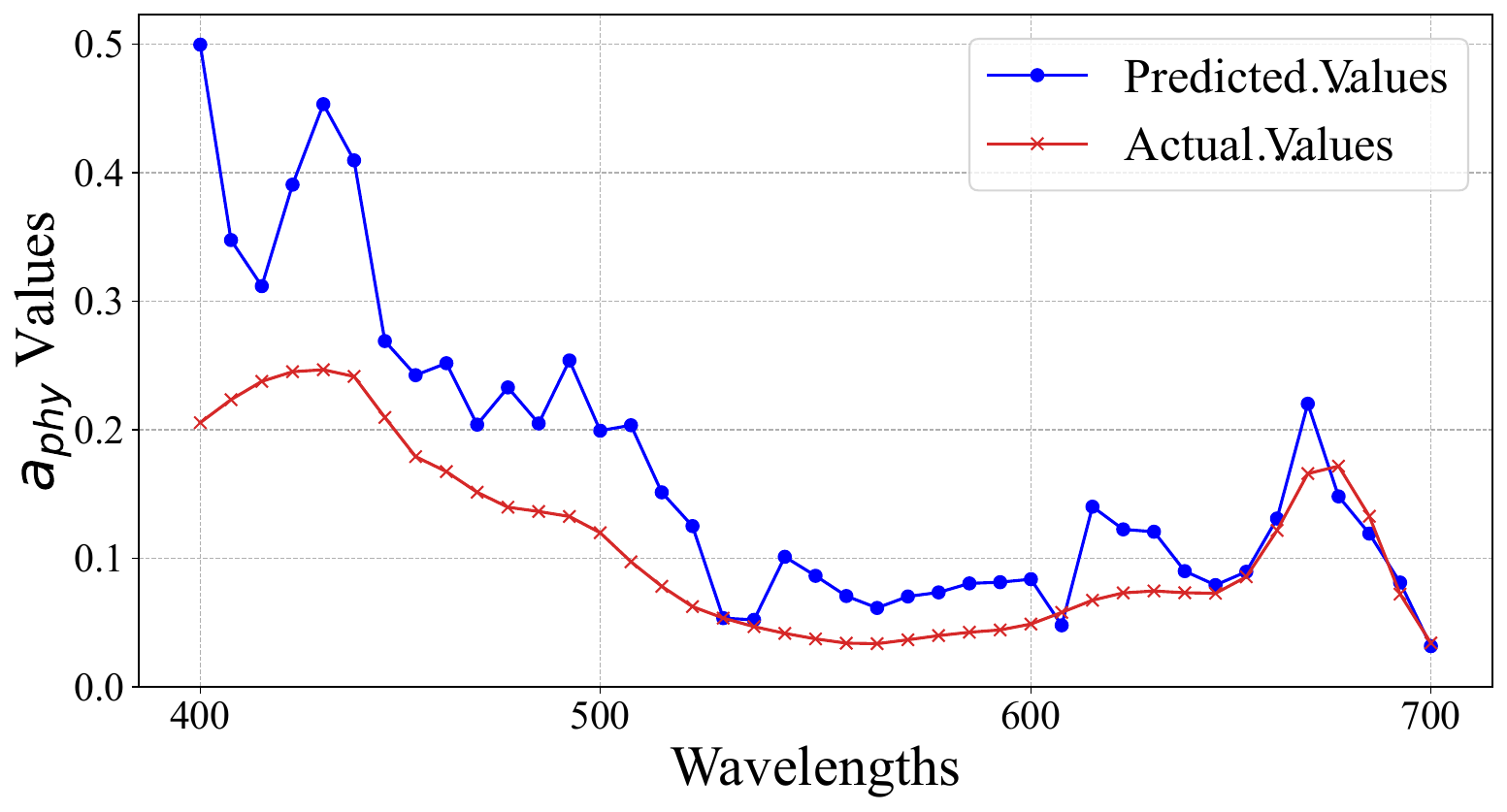}
		\label{MMDN_EMIT_line}
	}
	\caption{Comparison of actual and predicted $\aphy$ spectra in the 400–700 nm range: (a)–(c) show results for VAE, MDN, and M-MDN using PACE spectral setting, while (d)–(f) present corresponding results using EMIT wavelengths.}
	\label{fig:fit}
\end{figure*}

{
Moreover, both VAE and MDN models are also applied to the same randomly selected testing Rrs-$\aphy$ spectra, resampled at PACE and EMIT wavelength settings, to compare the predicted and actual $\aphy$ spectra. 
We also include a comparison with M-MDN for $\aphy$ predictions, which samples $\aphy$ from the learned mixture of Gaussian distributions instead of selecting the Gaussian mean with the highest weight as the output to handle one-to-many estimation problems.
Fig.~\ref{fig:fit} presents the actual and predicted spectra within the 400-700 nm range with Figures~\ref{fig:fit}(a)-(c) and Figures~\ref{fig:fit}(d)-(f) at PACE and EMIT spectral settings, respectively. }

The results demonstrate that the VAE-$\aphy$ (Figures~\ref{fig:fit}(a) and (d)) provide superior predictions, as the estimated  $\aphy$ curves (blue) closely align with the actual values (red) across the entire spectral range. 
In contrast, while the MDN model (Figures~\ref{fig:fit}(b) and (e)) preserves general spectral shape, it exhibits slight deviations between the predicted and actual spectra, which is most evident at shorter wavelengths in the blue range (400-500 nm). 
The M-MDN model (Figures~\ref{fig:fit}(c) and (f)) exhibits the highest level of instability, with substantial fluctuations and increased prediction noise across all wavelengths, particularly where high-frequency noise is more pronounced. 
These results suggest that M-MDN struggles to generalize effectively, introducing large uncertainties in $\aphy$ predictions when sampling $\aphy$ from the learned mixture of Gaussian distributions to address one-to-many issues.
The deterioration in $\aphy$ predictions from M-MDN is particularly pronounced at PACE wavelengths, which have a higher spectral resolution (2.5 nm) and consequently higher input and output dimensions.
M-MDN struggles with high-dimensional data, whereas VAE-$\aphy$ effectively maps high-dimensional data into a lower-dimensional latent variable space, leading to a better performance on PACE wavelengths.
This result underscores the reliability and accuracy of VAE-$\aphy$ model across varying wavelength conditions, highlighting its novelty in the applications of high-dimensional data predictions. 
%These findings highlight a fundamental limitation of MDN-based methods, such as those proposed in~\cite{o2023hyperspectral,pahlevan2021hyperspectral}, which, despite producing reasonable estimations, fail to truly resolve the one-to-many problem and cannot be easily improved through simple modifications.

% \begin{figure*}[ht]
% 	\centering
% 	\subfigure[RMSE on VAE ]{
% 		\includegraphics[width=0.3\linewidth]{figs/HICO_RMSE.pdf}
% 		\label{Fig:GAN-0-AIDS}
% 	}	
% 	\subfigure[Bias on VAE]{
% 		\includegraphics[width=0.3\linewidth]{figs/HICO_Bias.pdf}
% 		\label{Fig:ALS-0-AIDS}
% 	}
% 	\subfigure[$\beta$ on VAE]{
% 		\includegraphics[width=0.3\linewidth]{figs/HICO_Beta.pdf}
% 		\label{Fig:GAN-1-AIDS}
% 	}	
% 	\subfigure[RMSE on MDN ]{
% 		\includegraphics[width=0.3\linewidth]{figs/MDN_HICO_RMSE.pdf}
% 		\label{Fig:ALS-1-AIDS}
% 	}
% 	\subfigure[Bias on MDN ]{
% 		\includegraphics[width=0.3\linewidth]{figs/MDN_HICO_Bias.pdf}
% 		\label{Fig:GAN-6-AIDS}
% 	}	
% 	\subfigure[$\beta$ on MDN]{
% 		\includegraphics[width=0.3\linewidth]{figs/MDN_HICO_Beta.pdf}
% 		\label{Fig:ALS-6-AIDS}
% 	}
%   \vspace{-1em}
% 	\caption{Performance comparisons in terms of RMSE, Bias, and $\beta$ for $\aphy$ prediction on HICO across 400nm-700nm.}
%  \vspace{-1em}
% 	\label{fig:metrics:HICO}
% \end{figure*}

\subsection{Evaluations of Chl$\mathit{a}$ Predictions}

\begin{figure*}%[ht]
	\centering
	\subfigure[VAE on PACE]{
		\includegraphics[width=0.23\linewidth]{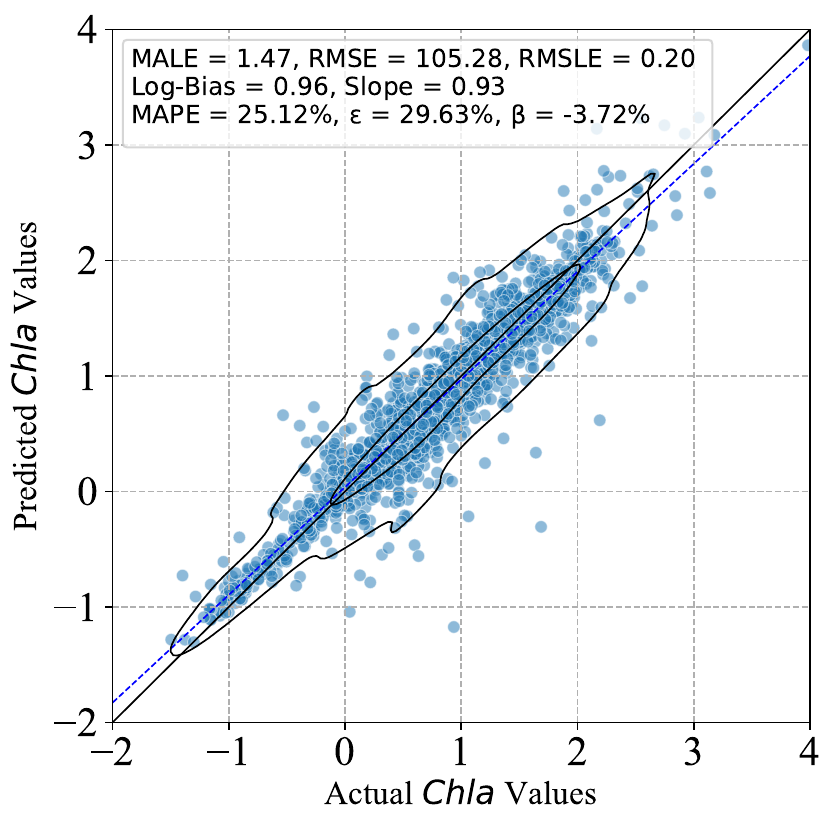}
		\label{fig:chl2}
	}
    	\subfigure[MDN on PACE]{
		\includegraphics[width=0.23\linewidth]{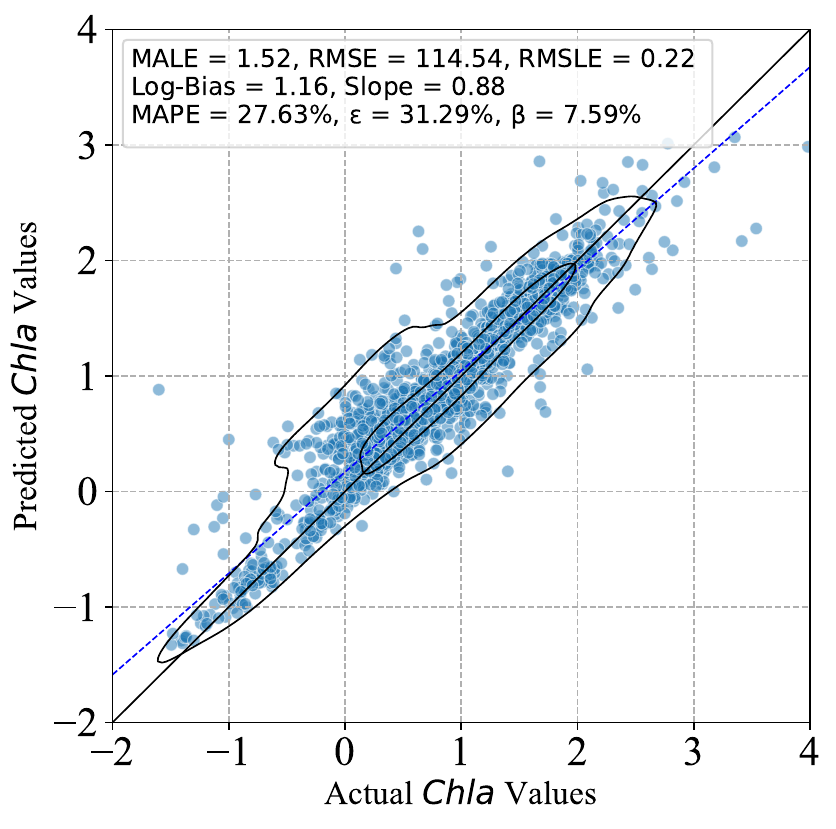}
		\label{fig:chl5}
	}	
	\subfigure[VAE on EMIT]{
		\includegraphics[width=0.23\linewidth]{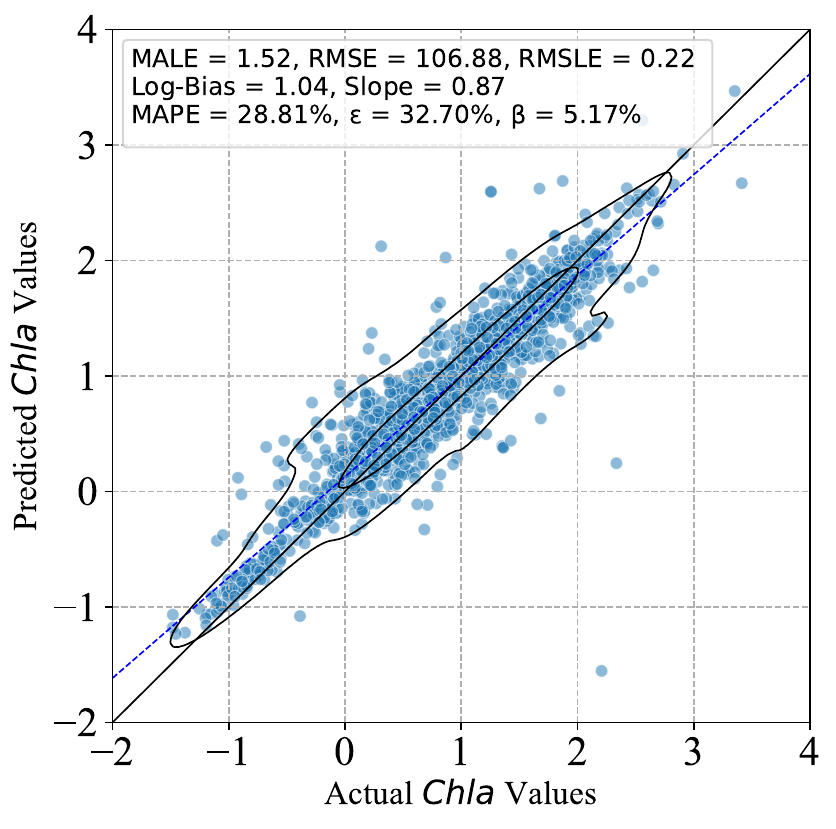}
		\label{fig:chl3}
	}	
	\subfigure[MDN on EMIT]{
		\includegraphics[width=0.23\linewidth]{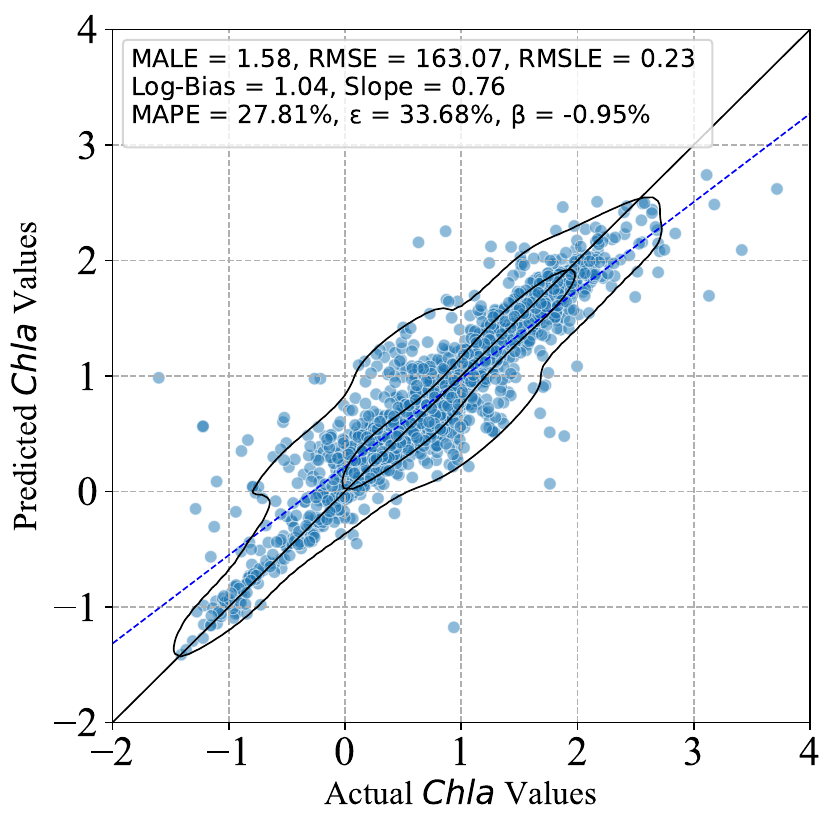}
		\label{fig:chl6}
	}
	\caption{Evaluation metrics for VAE and MDN predictions of Chl-a
using PACE and EMIT spectral settings. (a), (c) VAE for PACE and EMIT, and (b), (d) MDN for PACE and EMIT}
	\label{fig:chl}
\end{figure*}

In this section, we present the performance of VAE-Chl-a and MDN in predicting Chl-a using the $\Rrs$ PACE and EMIT spectral settings, respectively, in Fig.~\ref{fig:chl}. 
It is observed that the VAE-Chl-a model slightly outperforms the MDN model across all evaluation metrics. %(Fig.~\ref{fig:chl}). 
First, VAE-Chl-a model exhibits lower prediction errors, as evidenced by overall lower values in MALE, RMSE, and RMSLE.
For example, as shown in Figs~\ref{fig:chl2}, which are based on $\Rrs$ at PACE wavelengths, the VAE-Chl-a model shows an MALE of 1.47, an RMSE of 105.28, and an RMSLE of 0.20, while the MDN model yields an MALE of 1.52, an RMSE of 114.54, and an RMSLE of 0.22. 
Additionally, the prediction deviation of VAE-Chl-a model is better than that of the MDN model, as rendering Log-bias and slope values are closer to 1. 
However, unlike the substantial advantage of VAE for $\aphy$ predictions, VAE-Chl-a model achieves only minor improvements over the MDN model. 
This is because the prediction for Chl-a is a single-value prediction whereas $\aphy$ predictions involve high-dimensional spectral vectors. 
Consequently, the advantage of the VAE in handling high-dimensional prediction is not as evident in this VAE-Chl-a case. 
Despite this, the slight improvement still demonstrates the robustness of the VAE model as an ideal framework for adoption in the ocean color field for the inversion retrievals of IOPs and optical active biogeochemical parameters. 
%{
%Moreover, our VAE-Chl-a model can achieve the one-to-many Chl-a while MDN methods fail.

\begin{figure*}[]
	\centering
	\subfigure[VAE prediction on PACE example 1]{
		\includegraphics[width=0.3\linewidth]{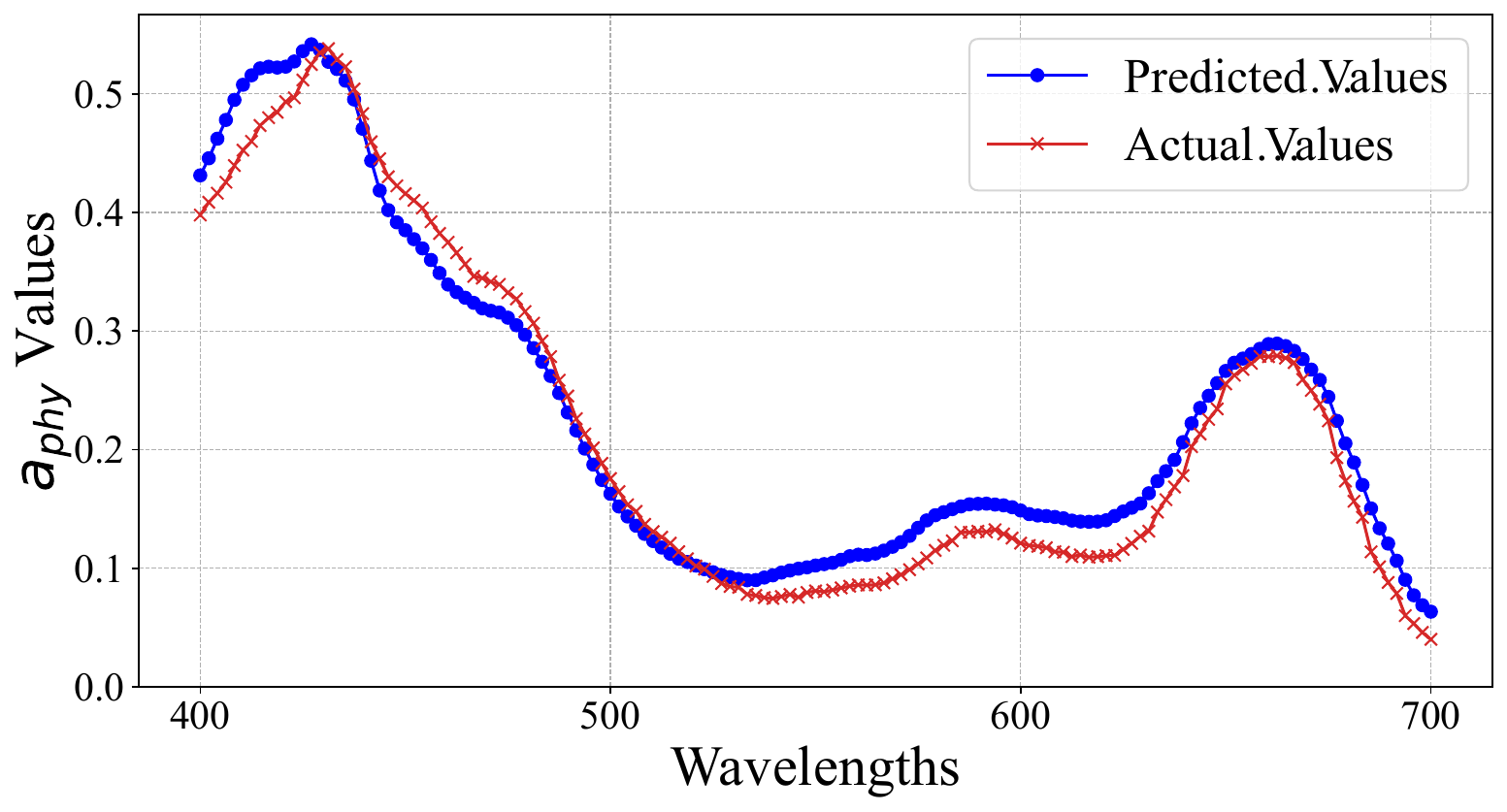}
		\label{fig:fit:real:1}
	}	
  \subfigure[VAE prediction on PACE example 2]{
		\includegraphics[width=0.3\linewidth]{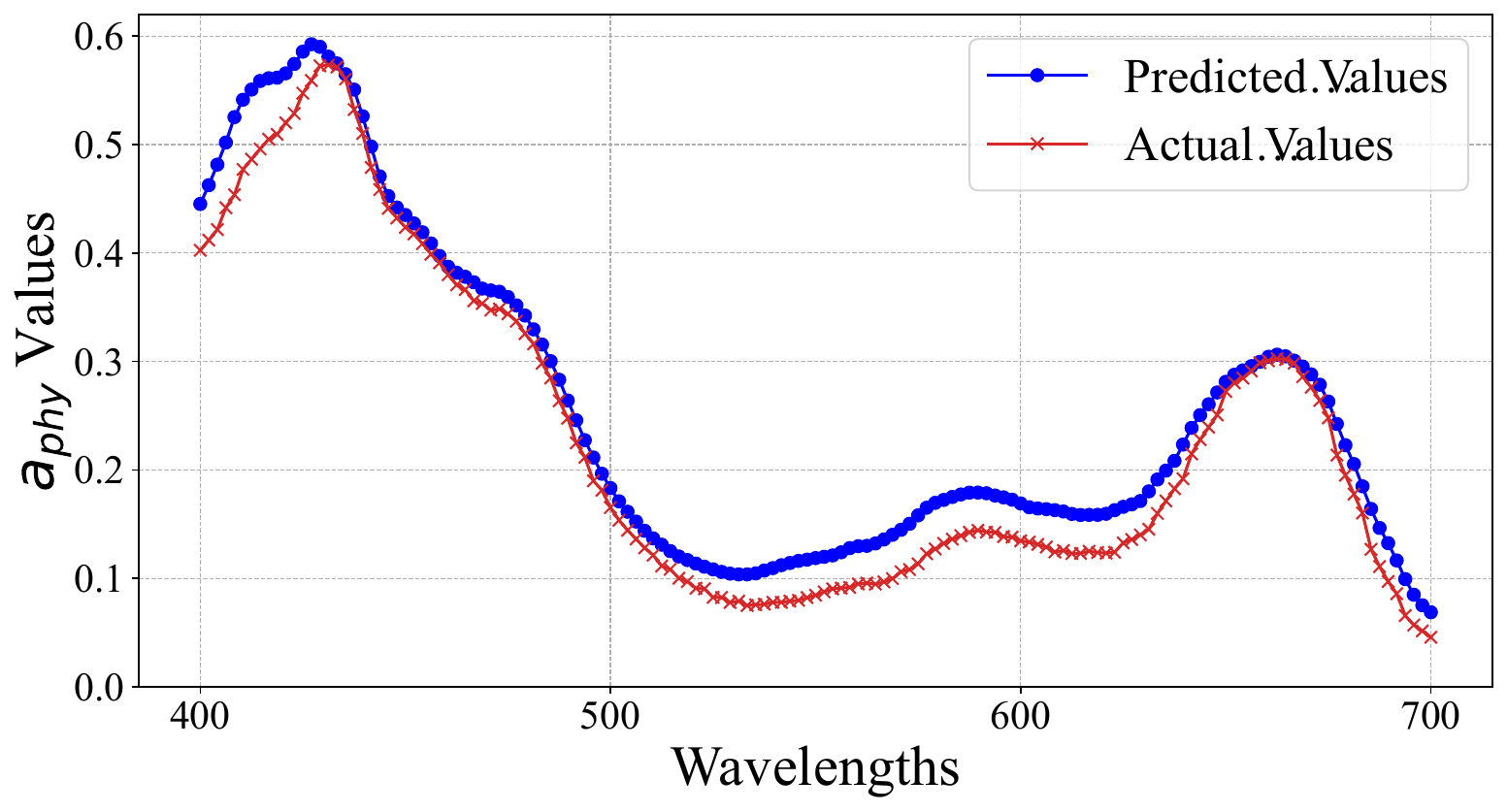}
		\label{fig:fit:real:2}
	}
  \subfigure[VAE prediction on PACE example 3]{
		\includegraphics[width=0.3\linewidth]{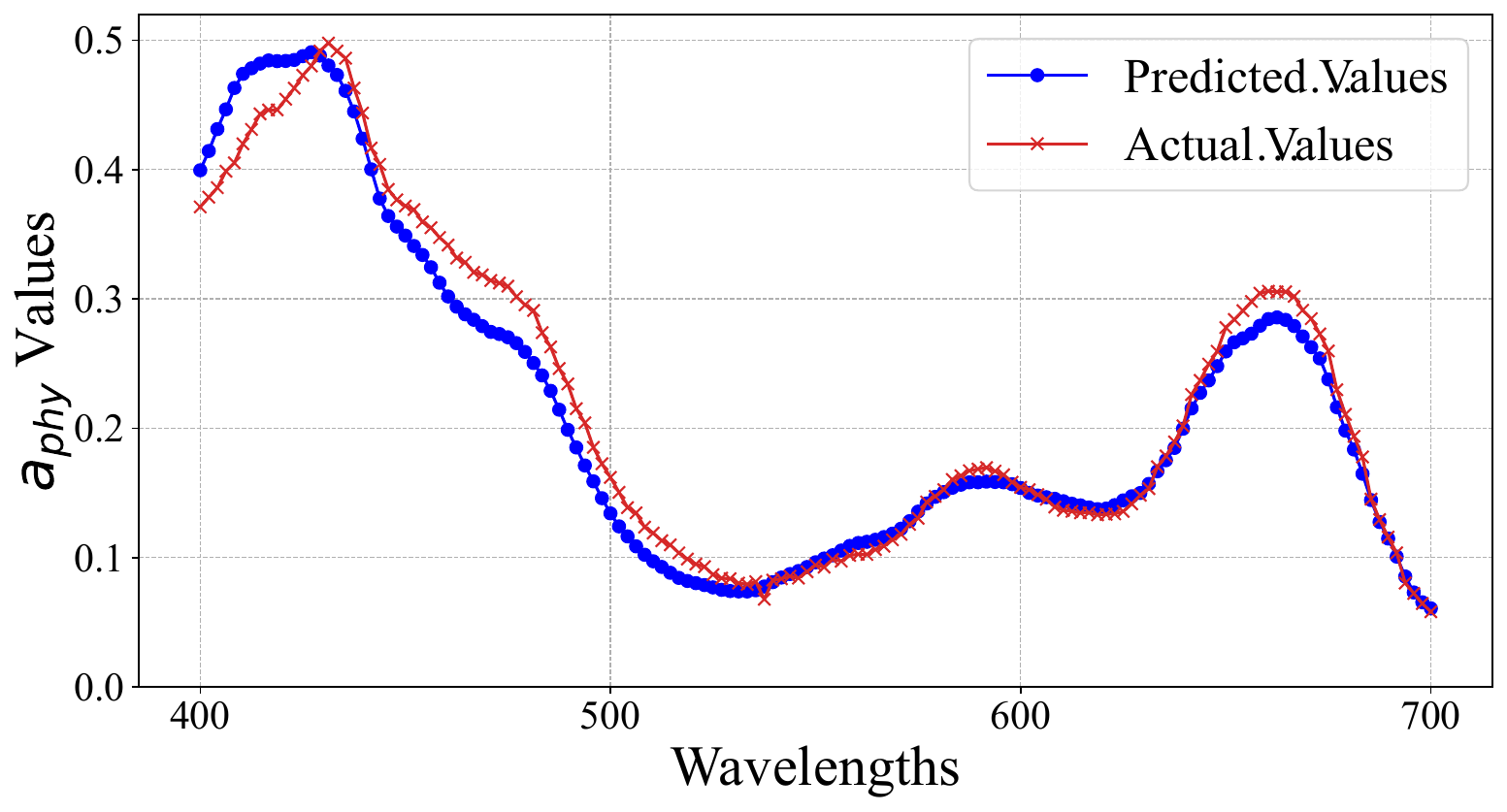}
		\label{fig:fit:real:3}
	}	
	% \subfigure[MDN for Chl-a levels at 5 $\mu$g $L^(-1)$]{
	% 	\includegraphics[width=0.3\linewidth]{figs/MDN_PACE_5chla.pdf}
	% 	\label{fig:fit:real:4}
	% }
	% \subfigure[MDN for Chl-a levels at 30 $\mu$g $L^(-1)$]{
	% 	\includegraphics[width=0.3\linewidth]{figs/MDN_PACE_30chla.pdf}
	% 	\label{fig:fit:real:5}
	% }
	% \subfigure[MDN for Chl-a levels at 50 $\mu$g $L^(-1)$]{
	% 	\includegraphics[width=0.3\linewidth]{figs/MDN_PACE_50chla.pdf}
	% 	\label{fig:fit:real:6}
 %        }
        \subfigure[M-MDN prediction on PACE example 1]{
		\includegraphics[width=0.3\linewidth]{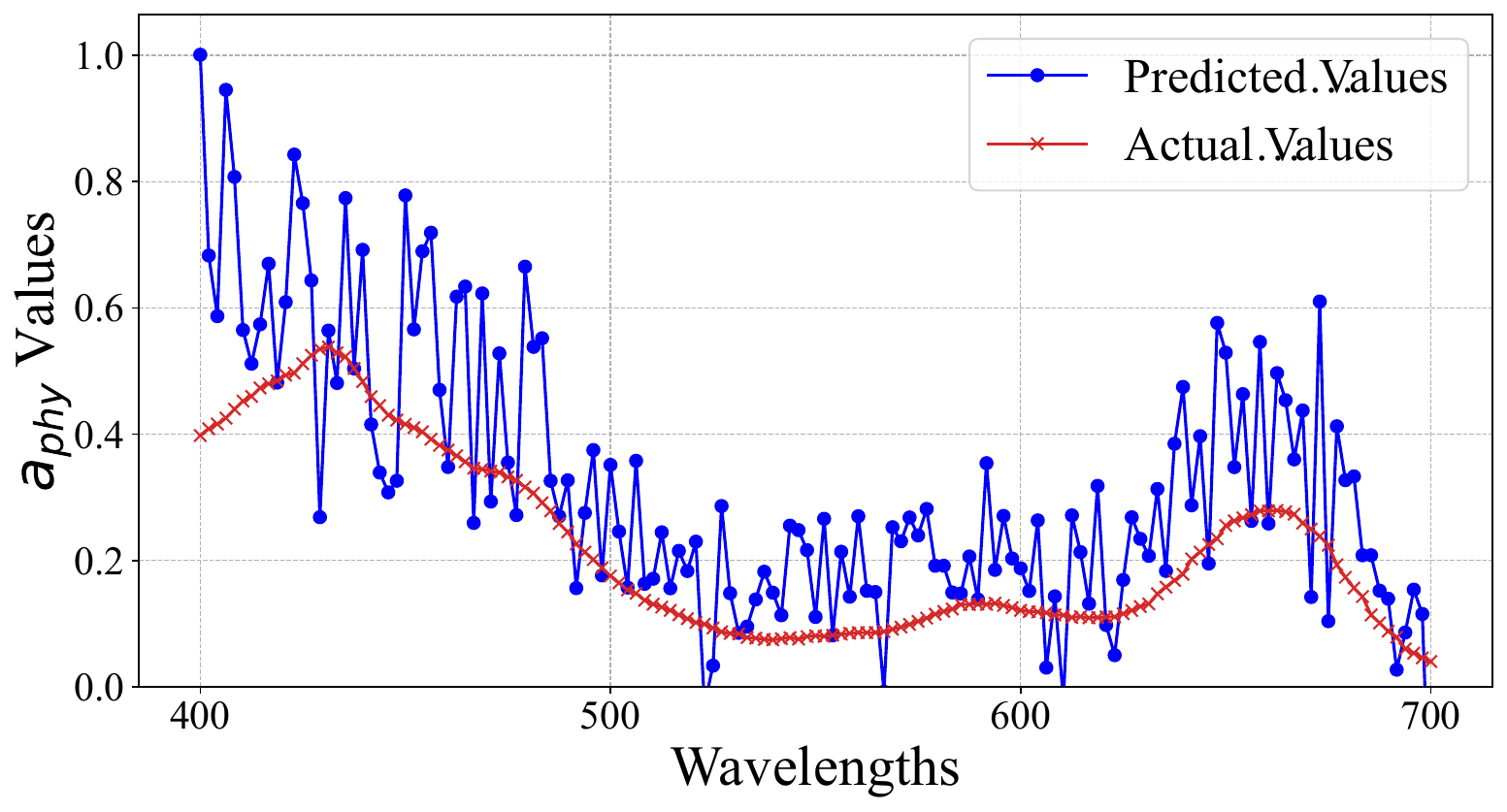}
		\label{fig:fit:real:7}
	}
	\subfigure[M-MDN prediction on PACE example 2]{
		\includegraphics[width=0.3\linewidth]{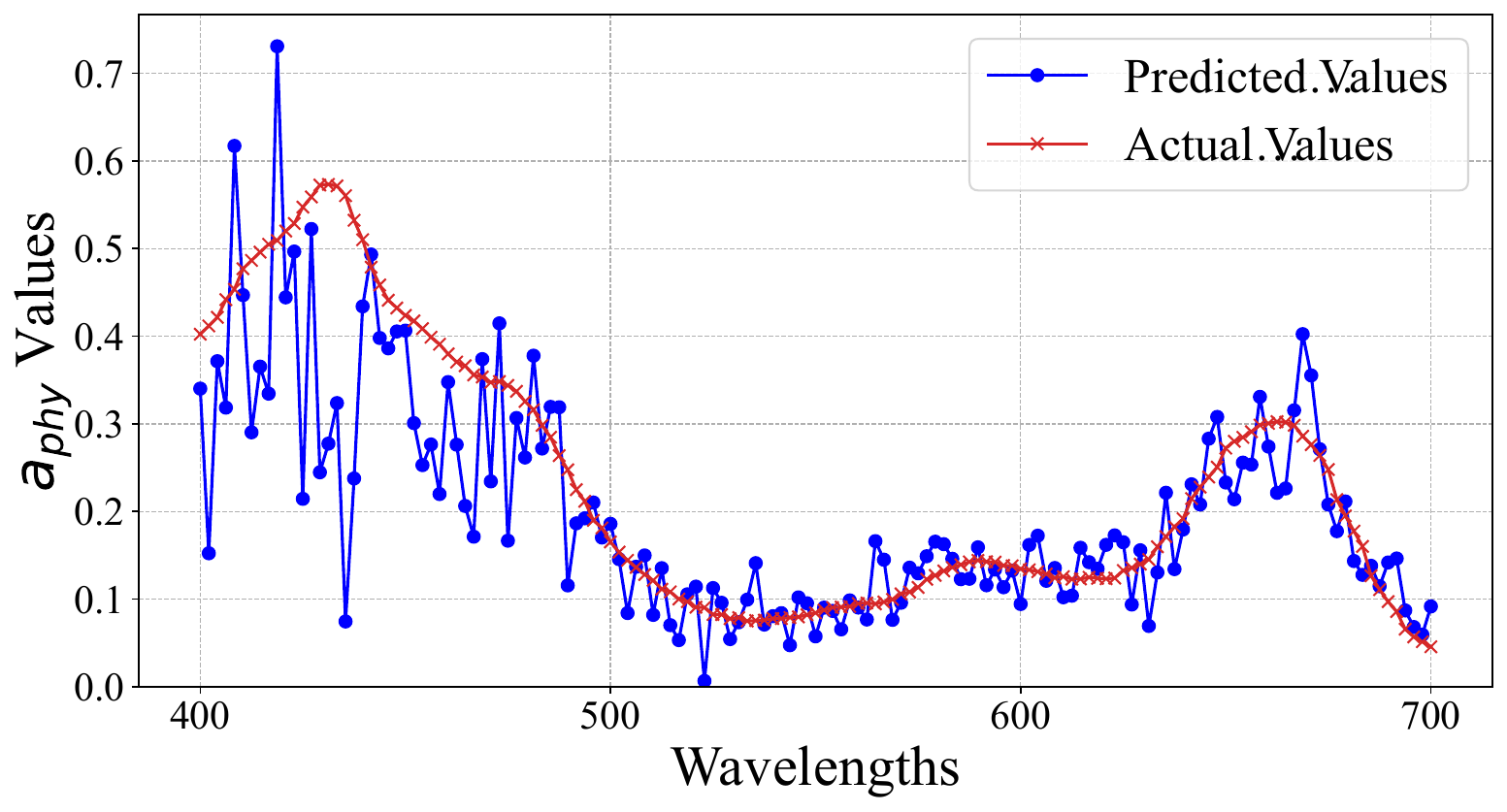}
		\label{fig:fit:real:8}
	}
	\subfigure[M-MDN prediction on PACE example 3]{
		\includegraphics[width=0.3\linewidth]{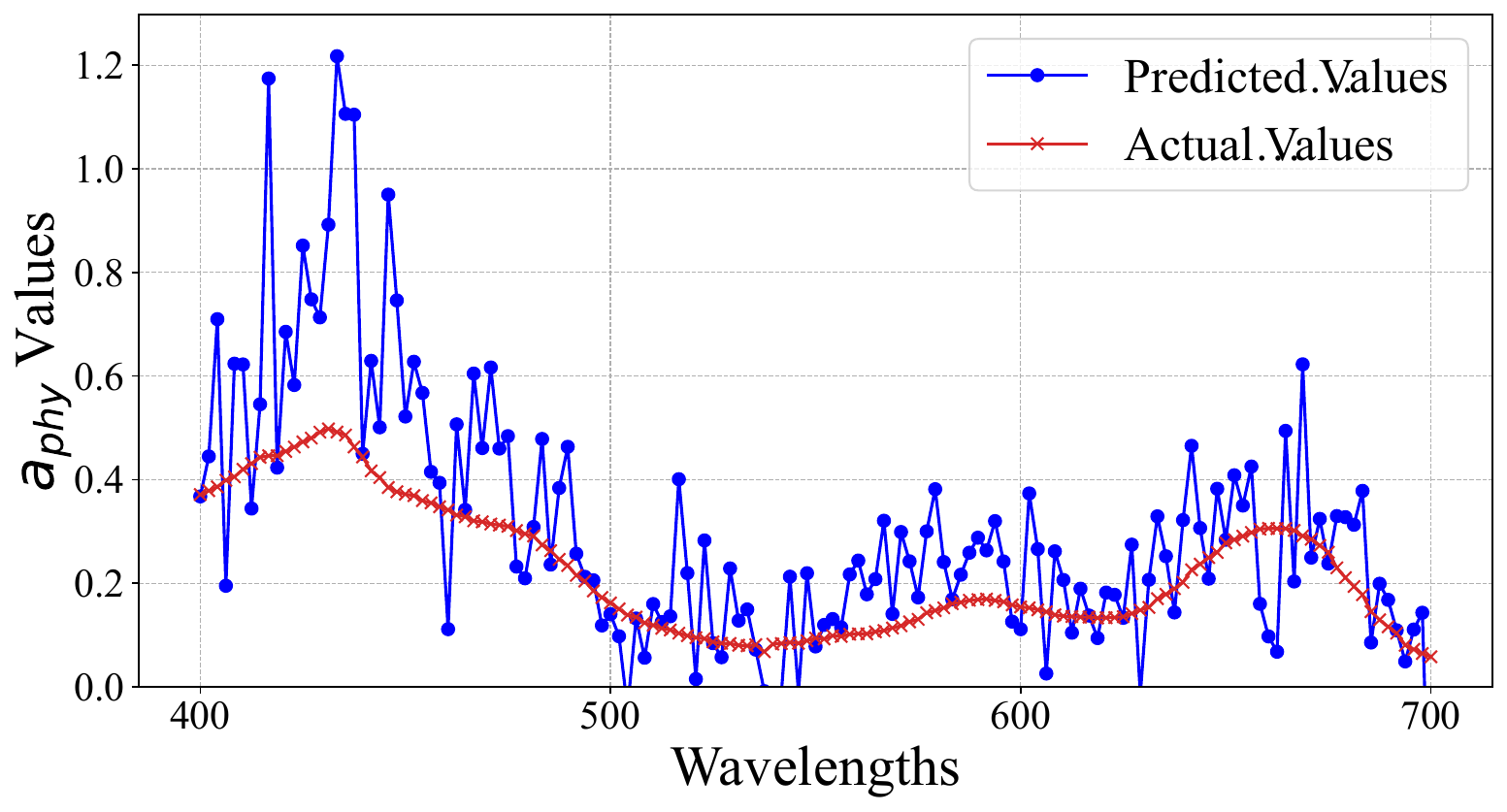}
		\label{fig:fit:real:9}
	}
	\caption{ Comparison of actual and predicted  $\aphy$ spectra at PACE wavelength setting in the range of 400–700 nm, using (a)–(c) VAE and (d)–(f) M-MDN..}
	\label{fig:fit:real}
\end{figure*}

\begin{figure*}[h]
	\centering
	\subfigure[VAE prediction on EMIT example 1]{
		\includegraphics[width=0.3\linewidth]{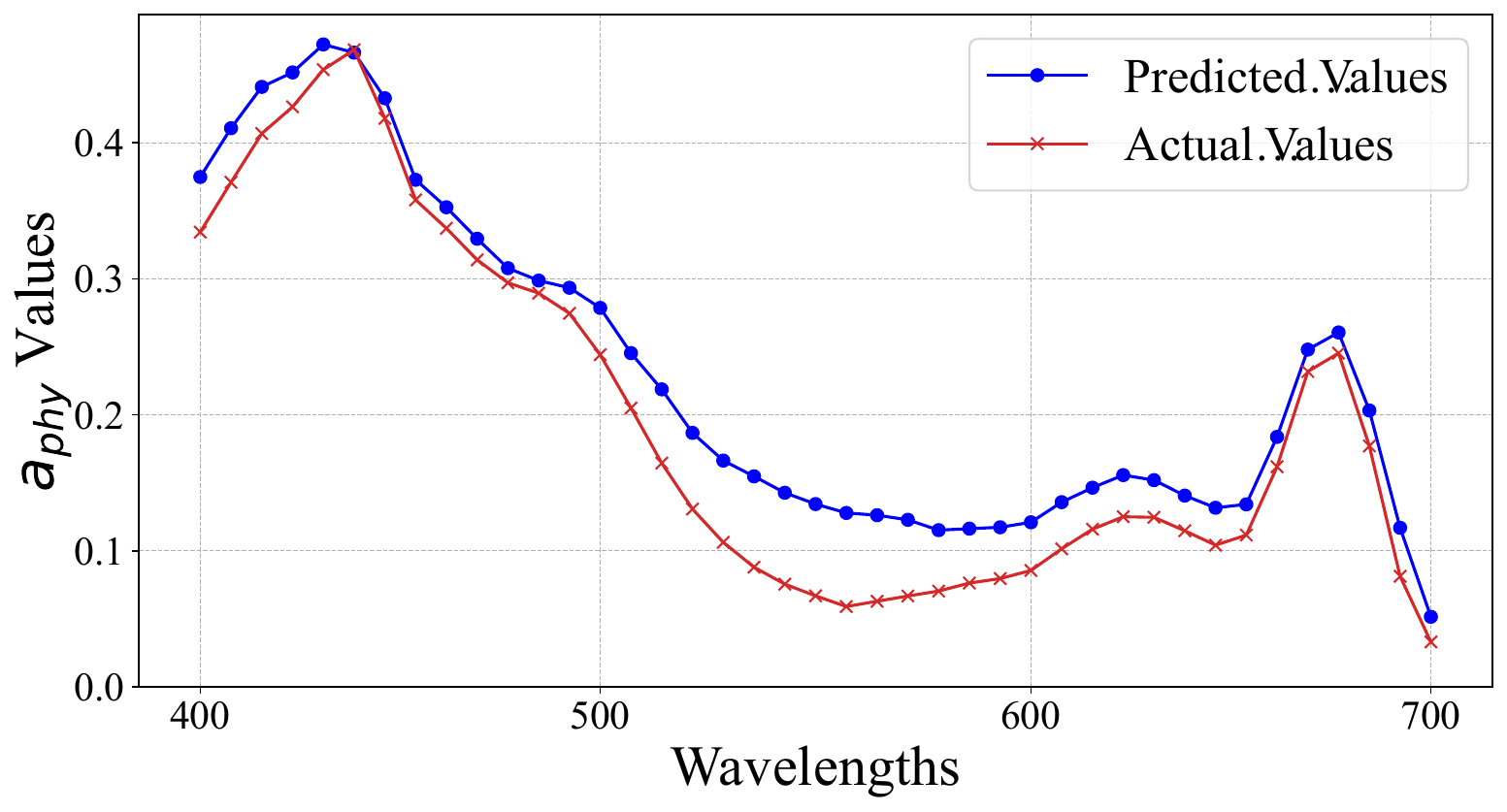}
		\label{fig:fit:real2:1}
	}	
  \subfigure[VAE prediction on EMIT example 2]{
		\includegraphics[width=0.3\linewidth]{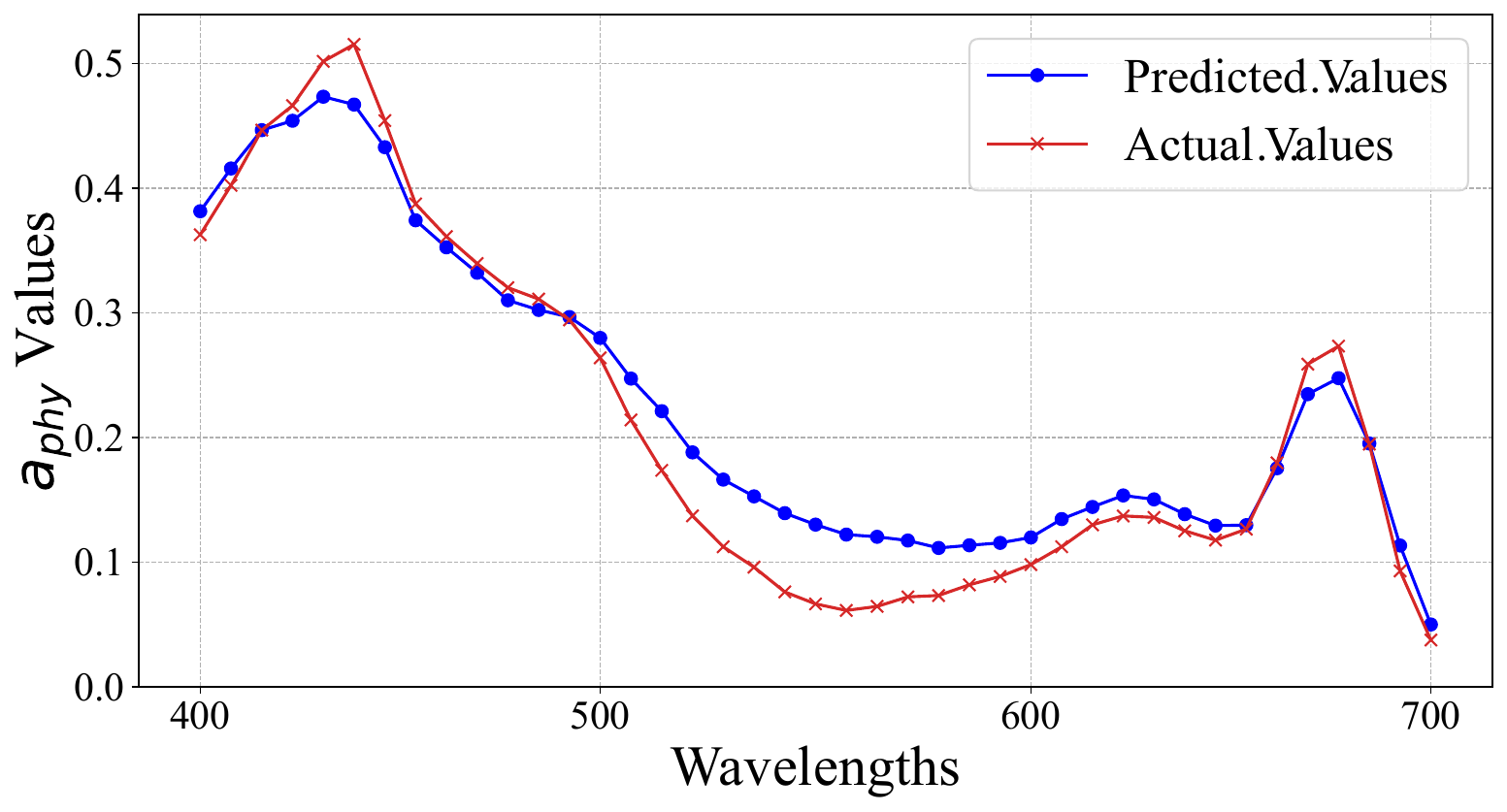}
		\label{fig:fit:real2:2}
	}
  \subfigure[VAE prediction on EMIT example 3]{
		\includegraphics[width=0.3\linewidth]{figs/EMIT_Oct_13.pdf}
		\label{fig:fit:real2:3}
	}	
	\subfigure[M-MDN prediction on EMIT example 1]{
		\includegraphics[width=0.3\linewidth]{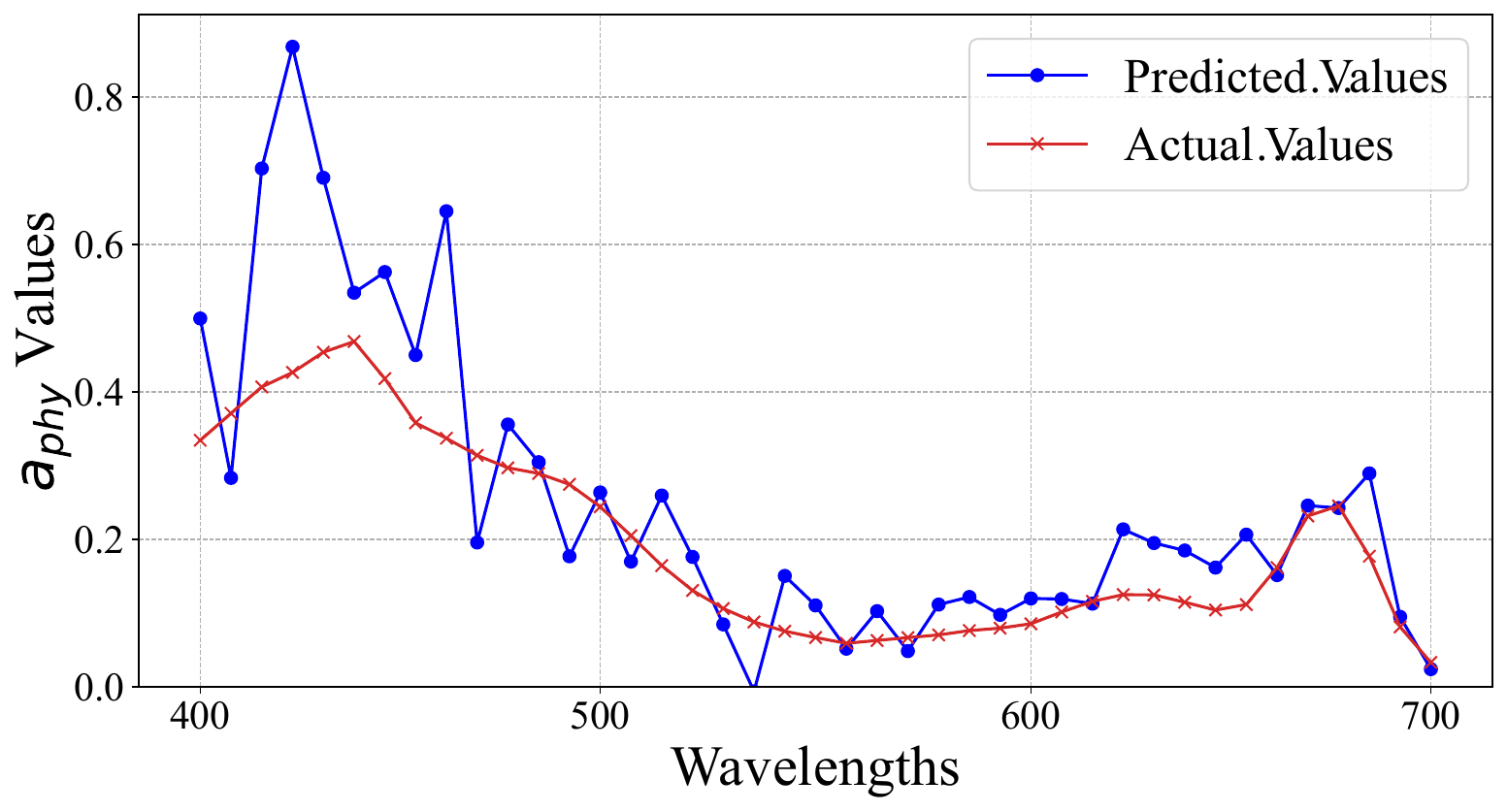}
		\label{fig:fit:real2:4}
	}
	\subfigure[M-MDN prediction on EMIT example 2]{
		\includegraphics[width=0.3\linewidth]{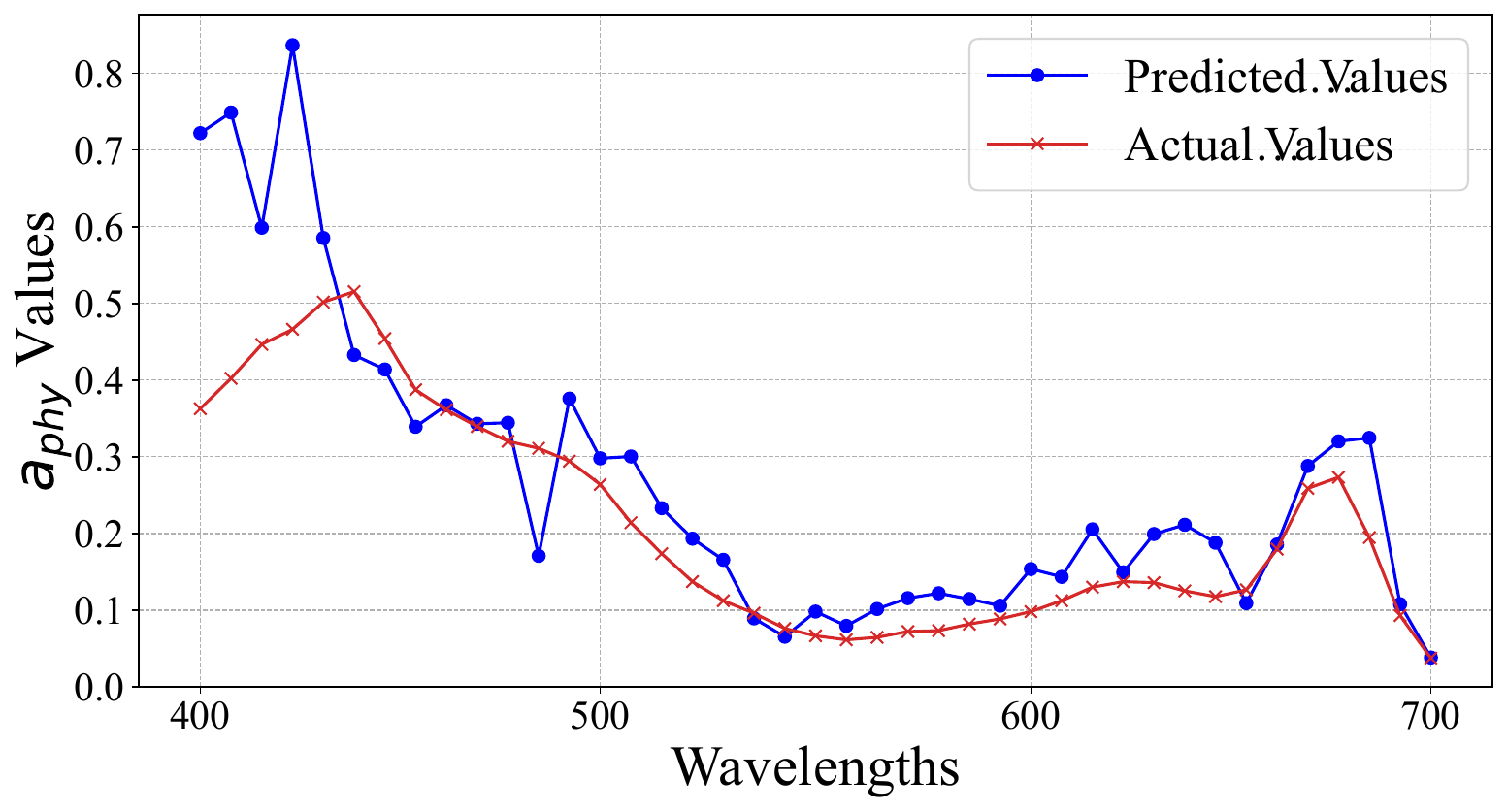}
		\label{fig:fit:real2:5}
	}
	\subfigure[M-MDN prediction on EMIT example 3]{
		\includegraphics[width=0.3\linewidth]{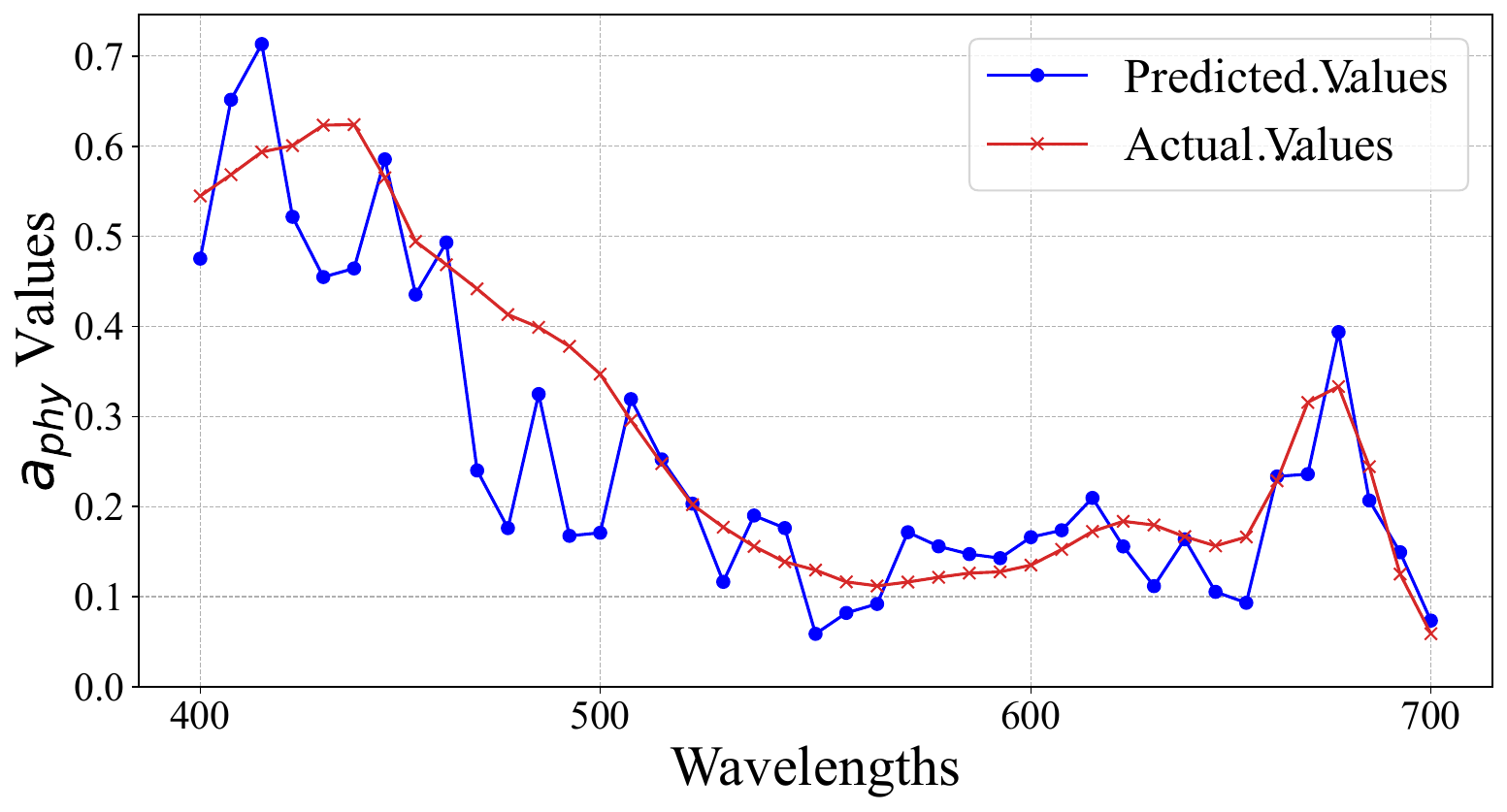}
		\label{fig:fit:real2:6
	}
    }
    
% \vspace{-1em}
	\caption{Comparison of actual and predicted  $\aphy$ spectra at EMIT wavelength setting in the range of 400–700 nm, using (a)–(c) VAE and (d)–(f) M-MDN.}
  % \vspace{-1em}
	\label{fig:fit:real2}
\end{figure*}

{
\subsection{Comparisons of Generality for $\aphy$ Predictions}
We further assess the robustness and generality of the trained VAE-$\aphy$ and M-MDN models by applying them to a separate in-situ dataset from Galveston Bay, TX, USA, which was not included in the training process. 
This assessment verifies the models’ ability to generalize $\aphy$ estimation to unseen data. 
This in-situ dataset from Galveston Bay was collected one and two months after Hurricane Harvey in 2017, during which the optical properties were highly complex. 
The one-month post-hurricane data represents turbid waters, while the two-month dataset captures productive waters with high Chl-a concentrations~\cite{liu2019floodwater}. 
Figs.~\ref{fig:fit:real} and~\ref{fig:fit:real2}  present three examples of actual and predicted $\aphy$ spectra across the 400-700 nm range at PACE and EMIT spectral settings, evaluating the ability of VAE and M-MDN models to generalize across different unseen geographic locations.
We consistently observe robust and stable estimations from VAE-$\aphy$ model in Figures~\ref{fig:fit:real} and~\ref{fig:fit:real2}.  
In addition to capturing the two pronounced Chl-a absorption peaks at 440 nm and 670 nm, the model effectively captures subtle variations in the range of 450-500 nm, a critical spectral region for differentiating phytoplankton groups. 
This also holds true for EMIT, which has a slightly lower spectral resolution (7.4 nm). 
Moreover, minor variations along the $\aphy$ spectra, such as those in the 480–500 nm and 600–620 nm ranges, were accurately predicted (Figure~\ref{fig:fit:real2}). 
However, predictions for EMIT show slightly greater deviations in the 500–600 nm range compared to PACE.
Overall, the $\aphy$ predictions from the VAE-$\aphy$ model closely align well with the actual values across all wavelengths, whereas M-MDN predictions exhibit zigzag patterns, indicating instability. 
This difference in performance stems from the VAE’s learning structure, which does not directly fit the $\aphy$ distribution from the training data but instead learns a Gaussian distribution to capture the high-level patterns in the $\Rrs$-$\aphy$ relationship for predictions. 
These experimental results highlight the practical advantages of using VAE, such as improved generality to unseen data and greater stability across wavelengths, in addressing one-to-many inversion challenges in ocean color applications. 
}

%It is observed that VAE-$\aphy$ performs well in predicting $\aphy$ at the higher Chl-a concentrations, such as, 30 and 50 $\mu$g$L^(-1)$, compared to lower Chl-a levels (e.g., 5.0 $\mu$g$L^(-1)$; Figs. 10 and 11), which could be attributed to underrepresentation of this water type included in the training datasets. 
%This lower Chl-a site was located at the entrance of Galveston Bay where it connects to the relatively clear seawaters of the northern Gulf of Mexico, is expected to exhibit different optical properties from those of the training dataset in [6] and other sites in the TX dataset [22]. 
%This difference might constrain the models' ability to learn these relevant patterns, resulting in reduced estimation accuracy for both MDN and VAE-$\aphy$ models (e.g., Figs. 10a and 10d). 
%While VAE-$\aphy$ predictions accurately capture the magnitude of $\aphy$ spectra at where Chl-a reach 30 and 50 $\mu$g$L^(-1)$, it is worth noting that a few minor absorption peaks associated with other pigments, such as Chl b, Chl c, and carotenoids in the 450-500 nm range (Figs. 10c and 11c) are missing from the predictions [19]. 
%This suggests that while VAE-$\aphy$ performs well overall, the training dataset may not be comprehensive enough to fully capture the diversity of water types, such as algal blooms. 
 
%At the same time, they highlight areas for improvement to enhance its applications in hyperspectral ocean color remote sensing.

\section{DISCUSSION and FUTURE WORK}
\label{discussion}

In this section, we discuss the advantages and reasoning behind employing the VAE structure for predicting $\aphy$ and Chl-a, as well as its potential for future applications in ocean color remote sensing. 

\subsection{Balancing Uncertainty and Reliability for Multi-value Prediction}
Predicting $\aphy$ and Chl-a through hyperspectral $\Rrs$ data involves addressing one-to-many problems, which stems from the fact that for a given $\Rrs$ spectrum, the inverse function of $\Rrs$ is not unique as a single $\Rrs$ may correspond to multiple combinations of IOPs and associated concentrations~\cite{defoin2007ambiguous,sydor2004uniqueness}, especially when the spectral resolution is limited. 
{
While MDN-based approaches can be rectified to support one-to-many predictions, such as the M-MDN method described in Section~\ref{revisingMDN}, obtaining a specific prediction requires random sampling from the learned distribution. 
This process inherently introduces additional uncertainty. 
For instance, when sampling from a mixture of Gaussian distributions with multiple peaks, the resulting predictions can vary significantly. 
This variability makes it challenging to achieve consistent and reliable estimates for $\aphy$ and Chl-a, as different samples may lead to fluctuating predicted values, reducing the model’s overall stability.}
In contrast, VAEs can solve the one-to-many prediction problem and avoid the uncertainty associated with the distribution nature. 
This can be explained from two perspectives. First, the encoder learns the latent vectors from the probability distribution learned from the input data $\Rrs$ and introduces randomness through the reparameterization operation~\cite{rezende2014stochastic,burda2015importance,doersch2016tutorial}, as shown in Eq. (4), enabling the VAE to generate different outputs from the same input. 
Second, the decoder directly learns the mapping between the learned variable latent vectors and the derived prediction data, allowing VAEs to generate accurate predictions from the latent distribution. 
This approach reduces the randomness and uncertainty associated with the sampling process, resulting in more reliable predictions. 

\begin{figure*}[]
	\centering
	\subfigure[444 nm under 40 bins]{
		\includegraphics[width=0.3\linewidth]{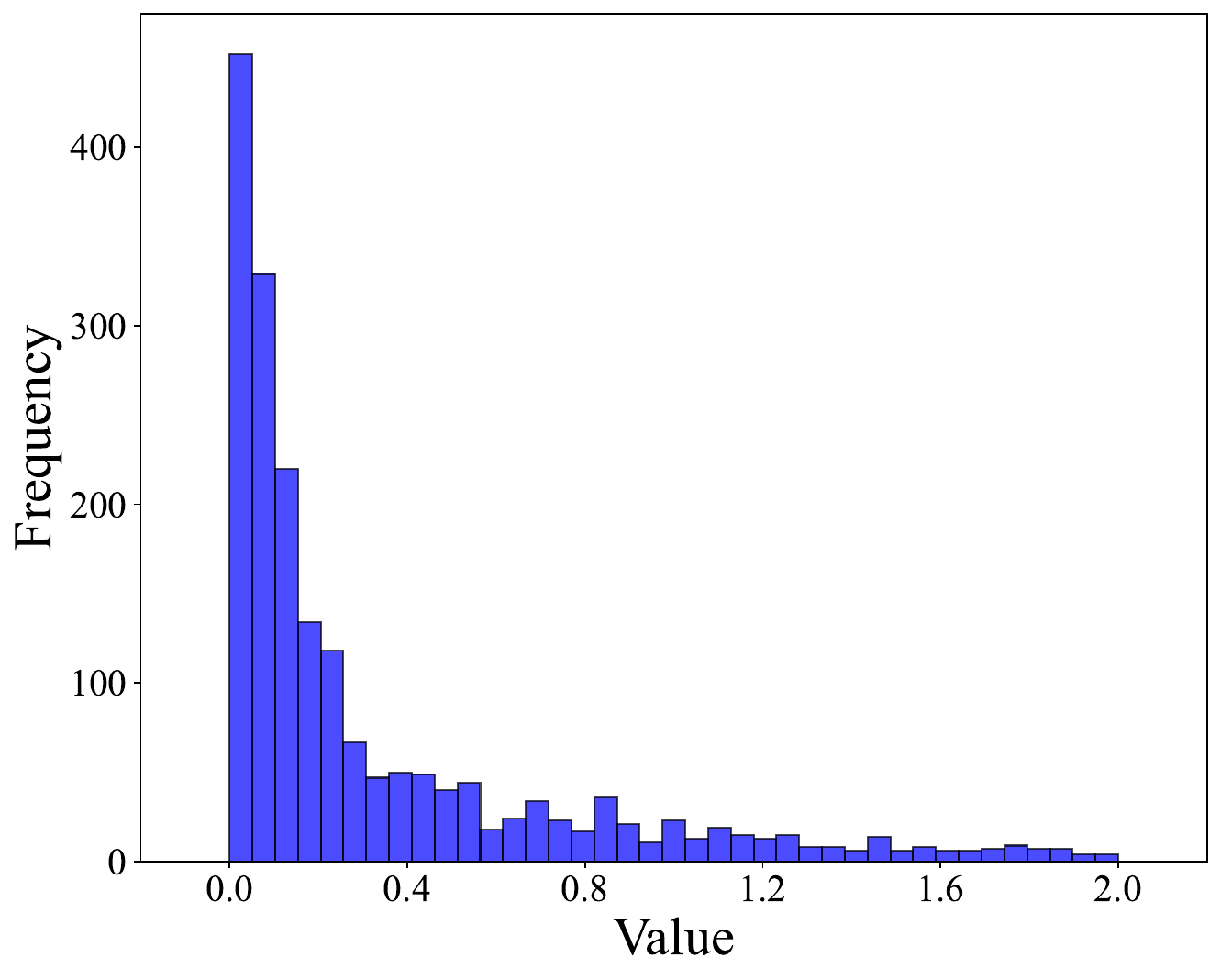}
		\label{fig:dis:1}
	}	
  \subfigure[444nm under 100 bins]{
		\includegraphics[width=0.3\linewidth]{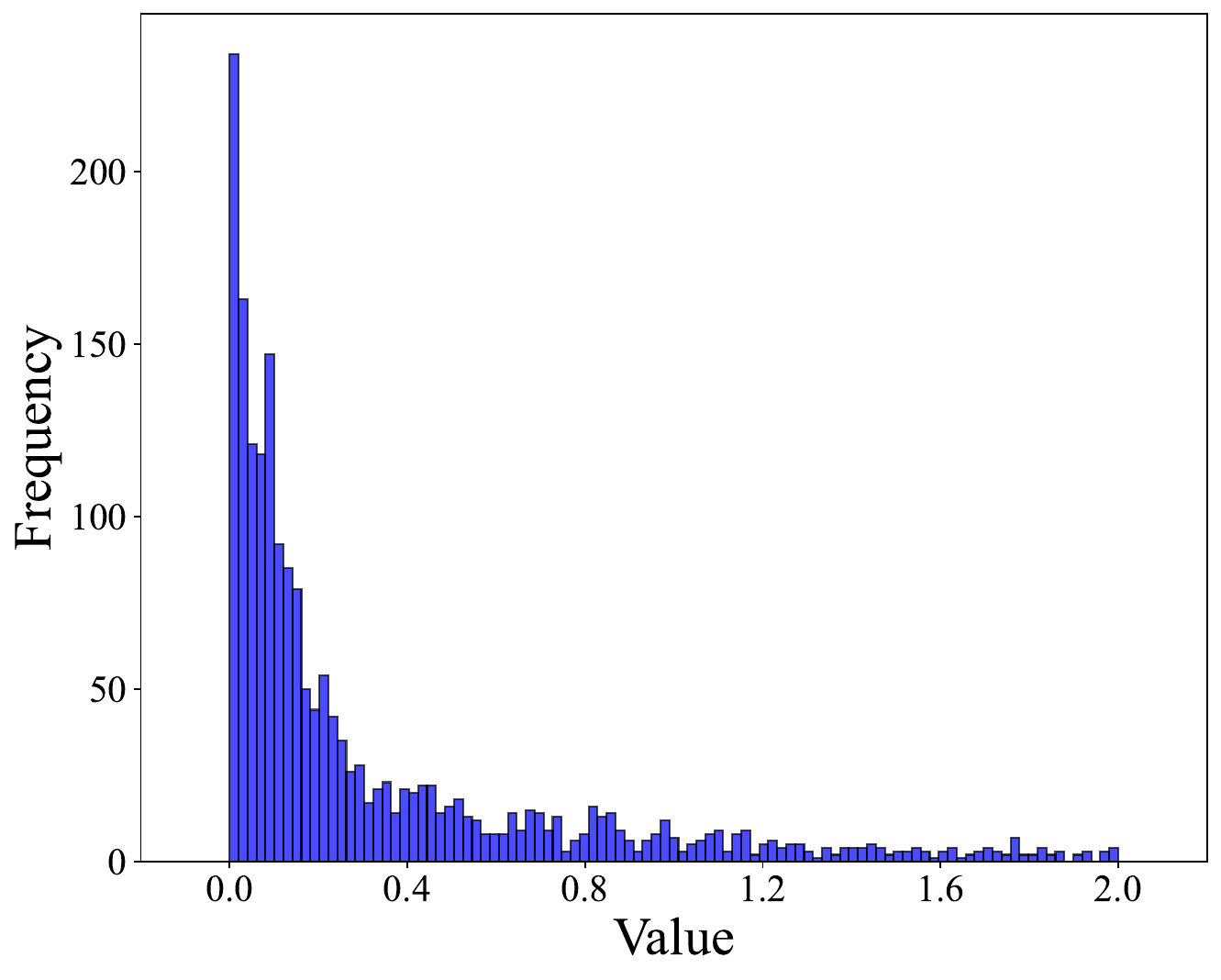}
		\label{fig:dis:2}
	}
  \subfigure[621 nm under 40 bins]{
		\includegraphics[width=0.3\linewidth]{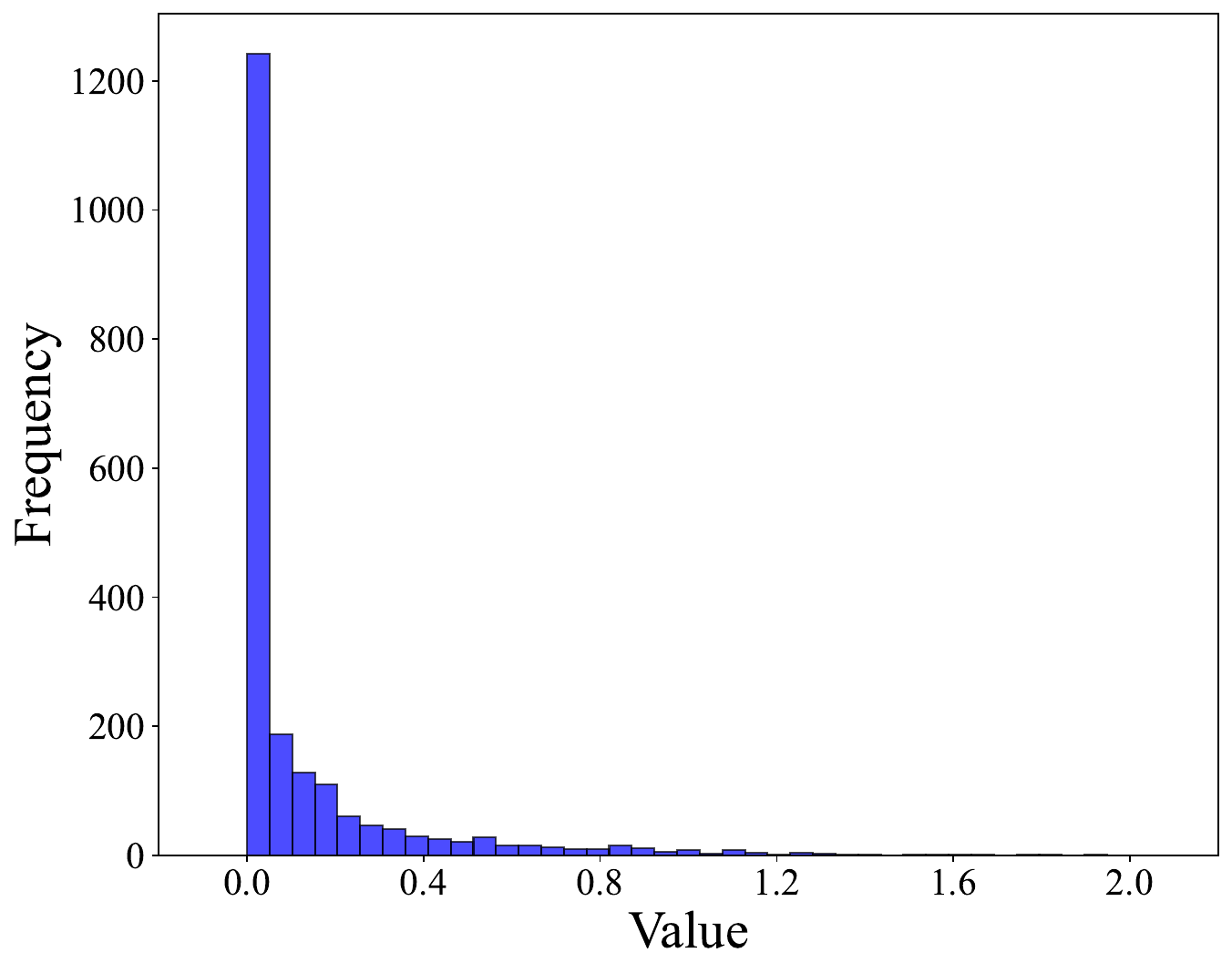}
		\label{fig:dis:3}
	}	
	\caption{ The $\aphy$ ($m^(-1)$) distribution at different wavelengths.}
	\label{fig:dis}
\end{figure*}

% \begin{figure}[h]
% 	\centering
% 	\includegraphics[width=0.8\linewidth]{figs/chla_dis.pdf}
% 	\caption{Distribution of data samples based on Chl a concentrations ($\mu$g $L^(-1)$ in the dataset.}
% 	\label{fig:chladis}
% \end{figure}

\subsection{Handling Complex Data Distributions}
Phytoplankton absorption coefficient $\aphy$ is one of the IOPs with the most complex distributions, often varying significantly across geographic locations. 
Figures~\ref{fig:dis}(a)-(b) display the $\aphy$ data distributions at 444 nm under different bin settings, where larger bin values indicate finer resolution of data variability, capturing multimodal distributions and avoiding oversimplification. %higher data precision, and distributions at 621nm. 
When comparing Figs.~\ref{fig:dis:1} and~\ref{fig:dis:2}, it is found that higher data precision leads to more peaks in the $\aphy$ distribution, suggesting that fitting the $\aphy$ data distribution at different scenarios may require different numbers of Gaussian components. 
Moreover, it is observed from Figs.~\ref{fig:dis:1} and~\ref{fig:dis:3} that the $\aphy$ distribution also varies significantly across different wavelengths due to absorption peaks of specific pigments. 
%The presence of more distinct peaks in the $\aphy$ distribution under 100 bins suggests that accurately predicting $\aphy$ distributions using these learning frameworks may require varying numbers of Gaussian components to account for the increased complexity in data with higher spectral resolution. 
%As observed in Figures 11a and 11b, the $\aphy$ distribution varies significantly across different wavelengths, primarily due to absorption peaks of specific pigments. 
As such, the MDN, which aims to fit $\aphy$ distributions at different wavelengths and Chl-a distributions, with a finite and fixed number of Gaussian distributions, may fail to produce accurate predictions in practice.  
In contrast, VAEs use Gaussian distributions to represent each element in the latent vector, aiming at capturing high-level patterns in the $\Rrs$ spectra for $\aphy$ and Chl-a prediction. Despite the complex distributions of $\aphy$ spectra, the underlying latent patterns can be more reasonably assumed to follow Gaussian distributions~\cite{blei2017variational,higgins2017beta}. 
By learning these latent patterns, VAE effectively captures the high-level inherent relationship between $\Rrs$ and $\aphy$ or Chl-a, eliminating the need to assume and fit mixed Gaussian distributions for target prediction. This enables VAEs to model intricate data distributions more efficiently and flexibly. In addition, it also supports VAE models in maintaining acceptable performance in prediction tasks on unseen data, even when the distributions differ from those of the training data, as shown in Figs.~\ref{fig:fit:real} and~\ref{fig:fit:real2}.

\subsection{Robustness to Noise}
The training data, i.e., $\Rrs$, $\aphy$, and Chl-a typically contain a certain level of noise. While data quality control and preprocessing help filter out clearly abnormal values, a certain degree of noise could remain inevitable due to in-situ sampling variability.
{
MDN models the relationship between $\Rrs$ and $\aphy$/Chl-a by fitting a mixture of Gaussian distributions to the training data.
However, when noise is present, MDN tends to fit Gaussian components to all observed variations, including those introduced by noise~\cite{balasubramanian2025mixture,rothfuss2019noise}. 
Since the model does not inherently distinguish between meaningful trends and random fluctuations, it may allocate Gaussian components to spurious variations rather than true signal patterns. 
This results in overfitting, where the learned distribution reflects noise-induced variability instead of capturing the underlying physical relationships. 
Consequently, MDN predictions can become unstable, especially when applied to new data with different noise characteristics.}
In contrast, VAE-based approaches are more robust to noisy data for two reasons. 
First, KL divergence loss, represented as the second term in Eqn.~(\ref{loss}), served as a regularization term. It constrains the learned latent distribution to remain close to a prior distribution, which helps prevent the latent vector from overfitting the noise~\cite{blei2017variational}. 
Additionally, the reparameterization trick naturally introduces noise to the latent vector, as shown in Eq.~(\ref{repara}), enabling the VAEs to inherently generate accurate predictions under noisy training data~\cite{burda2015importance,doersch2016tutorial}. 
These mechanisms help VAEs maintain stability and robustness, producing more consistent predictions for $\aphy$ and Chl-a even in noisy environments.
%{Adding noise to the test data is a straightforward and effective approach for evaluating the relative robustness of VAE and MDN to noise. 
%However, ensuring that the added noise is meaningful rather than purely random is crucial, as arbitrary noise could distort essential information in $R_{rs}$, potentially degrading the performance of both models. 
%In future work, we will further explore this aspect by deliberately designing noise patterns to more effectively assess model robustness.}

%While adding noise to the test data is a straightforward and effective experiment for evaluating the relative robustness of the VAE and MDN on noise. However, introducing random noise may fundamentally distort meaningful information in $R_{rs}$, leading to performance degradation in both MDN and VAE methods. In our future work, we will further explore this aspect by deliberately crafting noise to assess the model's robustness more effectively.}

{
\subsection{Generalizing to High-Dimensional Data}
In this study, VAE- and MDN-based learning frameworks are applied for $\aphy$ predictions for both the PACE and EMIT missions. 
EMIT has a lower-dimensional input space (N=41) compared to PACE, which has 141 spectral bands, making $\aphy$ predictions the highest-dimensional to date. 
This further emphasizes the need for robust learning frameworks capable of handling such high-dimensional data. 
Since PACE was just launched on February 8, 2024, no prior studies have yet been published exploring high-dimensional $\aphy$ predictions using learning frameworks. 
For example, in previous studies~\cite{pahlevan2021hyperspectral,pahlevan2020seamless,o2023hyperspectral}, hyperspectral data from HICO and PRISMA required at most 50-dimensional inputs for $\aphy$  prediction tasks and 67-dimensional inputs for phycocyanin prediction tasks. With the higher spectral resolution of PACE-OCI, VAEs, and MDNs are now expected to handle even higher-dimensional data input, enabling accurate predictions across finer wavelength intervals. However, MDNs encounter computational challenges in this setting. 
Since MDNs model the output as a mixture of multivariate Gaussian distributions, they need to learn a full covariance matrix for each Gaussian component. Specifically, for the input of $N$ dimensions with $M$ Gaussian components, the MDN must estimate an $M \times N \times times$ covariance matrix, meaning the number of parameters grows quadratically with respect to $N$. 
This will incur computational inefficiencies and challenges in model stability, primarily due to two key factors. First, as the dimensionality increases, the covariance matrix often becomes ill-conditioned or nearly singular, making the model highly sensitive to small numerical errors. 
These small numerical errors can result in inconsistent and unreliable parameter updates, leading to exploding or vanishing gradients during training. Second, the full covariance matrix introduces O($N^2$) parameters per component, significantly increasing the complexity of the loss function. 
This expansive parameter space makes it challenging for gradient-based optimizers to find stable updates, often resulting in slow convergence or divergence. In contrast, VAEs mitigate these challenges by mapping high-dimensional data into a latent space, where each dimension follows a simple Gaussian distribution. 
Rather than modeling full covariance matrices, VAEs learn independent latent features that capture essential spectral patterns, allowing for effective feature extraction and dimensionality reduction while preserving crucial spectral information. 
This structure significantly reduces the number of parameters to optimize and eliminates the numerical instability issues associated with MDNs. 
In our experiments, conducted on a workstation equipped with an Intel Core i9-13900K CPU and an NVIDIA RTX 4090 GPU (24GB VRAM), training the VAE for 2000 epochs took approximately 30 seconds, compared to roughly 500 seconds for MDN model with full covariance matrix. 
Given these advantages, VAEs present a more scalable alternative for hyperspectral inversion retrieval tasks, such as $\aphy$ and Chl-a prediction explored in this study. }
%VAE approach also demonstrates strong potential for scalability and generalizability in handling increasingly complex hyperspectral prediction challenges in the future.} 

\subsection{Forward Path}
\label{sub:forward}

Moving forward, the VAE framework introduced in this study will be applied to PACE and EMIT hyperspectral imagery; however, evaluating its performance on these satellite datasets remains an ongoing effort. 
PACE, launched on February 8, 2024, provides daily global hyperspectral imagery, though corresponding field validation is still in progress.
EMIT, on the other hand, has conducted irregular sampling across estuaries worldwide, making consistent match-up sampling a challenge—not only for this study but on a global scale. 
As these missions are still in their early stages, extensive field validation is required for both PACE and EMIT to better understand the uncertainties involved in estimating $a_{phy}$ and Chl-a from satellite-derived $R_{rs}$ using learning-based models. 
Importantly, these challenges do not conflict with the current work presented in this study; rather, they represent parallel and complementary efforts. 
As satellite $R_{rs}$ products continue to be calibrated and validated, the proposed algorithms can be more effectively applied, particularly as more paired \textit{in situ} $R_{rs}$–$a_{phy}$ datasets along with other IOPs become available on the same day as satellite overpasses (e.g., from PACE and EMIT).

On the other hand, while machine learning approaches are increasingly favored in hyperspectral ocean color remote sensing, there is a persistent need for large training datasets encompassing diverse water quality conditions. 
Consequently, there is a growing demand for more coincident reflectance-biophysical parameter pairs to be acquired across different water types, environmental conditions, and geographic locations. 
Unfortunately, the currently available in-situ $\Rrs$-$\aphy$ and $\Rrs$-Chl-a datasets are limited and contain uncertainties.
This is because the data have been collected by various groups using different instruments and calibration or correction approaches, all of which contribute to uncertainties in machine learning predictions. 
Additionally, current available global coastal bio-optical datasets—such as $\Rrs$–$\aphy$ and $\Rrs$–Chl-a in GLORIA—lack true one-to-many features.
To further enhance our VAE model’s generality across different geo-locations and water types, two lines of direction can be explored. 
First, considering the difficulty of enriching the real field data samples, the radiative transfer model (RTM), traditionally used in ocean color remote sensing to simulate apparent optical properties (AOPs), such as $\Rrs$, based on precise characterization of IOPs, offers a method to generate a large dataset to complement the real data. 
The simulated $\Rrs$ over global coastal waters using RTM, such as HydroLight, employing hyperspectral $\aphy$, and scattering phase functions measured from diverse phytoplankton groups/genus~\cite{lain2023simulated,lomas2024phytoplankton} to represent the dynamic and varied phytoplankton community composition, along with varying concentrations of CDOM and NAP independent of phytoplankton in a wide range of oceanic and coastal waters, is expected to capture detailed spectral signatures in response to phytoplankton community variability in biogeochemically dynamic and optically complex coastal-estuarine environments. 
With a large and high-quality simulation dataset, a plausible solution is to first use it to pre-train the VAE model, allowing it to learn the abundant patterns, and then fine-tune the model using real data samples. This approach is expected to improve both model prediction accuracy and generalizability. 
{
Secondly, the simulated data would further support the extension of VAE-based models into a multi-head framework, enabling simultaneous prediction of multiple parameters, similar to the prior work~\cite{o2023hyperspectral}. A complete set of IOPs and AOPs would allow multiple decoders to be trained concurrently,  with each decoder dedicated to predicting a single IOP. Such a design preserves the benefits of multi-head prediction, allowing a single encoder to capture relationships among different IOPs while minimizing error propagation through independent decoders. 
Nonetheless, further research is needed from both data and algorithm perspectives for future exploration. to refine the VAE architecture used in this study.  
%the authors proposed a multi-head MDN-based model to predict multiple parameters simultaneously. 
%Similarly, our VAE-based approach can be adapted for multi-head prediction by concatenating $\aphy$ and Chl a as a joint output vector. 
%However, this setup may introduce trade-offs between targets, where prediction errors from one IOP could propagate to others, potentially amplifying bias or variance, especially in optically complex environments.
%Additionally, given the PACE wavelength with 144 bands, the single Chl-a value may be disproportionately influenced by the high-dimensional $\aphy$ values. 
%To address this, one plausible solution is to  
}

%Another direction is to use imagery data from NASA’s hyperspectral missions, like PACE, to train a large foundation model~\cite{zhou2024comprehensive}, which is enriched by diverse geolocation data, and can learn hyperspectral $\Rrs$ patterns. This approach allows for the development of a wide range of downstream models with real data for tasks such as $\aphy$, Chl-a, as well as other IOPs and concentrations, effectively addressing the issue of data scarcity.

\section{Conclusion}

In this paper, VAEs—originally designed for data generation and reconstruction—were tailored for the first time to prediction tasks in ocean color remote sensing. Two VAE-based models, VAE-$\aphy$ for high-dimension predictions and VAE-Chl-a for single-value predictions in optically-complex waters, targeting two key indicators of phytoplankton, i.e., Chl-a for abundance and $\aphy$ for diversity. 
The dataset used by \cite{pahlevan2021hyperspectral} and ~\cite{o2023hyperspectral} for the MDN model was utilized for VAE model development, validation, and testing in this study. These models were applied to support NASA’s hyperspectral missions, including PACE and EMIT. 
The configurations of both the VAE and MDN models were provided in this study to benefit the broader community. The performance of the VAE and MDN models was evaluated using eight metrics—MALE, RMSE, RMSLE, Log-Bias, Slope, and three median value-oriented metrics—to provide a comprehensive assessment of prediction accuracy. 
Overall, the VAE presented greater stability and robustness in predicting $\aphy$ across the PACE and EMIT spectral settings, demonstrating higher precision and lower bias in all eight metrics. 
The benefits of applying VAEs to hyperspectral data for high-dimensional predictions, such as PACE-predicted $\aphy$, were discussed in depth across three aspects: uncertainty and reliability, robustness to noise, and generalization to high-dimensional data. 
Additionally, a new dataset collected post-Hurricane Harvey in 2017, which presented different conditions and distributions compared to the training data, was used as a test dataset. 
Both the VAE and MDN models perform well in the predictions of $\aphy$ at longer wavelengths, with the VAE-based solution demonstrating better robustness in the range of 400-500 nm when applied to unseen data collected in Galveston Bay. 
Moving forward, using simulated data to expand the training dataset, representing diverse coastal waters would be one of the key solutions to overcoming data scarcity issues and enhancing model generalization for $\aphy$ predictions. 
This approach will also further enhance support for the multi-head predictions. This study advances hyperspectral retrievals of phytoplankton metrics, paving the way for our understanding of phytoplankton community dynamics in optically complex coastal waters.

\bibliographystyle{plain}
\bibliography{main}

\end{document}